\documentclass{article}

\usepackage{microtype}
\usepackage{graphicx}
\usepackage{booktabs} 
\usepackage{subcaption}

\usepackage[pagebackref=true]{hyperref}

\makeatletter
\newcommand\footnoteref[1]{\protected@xdef\@thefnmark{\ref{#1}}\@footnotemark}
\makeatother

\usepackage[accepted]{icml2024}

\usepackage{amsmath}
\usepackage{amssymb}
\usepackage{mathtools}
\usepackage{amsthm}

\usepackage{enumitem}
\usepackage{bigstrut}

\usepackage[capitalize,noabbrev]{cleveref}

\theoremstyle{plain}
\newtheorem{theorem}{Theorem}[section]
\newtheorem{proposition}[theorem]{Proposition}
\newtheorem{lemma}[theorem]{Lemma}

\theoremstyle{definition}

\newtheorem{innercustomgeneric}{\customgenericname}
\providecommand{\customgenericname}{}
\newcommand{\newcustomtheorem}[2]{%
  \newenvironment{#1}[1]
  {%
   \renewcommand\customgenericname{#2}%
   \renewcommand\theinnercustomgeneric{##1}%
   \innercustomgeneric
  }
  {\endinnercustomgeneric}
}

\newcustomtheorem{customassumption}{Assumption}

\theoremstyle{remark}

\usepackage[textsize=tiny]{todonotes}

\newcommand{\appropto}{\mathrel{\vcenter{
  \offinterlineskip\halign{\hfil$##$\cr
    \propto\cr\noalign{\kern2pt}\sim\cr\noalign{\kern-2pt}}}}}

\icmltitlerunning{\hfill Sequential Neural Score Estimation ~\hfill \thepage}

\usepackage{xcolor}  
\usepackage{soul}    

\begin{document}

\twocolumn[
\icmltitle{Sequential Neural Score Estimation: Likelihood-Free Inference with Conditional Score Based Diffusion Models}



\icmlsetsymbol{equal}{*}

\begin{icmlauthorlist}
\icmlauthor{Louis Sharrock}{equal,yyy,zzz}
\icmlauthor{Jack Simons}{equal,zzz}
\icmlauthor{Song Liu}{zzz}
\icmlauthor{Mark Beaumont}{zzz}
\end{icmlauthorlist}

\icmlaffiliation{yyy}{Department of Mathematics and Statistics, Lancaster University, UK}

\icmlaffiliation{zzz}{School of Mathematics, University of Bristol, UK}

\icmlcorrespondingauthor{Louis Sharrock}{l.sharrock@lancaster.ac.uk}

\icmlkeywords{Machine Learning, ICML}

\vskip 0.3in
]



\printAffiliationsAndNotice{\icmlEqualContribution} 


\begin{abstract}
We introduce Sequential Neural Posterior Score Estimation (SNPSE), a score-based method for Bayesian inference in simulator-based models. Our method, inspired by the remarkable success of score-based methods in generative modelling, leverages conditional score-based diffusion models to generate samples from the posterior distribution of interest. The model is trained using an objective function which directly estimates the score of the posterior. We embed the model into a sequential training procedure, which guides simulations using the current approximation of the posterior at the observation of interest, thereby reducing the simulation cost. We also introduce several alternative sequential approaches, and discuss their relative merits. We then validate our method, as well as its amortised, non-sequential, variant on several numerical examples, demonstrating comparable or superior performance to existing state-of-the-art methods such as Sequential Neural Posterior Estimation (SNPE).
\end{abstract}

\section{Introduction}
\label{sec:intro}
Many applications in science, engineering, and economics make use of stochastic numerical simulations to model complex phenomena of interest. Such simulator-based models are often designed by domain experts, using knowledge of the underlying principles of the process of interest. They are thus well suited to domains in which observations are best understood as the result of mechanistic physical processes. These include, amongst others, neuroscience \cite{Sterratt2011,Goncalves2020}, evolutionary biology \cite{Beaumont2002,Ratmann2007}, ecology \cite{Beaumont2010,Wood2010}, epidemiology \cite{Corander2017}, climate science \cite{Holden2018}, cosmology \cite{Alsing2018}, high-energy physics \cite{Brehmer2021}, and econometrics \cite{Gourieroux1993}.

In many cases, simulator-based models depend on parameters $\theta$ which cannot be identified experimentally, and must be inferred from data $x$. Bayesian inference provides a principled approach for this task. In particular, given a {prior} $p(\theta)$ and a {likelihood} $p(x|\theta)$, Bayes' Theorem gives the {posterior} distribution over the parameters as
\begin{equation}
p(\theta|x) = \frac{p(x|\theta) p(\theta)}{p(x)}
\end{equation}
where $p(x) = \int p(x|\theta) p(\theta) \mathrm{d}\theta$ is known as the evidence or marginal likelihood. The major difficulty associated with simulator-based models is the absence of a tractable likelihood function $p(x|\theta)$. This precludes, in particular, the use of conventional likelihood-based Bayesian inference methods such as Markov chain Monte Carlo (MCMC) \cite{Brooks2011} or variational inference (VI) \cite{Blei2017}. The resulting inference problem is often referred to as likelihood-free inference or simulation-based inference (SBI)  \cite{Cranmer2020,Sisson2018}.

\begin{figure*}[h!]
\vspace{-3mm}
  \centering
\includegraphics[width=.92\textwidth, trim=70 0 170 0, clip]{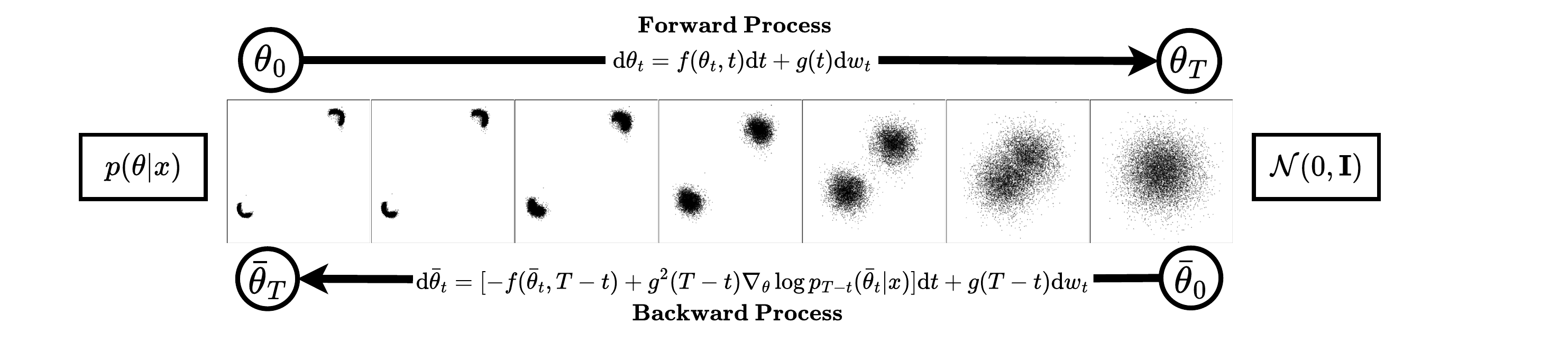}
  \vspace{-5mm}
  \caption{\textbf{Visualisation of posterior inference using Neural Posterior Score Estimation (NPSE) in the `Two Moons' experiment.} The forward process transforms samples from the target posterior distribution $\smash{p(\theta|x)}$ to a tractable reference distribution. The backward process transports samples from the reference to the target posterior. The backward process depends on the scores $\smash{\nabla_{\theta} \log p_t(\theta|x)}$, which can be estimated using score matching techniques given access to samples $\smash{(\theta,x)\sim p(\theta)p(x|\theta)}$ (see Section \ref{sec:SGM}).}
  \label{fig1}
  \vspace{-4mm}
\end{figure*}

Traditional methods for performing SBI include approximate Bayesian computation (ABC) \cite{Beaumont2002,Sisson2018}, whose variants include rejection ABC \cite{Tavare1997,Pritchard1999}, MCMC ABC \cite{Marjoram2003}, and sequential Monte Carlo (SMC) ABC \cite{Beaumont2009,Bonassi2015}. In such methods, one repeatedly samples parameters, and only accepts parameters for which the corresponding samples from the simulator are similar to the observed data $x_{\mathrm{obs}}$. 

More recently, a range of new SBI methods have been introduced, which leverage advances in machine learning such as normalising flows \cite{Papamakarios2017,Papamakarios2021} and generative adversarial networks \cite{Goodfellow2014}. These methods often include a sequential training procedure, which adaptively guides simulations to yield more informative data. Such methods include Sequential Neural Posterior Estimation (SNPE) \cite{Papamakarios2016,Lueckmann2017,Greenberg2019}, Sequential Neural Likelihood Estimation (SNLE) \cite{Lueckmann2019,Papamakarios2019}, and Sequential Neural Ratio Estimation (SNRE) \cite{Durkan2020,Hermans2020,Miller2021,Thomas2022}.  Other more recent algorithms of a similar flavour include Sequential Neural Variational Inference (SNVI) \cite{Glockler2022},  Generative Adversarial Training for SBI (GATSBI) \cite{Ramesh2022}, Truncated SNPE (TSNPE) \cite{Deistler2022}, and Sequential Unnormalized Neural Likelihood Estimation (SUNLE) \cite{Glaser2022}.

In this paper, we present Neural Posterior Score Estimation (NPSE), as well as its sequential variant (SNPSE). Our method, inspired by the remarkable success of score-based generative models \cite{Song2019,Song2021,Ho2020}, utilises a conditional score-based diffusion model to generate samples from the posterior of interest. While similar approaches \citep[e.g.,][]{Batzolis2021,Dhariwal2021,Song2021,Tashiro2021,Chao2022,Chung2022} have previously found success in a variety of problems, their application to SBI has not yet been widely investigated.\footnote{In parallel with an early version of this work, \citet{Geffner2022} also studied the use of diffusion models for SBI. We provide a comparison with this paper in Section \ref{subsec:SM_and_LFI} and Appendix \ref{app:LFI_multiple_obs}.}

In contrast to existing SBI approaches based on normalising flows (e.g., SNLE, SNPE), our approach only requires estimates for the gradient of the log density, or {score function}, of the intractable likelihood or the posterior, which can be approximated using a neural network via score matching techniques \cite{Hyvarinen2005,Vincent2011,Song2020}. Since we do not require a normalisable model, our method avoids the need for any strong restrictions on the model architecture. In addition, unlike methods based on generative adversarial networks (e.g., GATSBI), we do not require adversarial training objectives, which are notoriously unstable \cite{Metz2017,Salimans2016}.

We first discuss how conditional score-based diffusion models can be used for SBI. We then outline how our approach can be embedded within a principled sequential training procedure, which guides simulations towards informative regions using the current approximation of the posterior. We outline in detail a number of possible sequential procedures, several of which could also be used to develop sequential variants of amortised algorithms more recently proposed in the SBI literature \citep[e.g.,][]{Dax2023}. We then advocate for our preferred method, Truncated Sequential NPSE (TSNPSE), which uses a series of truncated proposals inspired by the approach in \citet{Deistler2022}. We validate our methods on several benchmark SBI problems as well as a real-world neuroscience problem, obtaining comparable or superior performance to other state-of-the-art methods.

\section{Simulation-Based Inference with Diffusion Models}
\label{sec:score-matching}

\subsection{Simulation-Based Inference}
\label{subsec:LFI}
Suppose that we have access to a simulator which, given input parameters $\theta\in\mathbb{R}^d$,  generates synthetic data $x\in\mathbb{R}^p$. We assume that parameters are distributed according to some known prior $p(\theta)$, but that the likelihood $p(x|\theta)$ is intractable. Given an observation $x_{\mathrm{obs}}$, we are interested in generating samples from the posterior distribution $p(\theta|x_{\mathrm{obs}}) \propto p(\theta)p(x_{\mathrm{obs}}|\theta)$, given a finite number of i.i.d. samples $\{(\theta_i,x_i)\}_{i=1}^N\sim p(\theta) p(x|\theta)$.

\subsection{Diffusion Models for Simulation-Based Inference}
\label{sec:SGM}
We propose to tackle this problem using conditional score-based diffusion models \citep[e.g.,][]{Song2021}. In such models, noise is gradually added to the target distribution using a diffusion process, resulting in a tractable reference distribution, e.g., a standard Gaussian. The time-reversal of this process is also a diffusion process, whose dynamics can be approximated using score matching \cite{Hyvarinen2005,Vincent2011,Song2020a,Song2021}. One can thus generate samples from the target distribution by simulating the approximate reverse-time process, initialised at samples from the reference distribution.

More concretely, we begin by defining a forward {noising} process $(\theta_t)_{t\in[0,T]}$ which, initialised at $\theta_0\sim p(\cdot|x)$, evolves according to the stochastic differential equation (SDE)
\begin{align}
\mathrm{d}\theta_t &= f(\theta_t,t) \mathrm{d}t + g(t)\mathrm{d}w_t,
\label{eq:forwardSDE} 
\end{align}
where $f:\mathbb{R}^d\times \mathbb{R}_{+}\rightarrow\mathbb{R}^d$ is the drift coefficient, $g:\mathbb{R}_{+}\rightarrow\mathbb{R}^d$ is the diffusion coefficient, and $(w_t)_{t\geq0}$ is a standard $\mathbb{R}^d$-valued Brownian motion. The coefficients $f$ and $g$ are chosen such that, for all $x\in\mathbb{R}^p$, the forward noising process admits a unique stationary distribution $\pi$ from which it is easy to sample, e.g., a standard Gaussian. 

Under mild conditions, the time-reversed process $(\bar{\theta}_{t})_{t\in[0,T]}:=(\theta_{T-t})_{t\in[0,T]}$ is also a diffusion process \citep{Anderson1982,Follmer1985,Haussmann1986}. Initialised at $\bar{\theta}_{0}\sim p_T(\cdot|x)$, this process evolves according to
\begin{align}
\mathrm{d}{\bar{\theta}}_{t} &= \left[-f(\bar{\theta}_{t},T-t) + g^2(T-t) \nabla_{\theta} \log p_{T-t}(\bar{\theta}_{t}|x)\right] \mathrm{d}t \nonumber \\
&\hspace{45mm}+g(T-t)\mathrm{d}{w}_{t},  \label{eq:backwardSDE}
\end{align}
where $p_{t}(\cdot|x)= \int p_{t|0}(\cdot|\theta_0)p(\theta_0|x)\mathrm{d}\theta_0$ denotes the time marginal density of  $\theta_{t}$, conditioned on $x$. By definition, the marginals  of $(\bar{\theta}_{t})_{t\in[0,T]}|x$ are equal to those of $(\theta_{T-t})_{t\in[0,T]}|x$. Thus, in particular, $\bar{\theta}_{T} \sim p_0(\cdot|x):=p(\cdot|x)$. 
Hence, if we could sample $\bar{\theta}_0\sim p_{T}(\cdot|x)$, and simulate $(\bar{\theta}_{t})_{t\in[0,T]}$ according to \eqref{eq:backwardSDE}, then its final distribution would be the desired posterior distribution. This process is visualised in \cref{fig1}.

Although this procedure provides an elegant sampling mechanism, it does not allow us to evaluate the density $p_0(\theta|x):=p(\theta|x)$ of these samples. Fortunately, there exists an ODE with the same marginals as \eqref{eq:forwardSDE}, which does enable density evaluation. This deterministic process, known as the {probability flow ODE} \citep{Song2021}, defines $(\theta_t)_{t\in[0,T]}$ according to
\begin{equation}
\frac{\mathrm{d}\theta_{t}}{\mathrm{d}t} = \left[ f(\theta_{t},t) - \frac{1}{2}g^2(t)\nabla_{\theta}\log p_{t}(\theta_{t}|x)\right], \label{eq:backwardODE}
\end{equation}
where once again $\smash{\theta_0 \sim p(\cdot|x)}$. In this case, the log densities $\smash{\log p_{t}(\theta_{t}|x)}$ \emph{can} be computed exactly via the {instantaneous change-of-variables} formula \cite{Chen2018b}:
\begin{align} \label{eq:change_of_var}
    &\frac{\mathrm{d} \log p_{t}(\theta_{t}|x)}{\mathrm{d}t} \\
    &= - \mathrm{Tr}\big[\nabla_{\theta}\big( f(\theta_{t},t) - \frac{1}{2}g^2(t)\nabla_{\theta}\log p_{t}(\theta_{t}|x)\big)\big]. \nonumber
\end{align}
In practice, we cannot simulate \eqref{eq:backwardSDE} or \eqref{eq:backwardODE} directly, since we do not have access to $p_T(\cdot|x)$, or the scores $\nabla_{\theta}\log p_t(\theta_t|x)$. We will therefore rely on two approximations. First, we will assume that $p_{T}\approx \pi$. Second, we will approximate $\nabla_{\theta}\log p_t(\theta_t|x)$ using score matching \citep[e.g.,][]{Song2021}, and substitute this approximation into \eqref{eq:backwardSDE} or \eqref{eq:backwardODE}. In this case, the ODE in \eqref{eq:backwardODE} is an instance of a continuous normalising flow (CNF) \cite{Grathwohl2019}. 

There are various ways in which we can obtain this approximation. Here, we choose to train a time-varying score network $s_{\psi}(\theta_t,x,t)\approx \nabla_{\theta}\log p_t(\theta_t|x)$ to directly approximate the score of the perturbed posterior \citep{Dhariwal2021,Song2021,Batzolis2021}.\footnote{In Appendix \ref{sec:nlse}, we outline an alternative approach which instead trains a score-network to approximate the score of the perturbed likelihood $\nabla_{\theta}\log p_t(x|\theta_t)$. We refer to this approach as Neural Likelihood Score Estimation (NLSE).} In this case, a natural objective is the weighted Fisher divergence 
\begin{align}
&\mathcal{J}_{\mathrm{post}}^{\mathrm{SM}}(\psi) =\frac{1}{2}\int_0^T\lambda_t
\label{eq:weightedposteriorSM}  \\ 
 &\hspace{10mm}\mathbb{E}_{p_t(\theta_t,x)} \left[ ||s_{\psi}(\theta_t,x,t) -  \nabla_{\theta} \log p_t(\theta_t|x) ||^2\right] \mathrm{d}t, \nonumber 
\end{align}
where $\lambda_t:[0,T]\rightarrow\mathbb{R}_{+}$ is a positive weighting function, and $p_{t}(\theta_t,x)$ denotes the joint distribution of $(\theta_t,x)$. In practice, this objective cannot be evaluated directly, since it depends on the posterior scores $\nabla_{\theta}\log p_t(\theta_t|x)$. Fortunately, one can show (e.g., \citealp{Batzolis2021,Tashiro2021}; \cref{ap:npse-1}) that it is equivalent to minimise the conditional denoising posterior score matching objective, given by  
\begin{align}
&\mathcal{J}^{\mathrm{DSM}}_{\mathrm{post}}(\psi)
= \frac{1}{2}\int_{0}^T\lambda_t \mathbb{E}_{p_{t|0}({\theta}_t|\theta_0) p(x|\theta_0) p(\theta_0)}\label{eq:weightedposteriorDSM} \\ &\hspace{20mm}  \big[ || s_{{\psi}}({\theta}_t,x,t) - \nabla_{{\theta}_t} \log p_{t|0}({\theta}_t|\theta_0)||^2\big]\mathrm{d}t , \nonumber 
\end{align}
where $p_{t|0}(\theta_t|\theta_0)$ denotes the transition density defined by \eqref{eq:forwardSDE}. In particular, this objective is minimised when $\smash{s_{\psi}(\theta_t,x,t) = \nabla_{\theta}\log p_t(\theta_t|x)}$ for almost all $\theta_t\in\mathbb{R}^d$, $x\in\mathbb{R}^p$, and $t\in[0,T]$. 

The expectation in \eqref{eq:weightedposteriorDSM} only depends on samples $\theta_0\sim p(\theta)$ from the prior, $x\sim p(x|\theta_0)$ from the simulator, and $\theta_t\sim p_{t|0}(\theta_t|\theta_0)$ from the forward diffusion \eqref{eq:forwardSDE}. Moreover, given a suitable choice for the drift and diffusion coefficients in \eqref{eq:forwardSDE}, the scores $\nabla_{\theta_t}\log p_{t|0}(\theta_t|\theta_0)$ can be computed in closed form. We can thus compute a Monte Carlo estimate of \eqref{eq:weightedposteriorDSM}, and minimise this to obtain $s_{\psi}(\theta_t,x,t)\approx \nabla_{\theta} \log p_t(\theta_t|x)$. 

We now have all of the necessary ingredients to generate approximate samples from the target posterior distribution:
\begin{itemize}[leftmargin=*]
\item[(i)] Draw samples $\theta_0\sim p(\theta)$ from the prior, $x\sim p(x|\theta_0)$ from the likelihood, and $\theta_t\sim p_{t|0}(\theta_t|\theta_0)$ using the forward process \eqref{eq:forwardSDE}.  
\item[(ii)] Using these samples, train a time-varying score network $s_{\psi}(\theta_t,x,t)\approx \nabla_{\theta}\log p_t(\theta_t|x)$ by minimising a Monte Carlo estimate of  \eqref{eq:weightedposteriorDSM}.
\item[(iii)] Draw samples $\smash{\bar{\theta}_0\sim \pi(\cdot)}$. Simulate an approximation of the reverse-time process in \eqref{eq:backwardSDE}, or the time-reversal of the probability flow ODE in \eqref{eq:backwardODE}, 
with $x=x_{\mathrm{obs}}$, replacing $\nabla_{\theta}\log p_{t}(\theta_t|x_{\mathrm{obs}})\approx s_{\psi}(\theta_t,x_{\mathrm{obs}},t)$. 
\end{itemize}

In line with the current SBI taxonomy, we will refer to this approach as Neural Posterior Score Estimation (NPSE).

In \cref{ap:npse-error-bounds}, we provide error bounds for NPSE in the fully deterministic sampling regime, assuming an $L^2$ bound on the approximation error and a mild regularity condition on the target posterior $p(\cdot|x_{\mathrm{obs}})$. Our result is adapted from \citet[Theorem 6]{Benton2024}.

\section{Sequential Neural Score Estimation}
\label{sec:sequential}
Given enough data and a sufficiently flexible model, the optimal score network $s_{\psi^{*}}(\theta_t,x,t)$ will equal $\nabla_{\theta}\log p_t(\theta_t|x)$ for almost all $x\in\mathbb{R}^p$, $\theta_t\in\mathbb{R}^d$, and $t\in[0,T]$. Thus, in theory, we can use the methods in the previous section to generate samples $\theta\sim p(\theta|x)$ for any observation $x$. 

In practice, we are often only interested in sampling from the posterior for a particular experimental observation $x=x_{\mathrm{obs}}$. Thus, given a finite simulation budget, it may be more efficient to train the score network using simulated data which is close to $x_{\mathrm{obs}}$, and thus more informative for learning the posterior scores $\nabla_{\theta} \log p_t(\theta_t|x_{\mathrm{obs}})$. This can be achieved by drawing initial parameter samples from a suitably chosen proposal prior, $\theta_0 \sim \tilde{p}(\theta)$, rather than the true prior $\theta_0 \sim p(\theta)$. This idea is central to existing sequential SBI algorithms, which use a sequence of adaptively chosen proposals in order to guide simulations towards more informative regions. The central challenge associated with developing a successful sequential algorithm is how to effectively correct for the mismatch between the so-called {proposal posterior} 
\vspace{-0.3em}
\begin{equation}
    \tilde{p}(\theta|x) = p(\theta|x) \frac{\tilde{p}(\theta)}{p(\theta)}\frac{{p}(x)}{\tilde{p}(x)}, \label{eq:snpe_identity}
\vspace{-0.3em}
\end{equation} 
and the true posterior $p(\theta|x)\propto p(\theta)p(x|\theta)$. In the following sections, we introduce several possible sequential variants of NPSE, which we collectively refer to as SNPSE. We note, as pointed out in the introduction, that in principle these approaches could also be used to develop sequential variants of the recently proposed flow-matching posterior estimation (FMPE) algorithm \citep{Dax2023}.

We begin by outlining some generic features of the sequential procedure, which hold irrespective of the specific sequential method employed (see Sections \ref{sec:truncated} - \ref{sec:alternatives}). In all cases, the sequential procedure will take place over $R$ rounds, indexed by $r\geq 1$. Given a total budget of $N$ simulations, we assume the simulations are evenly distributed across rounds: $N_r = N/R = M$ for $r=1,\dots,R$, where $N_r$ is the number of simulations in round $r$. In the first round, we follow the standard NPSE algorithm (Section \ref{sec:score-matching}). In particular, we first generate $\smash{\{\theta_{0,i}^{1}\}_{i=1}^{M}\sim p(\theta)}$ from the prior, and $\smash{\{x_i^{1}\}_{i=1}^{M}\sim p(x|\theta_{0,i})}$ using the simulator. These samples are used to train a score network $s_\psi(\theta_t, x, t) \approx \nabla_{\theta}\log p_t(\theta_t|x)$ by minimising \eqref{eq:weightedposteriorDSM}. By substituting this into \eqref{eq:backwardSDE}, we can generate samples approximately from the target posterior.
Following the initial round, there are several conceivable sequential procedures one could use to generate samples from $p(\theta|x_{\mathrm{obs}})$. We now describe several such methods. Broadly speaking, these procedures differ in (i) how they define the proposal prior; and (ii) how they correct for the mismatch between the proposal posterior and the true posterior. 

\subsection{Truncated Approach}
\label{sec:truncated}
We first introduce our preferred method: Truncated SNPSE (TSNPSE). This algorithm - summarised in Algorithm \ref{alg:tsnpse} - utilises a series of proposals given by truncated versions of the prior, inspired by the approach 
in \citet{Deistler2022}. For $r\geq 1$, let $\smash{{p}_{\psi}^{r-1}(\theta|x_{\mathrm{obs}})}$ denote the approximation to the target posterior learned in the $\smash{(r-1)^{\mathrm{th}}}$ round, with the convention that $\smash{p_{\psi}^{0}(\theta):= p(\theta)}$. Then, in the $r^{\mathrm{th}}$ round, we will use the highest-probability region of this approximation to define a truncated version of the prior.
To be precise, in the $r^{\mathrm{th}}$ round, suppose we define
\begin{equation}
    \bar{p}^{r}(\theta) \propto p(\theta) \cdot \mathbb{I}\{\theta\in\mathrm{HPR}_\varepsilon({p}_{\psi}^{r-1}(\theta|x_\mathrm{obs}))\}, \label{eq:truncation}
\end{equation}
where $\mathrm{HPR}_{\varepsilon}(\cdot)$ denotes the highest $1-\varepsilon$ probability region, defined as the smallest region which contains $1-\varepsilon$ of the mass; and we adopt the convention that $\bar{p}^{0}(\theta) = p(\theta)$. We then define the proposal distribution for this round as $\tilde{p}^{r}(\theta) = \frac{1}{r}\sum_{s=0}^{r-1}\bar{p}^{s}(\theta)$. Additional details regarding how to compute and sample from this proposal distribution are provided in Appendix \ref{sec:implementation}.

Crucially, under the assumption that we do not truncate regions which have non-zero mass under the true posterior $p(\theta|x_{\mathrm{obs}})$, this proposal distribution is proportional to the prior within the support of the posterior. Thus, we do not need to perform a correction. In particular, our loss function remains minimised at the score of the target posterior. This statement is formalised in the following proposition.

\begin{proposition}
\label{prop:main-result}
    Let $\tilde{p}^{r}(\theta) = \frac{1}{r}\sum_{s=0}^{r-1} \bar{p}^{s}(\theta)$, where $\bar{p}^{0}(\theta) = p(\theta)$ and $\bar{p}^{s}(\theta)$ is defined by \eqref{eq:truncation} for all $s\geq 1$. Suppose that $\Theta_{\mathrm{obs}} \subseteq \mathrm{HPR}_\epsilon({p}_{\psi}^{s}(\theta|x_\mathrm{obs}))$ for all $s\geq 1$, where $\Theta_{\mathrm{obs}} = \mathrm{supp}(p(\cdot|x_{\mathrm{obs}}))$. 
    Then, writing $\tilde{p}_t^{r}(\theta_t, x)$ for the distribution of $(\theta_t, x)$ when $(\theta_0,x) \sim \tilde{p}^{r}(\theta,x)$, the minimiser $\psi^{*}$ of the loss function
    \begin{align}
    \label{eq:tsnpse-loss}
&\hspace{-4mm}\mathcal{J}^{\mathrm{TSNPSE-SM}}_{\mathrm{post}}(\psi) 
    = \frac{1}{2} \int_{0}^{T} \lambda_t \mathbb{E}_{\tilde{p}_t^{r}(\theta_t, x)} \\
    &\hspace{20mm}[|| s_{\psi}(\theta_t, x, t) - \nabla_{\theta} \log p_t(\theta_t|x) ||^2] \mathrm{d}t, \nonumber \\
\intertext{or, equivalently, of the loss function}
    \label{eq:tsnpse-loss-denoising}
&\hspace{-4mm}\mathcal{J}^{\mathrm{TSNPSE-DSM}}_{\mathrm{post}}(\psi) 
    = \frac{1}{2} \int_{0}^{T} \lambda_t \mathbb{E}_{p_{t|0}(\theta_t|\theta_0)p(x|\theta_0)\tilde{p}^r(\theta_0)} \\
    &\hspace{20mm}[|| s_{\psi}(\theta_t, x, t) - \nabla_{\theta} \log p_{t
0}(\theta_t|\theta_0) ||^2] \mathrm{d}t, \nonumber
\end{align}
satisfies $s_{\psi^{\star}}(\theta_t, x_\mathrm{obs}, t) = \nabla_{\theta} \log p_t(\theta_t|x_{\mathrm{obs}})$.
\end{proposition}
\vspace{-3mm}
\begin{proof}
See Appendix \ref{sec:tnspse}.
\end{proof}

\vspace{-3mm}
\begin{algorithm}[H]
\caption{TSNPSE}
\label{alg:tsnpse}
\begin{algorithmic}
    \STATE \textbf{Inputs:} Observation $x_\mathrm{obs}$, prior $p(\theta) =: \bar{p}^{0}(\theta)$, simulator $p(x|\theta)$, simulation budget $N$, number of rounds $R$, (simulations-per-round $M=N/R$), dataset $\mathcal{D} = \{\}$.
    \STATE \textbf{Outputs: $p_\psi(\theta|x_{\mathrm{obs}}) \approx  p(\theta|x_{\mathrm{obs}}).$} 
    \FOR{$r = 1,\dots,R$}
        \FOR{$i = 1,\dots,M$}
            \STATE Draw $\theta_i \sim \bar{p}^{r-1}(\theta)$, $x_i \sim p(x|\theta_{i})$.
            \STATE Add $(\theta_i, x_i)$ to $\mathcal{D}$.
        \ENDFOR
        \STATE Learn $s_{\psi}(\theta_t,x,t) \approx \nabla_{\theta} \log p_t(\theta_t|x)$ by minimising 
        a Monte Carlo estimate of \eqref{eq:tsnpse-loss-denoising} based on dataset $\mathcal{D}$. \\[.5mm]
        \STATE Compute $\bar{p}^{r}(\theta)$ in \eqref{eq:truncation} using $s_{\psi}(\theta_t,x_{\mathrm{obs}},t)$. See Appendix \ref{sec:implementation} for details.
    \ENDFOR
    \STATE Get $p_\psi(\theta|x_{\mathrm{obs}})$ sampler by substituting $s_{\psi}(\theta_t,x_{\mathrm{obs}},t)\approx\nabla_{\theta} \log p_t(\theta_t|x_{\mathrm{obs}}) $ in \eqref{eq:backwardODE}.
    \STATE \textbf{Return:} $p_\psi(\theta|x_{\mathrm{obs}})$.
\end{algorithmic}
\end{algorithm}

\subsection{Alternative Approaches}
\label{sec:alternatives}
We now outline several other possible sequential approaches for NPSE. An extensive and detailed discussion of these methods, as well as supporting numerical results, can be found in Appendix \ref{sec:sequential-additional}. Broadly speaking, these methods can be viewed as score-based analogues of existing sequential variants of NPE, namely, SNPE-A \cite{Papamakarios2016}, SNPE-B \cite{Lueckmann2017}, and SNPE-C \cite{Greenberg2019}. We refer to, e.g., \citet{Durkan2020} for a concise overview of SNPE-A, SNPE-B, and SNPE-C.

Unlike TSNPSE, in each of these methods, the proposal prior is defined \emph{directly} in terms of the most recent approximation of the posterior. In particular, in the $r^{\mathrm{th}}$ round, we now sample new parameters $\smash{\{\theta_{0,i}^{r}\}_{i=1}^{M} \sim {p}_{\psi}^{r-1}(\theta|x_{\mathrm{obs}})}$ and simulate new data $\smash{\{x_i^{r}\}_{i=1}^{M} \sim p(x|\theta_{0,i}^{r})}$. We then concatenate these samples with those from previous rounds to form $\bigcup_{s=1}^{r}\{(\theta_{0,i}^{s}, x_i^{s})\}_{i=1}^{M} \sim \tilde{p}^{r}(\theta)p(x|\theta)$, where $\tilde{p}^{r}(\theta) = \frac{1}{r}\sum_{s=0}^{r-1} p_{\psi}^{s}(\theta|x_\mathrm{obs})$, and $p_{\psi}^{0}(\theta|x_\mathrm{obs}) := p(\theta)$. 

In this case, if were to minimise the original score matching objective \eqref{eq:weightedposteriorDSM}, but using samples $\theta_0\sim \tilde{p}^{r}(\theta)$ rather than $\theta_0\sim p(\theta)$, we would learn a score network which approximates $\nabla_{\theta}\log \tilde{p}^{r}_t(\theta_t|x)$, rather than $\nabla_{\theta} \log p_t(\theta_t|x)$, where $\tilde{p}^{r}_t(\theta_t|x) = \int_{\mathbb{R}^d} p_{t|0}(\theta_t|\theta_0)\tilde{p}^{r}(\theta_0|x)\mathrm{d}\theta_0$, and ${\tilde{p}^{r}(\theta|x) = \frac{\tilde{p}^{r}(\theta)p(x|\theta)}{\tilde{p}^{r}(x)}}$. Substituting this score network, evaluated at $x=x_{\mathrm{obs}}$,
into  \eqref{eq:backwardSDE} or \eqref{eq:backwardODE}, would then result in samples $\theta\sim \tilde{p}^{r}(\theta|x_{\mathrm{obs}})$, rather than $\theta\sim p(\theta|x_{\mathrm{obs}})$. We thus require a correction to recover samples from the correct posterior.

\paragraph{SNPSE-A.} The first approach is to perform a {post-hoc} importance weight correction using, e.g., sampling-importance resampling (SIR) \cite{Rubin1987,Rubin1988,Smith1992,Gelman1995}. According to this approach, we first generate $\smash{\{{\tilde{\theta}}_i\}_{i=1}^{M'}\sim {\tilde{p}}_{\psi}^{r}({\cdot}|x_{\mathrm{obs}})}$, where $\smash{{\tilde{p}}_{\psi}^{r}(\cdot|x_{\mathrm{obs}})}$ denotes the approximate proposal posterior obtained in the $r^{\mathrm{th}}$ round, and $M'\geq M$. We then draw samples $\{\theta_i\}_{i=1}^M$ with or without replacement from $\smash{\{\tilde{\theta}_i\}_{i=1}^{M'}}$, with sample probabilities, $\tilde{w}_i$, proportional to the importance ratios 
\begin{equation}
\label{eq:sir}
    \tilde{h}_i = \frac{p(\tilde{\theta}_i|x_{\mathrm{obs}})}{{\tilde{p}}_{\psi}^{r}(\tilde{\theta}_i|x_{\mathrm{obs}})}.
\end{equation}
In the limit as $M'\rightarrow\infty$, this sample will consist of independent draws from $p(\cdot|x_{\mathrm{obs}})$ \citep[e.g.,][]{Smith1992}. In practice, we cannot evaluate $p(\cdot|x_{\mathrm{obs}})$ in \eqref{eq:sir}, and thus will instead use sample probabilities $w_i$ proportional to
\begin{equation}
\label{eq:sir-approx}
    {h}_i = \frac{p(\tilde{\theta}_i)}{\tilde{p}^{r}(\tilde{\theta}_i)}.
\end{equation}
The importance ratios in \eqref{eq:sir-approx} are approximately proportional to the correct importance ratios in \eqref{eq:sir}, since 
\begin{equation}
\label{eq:importance-sampling}
    h_i = \frac{p(\tilde{\theta}_i)}{\tilde{p}^{r}(\tilde{\theta}_i)}\propto \frac{p(\tilde{\theta}_i|x_{\mathrm{obs}})}{\tilde{p}^{r}(\tilde{\theta}_i|x_{\mathrm{obs}})} \approx \frac{p(\tilde{\theta}_i|x_{\mathrm{obs}})}{{\tilde{p}}_{\psi}^{r}(\tilde{\theta}_i|x_{\mathrm{obs}})} = \tilde{h}_i.
\end{equation}
Although SNPSE-A can work well in simple settings, it is fundamentally limited by the approximation introduced in \eqref{eq:importance-sampling}. In particular, when there is a significant mismatch between the true proposal, $\tilde{p}^{r}(\cdot|x_{\mathrm{obs}})$, and the approximate (learned) proposal, ${\tilde{p}}_{\psi}^{r}(\cdot|x_{\mathrm{obs}})$, this approach can lead to inaccurate inference (see Appendix \ref{sec:snpse-a}).

\paragraph{SNPSE-B.} The second approach is to include an importance weight correction within the denoising score matching objective \eqref{eq:weightedposteriorDSM}. In particular, in the $r^{\mathrm{th}}$ round, we now minimise a Monte Carlo estimate of  
\begin{align}
&\mathcal{J}^{\mathrm{SNPSE-B}}_{\mathrm{post}}(\psi)
= \frac{1}{2}\int_{0}^T\lambda_t \mathbb{E}_{p_{t|0}({\theta}_t|\theta_0) p(x|\theta_0) \tilde{p}^{r}(\theta_0)}\label{eq:weightedposteriorDSM-IW} \\ &\hspace{10mm}  \bigg[ \frac{p(\theta_0)}{\tilde{p}^{r}(\theta_0)}|| s_{{\psi}}({\theta}_t,x,t) - \nabla_{{\theta}_t} \log p_{t|0}({\theta}_t|\theta_0)||^2\bigg]\mathrm{d}t. \nonumber 
\end{align}
It is straightforward to show that this objective is minimised at the score of the true posterior, that is, by $\psi^{*}$ such that $s_{{\psi}^{*}}({\theta}_t,x,t) = \nabla_{\theta}\log p_t(\theta_t|x)$ (see Appendix \ref{sec:snpse-b}). Unfortunately, similar to SNPE-B \cite{Lueckmann2017}, the importance weights are often high variance, resulting in unstable training and poor overall algorithm performance \citep[e.g.,][]{Papamakarios2019,Durkan2019}.

\paragraph{SNPSE-C.} The third approach is to include a score-based correction within the denoising posterior score matching objective \eqref{eq:weightedposteriorDSM}. In this case, we minimise \eqref{eq:weightedposteriorDSM}, now over samples from the proposal prior, to learn an estimate $\smash{\tilde{s}^{r}_{\psi}(\theta_t,x,t)\approx \nabla_{\theta} \log \tilde{p}_t^{r}(\theta_t|x)}$ of the proposal posterior. We would like to use this to automatically recover an estimate of $\smash{\nabla_{\theta}\log p_t(\theta_t|x)}$. To do so, observe that 
\begin{align}
   \hspace{-2mm} \nabla_{\theta}\log p_t(\theta_t|x) &= \nabla_{\theta}\log p_t(\theta_t)+ \nabla_{\theta}\log  p_t(x|\theta_t)\\
   \hspace{-2mm} \nabla_{\theta}\log \tilde{p}_t^{r}(\theta_t|x) &= \nabla_{\theta}\log 
\tilde{p}_t^{r}(\theta_t) + \nabla_{\theta}\log \tilde{p}_t^{r}(x|\theta_t)
\end{align}
where $\smash{p_t(x|\theta_t) = \int p(x|\theta_0) p_{0|t}(\theta_0|\theta_t)\mathrm{d}\theta_0}$ and $\tilde{p}_t^{r}(x|\theta_t) = \int p(x|\theta_0) \tilde{p}^{r}_{0|t}(\theta_0|\theta_t)\mathrm{d}\theta_0$. Thus, in particular, 
\begin{align}
    \nabla_{\theta}\log \tilde{p}_t^{r}(\theta_t|x)&=\nabla_{\theta}\log p_t(\theta_t|x) \label{eq:bayes-perturbed-score} \\
    &+ \nabla_{\theta}\log 
 \tilde{p}^{r}_t(\theta_t) + \nabla_{\theta}\log \tilde{p}_t^{r}(x|\theta_t) \nonumber \\
 &- \nabla_{\theta}\log p_t(\theta_t)- \nabla_{\theta}\log  p_t(x|\theta_t). \nonumber 
\end{align}
This identity suggests defining $\smash{\tilde{s}^{r}_{\psi}(\theta_t,x,t)}$ in terms of another score network $\smash{s_{\psi}(\theta_t,x,t)}$ according to
\begin{align}
    \tilde{s}^{r}_{\psi}(\theta_t,x,t)&=s_{\psi}(\theta_t,x,t) \label{eq:new-score-est} \\
    &+ \nabla_{\theta}\log 
 \tilde{p}^{r}_t(\theta_t) + \nabla_{\theta}\log \tilde{p}_t^{r}(x|\theta_t) \nonumber \\
 &- \nabla_{\theta}\log p_t(\theta_t)- \nabla_{\theta}\log  p_t(x|\theta_t). \nonumber
\end{align}
In this case, given $\smash{\tilde{s}^{r}_{\psi}(\theta_t,x_{\mathrm{obs}},t) \approx \nabla_{\theta} \log \tilde{p}^{r}_{t}(\theta_t|x_{\mathrm{obs}})}$, we also have $\smash{s_{\psi}(\theta_t,x_{\mathrm{obs}},t) \approx \nabla_{\theta} \log p_t(\theta_t|x_{\mathrm{obs}})}$ from \eqref{eq:bayes-perturbed-score} - \eqref{eq:new-score-est}, as required. Unlike SNPSE-A and SNPSE-B, SNPSE-C has the advantage of not requiring importance weights. Moreover, since the corrections are performed `in the score space', it does not require us to evaluate $\smash{\tilde{p}^{r}(\cdot)}$, and thus does not necessitate calculating likelihoods via \eqref{eq:backwardODE} and \eqref{eq:change_of_var}. On the other hand, it does require knowledge of $\smash{\nabla_{\theta} \log \tilde{p}^{r}_t(\theta_t)}$, $\smash{\nabla_{\theta} \log \tilde{p}^{r}_t(x|\theta_t)}$, $\smash{\nabla_{\theta} \log p_t(\theta_t)}$, and $\smash{\nabla_{\theta} \log p_t(x|\theta_t)}$, which are not immediately available. Thus, in practice, this approach depends on several additional approximations. We provide further details in Appendix \ref{sec:snpse-c}.

In empirical testing, the corrections required for SNPSE-A, SNPSE-B, and SNPSE-C, lead to significantly worse performance than TSNPSE (see Appendix \ref{sec:sequential-additional}). On this basis, we advocate for TSNPSE as the preferred sequential method, and focus exclusively on this approach in our subsequent numerics (see Section \ref{sec:numerics}).

\section{Related Work}
\label{sec:related_work}

\subsection{Simulation-Based Inference}

\paragraph{Approximating the Posterior.} 
Many modern SBI algorithms are based on learning a conditional neural density estimator $q_{\psi}(\theta|x)$ to approximate the posterior $p(\theta|x)$, often over a number of rounds of training \cite{Papamakarios2016,Lueckmann2017,Greenberg2019}. This approach is known as SNPE. Such methods circumvent the bias introduced by the use of a proposal prior in various ways, including a post-hoc importance weight correction (SNPE-A) \cite{Papamakarios2016}, minimising an importance weighted loss function (SNPE-B) \cite{Lueckmann2017}, and re-parametrising the proposal posterior objective (SNPE-C) \cite{Greenberg2019}. Alternatively, the use of a truncated prior as the proposal circumvents the need for a correction (TNPSE) \cite{Deistler2022}. As noted in Section \ref{sec:sequential}, our sequential methods can loosely be viewed as analogues of these approaches suitable for diffusion models.  

\paragraph{Approximating the Likelihood.}
Rather than approximating the posterior directly, another approach is to learn a model $q_{\psi}(x|\theta)$ for the intractable likelihood $p(x|\theta)$. Such methods are sometimes referred to as Synthetic Likelihood approaches \cite{Wood2010,Ong2018,Price2018,Frazier2022}. Early examples of this approach assume that the likelihood can be parameterised as a single Gaussian \cite{Wood2010}, or a mixture of Gaussians \cite{Fan2013}. More recent approaches, referred to as SNLE, train conditional neural density estimators, over a number of rounds \cite{Lueckmann2019, Papamakarios2019}. While SNLE does not require a correction, it does rely on MCMC to generate posterior samples. This can be costly, and may prove prohibitive for posteriors with complex geometries.

\paragraph{Approximating the Likelihood Ratio.} 
Another approach to simulation-based inference is based on learning a parametric model for the likelihood-to-marginal ratio $r(x,\theta) = {p(x|\theta)}/{p(x)} = {p(\theta|x)}/{p(\theta)}$  \cite{Izbicki2014,Tran2017,Durkan2020,Hermans2020,Miller2021,Simons2021,Thomas2022}, or the likelihood ratio $r(x,\theta_1,\theta_2) = {p(x|\theta_1)}/{p(x|\theta_2)}$ \cite{Pham2014,Cranmer2016,Gutmann2018,Stoye2019,Brehmer2020}. In the first case, one trains a binary classifier to approximate this ratio.
Using the fact that $p(\theta|x_{\mathrm{obs}}) = p(\theta) r(x_{\mathrm{obs}},\theta)$, one can then use MCMC to generate posterior samples. This approach is also amenable to a sequential implementation, known as SNRE \cite{Durkan2020}. 

\paragraph{Approximating the Posterior and the Likelihood.} 
Two recent methods aim to combine the advantages of SNLE (or SNRE) and SNPE, while addressing their shortcomings \cite{Wiqvist2021,Glockler2022}. In particular, SNVI \cite{Glockler2022} and Sequential Neural Posterior and Likelihood Approximation (SNPLA) \cite{Wiqvist2021} first train a neural density estimator $q_{\psi_{\text{lik}}}(x|\theta)$ to approximate the likelihood, or the likelihood ratio. Once this model has been trained, one trains a parametric approximation $q_{\psi_{\text{post}}}(\theta)$ for the posterior, using variational inference with normalising flows. These methods differ in their variational objectives: SNVI uses the forward KL divergence, the importance weighted ELBO, or the Renyi $\alpha$-divergence, while SNPLA uses the reverse KL divergence.
 
\subsection{Diffusion Models}
Diffusion models \cite{Sohl-Dickstein2015,Song2019,Ho2020,Song2021} have recently emerged as a new class of generative models. These models offer high quality generation and sample diversity, do not require adversarial training, and have achieved state-of-the-art performance in a range of applications, including image generation \cite{Dhariwal2021,Ho2020,Song2021}, audio synthesis \cite{Chen2021a,Kong2021,Popov2021}, shape generation \cite{Cai2020}, music generation \cite{Mittal2021}, and video generation \citep{Ho2022a}.

Conditional diffusion models \cite{Song2019,Song2021,Batzolis2021,Dhariwal2021,Chao2022}
extend this framework to allow for conditional generation, allowing for tasks such as image in-painting \cite{Song2021}, time series imputation \cite{Tashiro2021}, image colourisation \cite{Song2021}, and medical image reconstruction \cite{Song2022}. In such applications, the `prior' typically corresponds to an unknown data distribution, whose score is estimated using score matching. 
Meanwhile, the `likelihood' is often known, or else corresponds to a differentiable classifier \citep[e.g.,][]{Song2021}. This is rather different to our setting, in which the prior is typically known, while the likelihood is intractable. 

\subsection{Diffusion Models and Simulation-Based Inference} \label{subsec:SM_and_LFI}
Surprisingly, the application of diffusion models to problems of interest to the SBI community \citep[see, e.g.,][]{Lueckmann2021} has not previously been investigated. In parallel with this work, \citet{Geffner2022} also considered the use of diffusion models for SBI. While related to our work, \citet{Geffner2022} focused specifically on how to use NPSE for sampling from $p(\theta|x_{\mathrm{obs}}^1,\dots,x_{\mathrm{obs}}^n)$, for any set of observations $\{x_{\mathrm{obs}}^1, \dots, x_{\mathrm{obs}}^n\}$. Meanwhile, we introduce sequential variant(s) of NPSE (see Section \ref{sec:sequential}). We provide a more detailed comparison with this work in Appendix \ref{app:LFI_multiple_obs}.

More recently, several other authors have proposed SBI algorithms which are closely related to diffusion models. In particular, \citet{Dax2023} propose {flow matching posterior estimation} (FMPE), an SBI algorithm which approximates the posterior $p(\theta|x)$ using a CNF trained via flow matching \cite{Lipman2023}. This approach includes NPSE, when using the deterministic probability-flow ODE, as a special case. Meanwhile, \citet{Schmitt2023} introduce {consistency model posterior estimation} (CMPE), which applies consistency models \cite{Song2023} to SBI. In contrast to our work, both of these papers consider only the amortised setting, and do not introduce sequential variants of their algorithms. 

\section{Numerical Experiments}
 In this section we benchmark the numerical performance of NPSE and TSNPSE. Code to reproduce our numerical results can be found at \url{https://github.com/jacksimons15327/snpse_icml}.

\label{sec:numerics}
\subsection{Experimental Details}
In all experiments, our score network is comprised of independent multilayer perceptron (MLP) embedding networks for $\theta_t$ and $x$. A sinusoidal embedding is employed for $t$. The embeddings of $\theta_t, x, t$ are concatenated and input to a MLP. All MLP networks have $3$ fully connected layers, each with $256$ neurons and SiLU activation functions. We use Adam \cite{Kingma2015} to train the networks, with a learning rate of $10^{-4}$. We hold back $15\%$ of the data to be used as a validation set for early stopping. We provide details of any additional hyperparameters in Appendix \ref{ap:hyperparameters}.

\subsection{Benchmark Results}
\label{subsec:results}

We first provide results for eight popular SBI benchmarks described in \citet{Lueckmann2021} (see Appendix \ref{sec:benchmarks} for details). We consider simulation budgets of $1000$, $10000$ and $100000$. In all cases, we report the classification-based two-sample test (C2ST) score \cite{Lopez-Paz2017}, which varies between $0.5$ and $1$ (lower is better), with a score of $0.5$ indicating perfect posterior estimation. 

\begin{figure*}[h!]
  \centering
  \includegraphics[width=\textwidth, trim=90 20 40 20, clip]{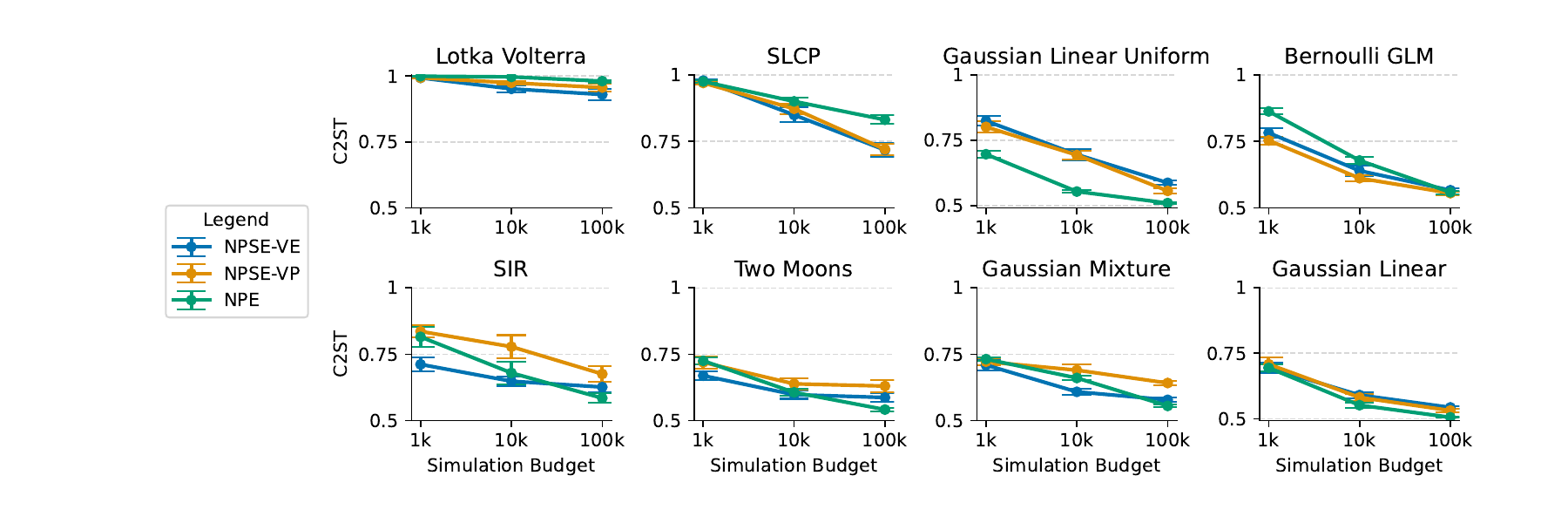}
  \vspace{-7mm}
  \caption{\textbf{Results on eight benchmark tasks (non-sequential methods).}}
  \label{fig:non-seq}
\end{figure*}

\begin{figure*}[h!]
\centering
  \includegraphics[width=\textwidth, trim=90 20 40 20, clip]{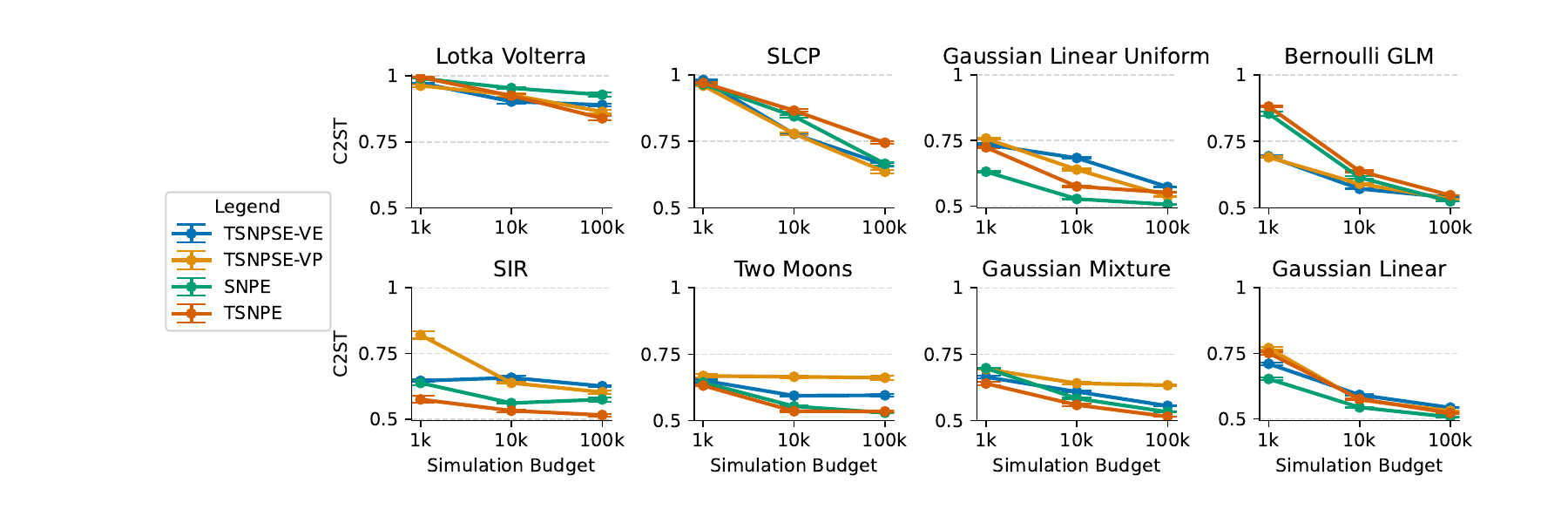}
  \vspace{-7mm}
  \caption{\textbf{Results on eight benchmark tasks (sequential methods).}}
  \label{fig:seq}
\vspace{-3mm}
\end{figure*}

For both our non-sequential (NPSE) and sequential (TSNPSE) methods, we consider two choices of dynamics for the forward noising process: a variance-exploding SDE (VE SDE) and a variance-preserving SDE (VP SDE) \citep{Song2021}. Further details can be found in Appendix \ref{ap:sde}. For reference, we compare our non-sequential method (NPSE) with NPE \cite{Papamakarios2016}; and our sequential method (TSNPSE) with SNPE-C \cite{Greenberg2019} and TSNPE \cite{Deistler2022}. For these algorithms, we obtain results using the python toolkit \texttt{sbibm} \cite{Lueckmann2021}.  We include an additional comparison with FMPE \cite{Dax2023} in Appendix \ref{app:fmpe}.

Our results, provided in Figures \ref{fig:non-seq} and \ref{fig:seq}, demonstrate that diffusion models provide an accurate and robust alternative to state-of-the-art SBI methods based on posterior density estimation with (discrete) normalising flows. Notably, for the two most challenging benchmark experiments, SLCP and Lotka Volterra, our methods outperform their competitors, providing evidence that our proposed algorithms scale well to high-dimensions. For the remaining benchmark experiments, the results are more mixed, with the best performing method varying based on the task at hand as well as the simulation budget. It is worth emphasising that our algorithms employ the same hyperparameter settings (e.g., neural network architecture, optimizer, etc.) across all experiments, including both the benchmarks and the real-world experiment in Section \ref{sec:pyloric}, and that we did not perform an extensive hyperparameter search. We suspect that the performance of (TS)NPSE could be further improved with additional tuning.

We also note that the choice of dynamics (e.g., VE SDE or VP SDE) for the forward noising process can have a significant impact on the quality of the posterior inference, although the best performing method can fluctuate based on the task at hand. In general, based on our empirical results, we recommend VE SDE for low dimensional experiments, and VP SDE for high dimensional experiments. 

\subsection{Real-world Neuroscience Problem}
\label{sec:pyloric}
We also apply TSNPSE to a challenging real-world neuroscience problem: inference for the parameters of a simulator model of the pyloric network of the stomatogastric ganglion in the crab \emph{Cancer Borealis} \citep{Prinz2003,Prinz2004}. In this case, the model simulates 3 neurons, whose behaviours are governed by synapses and membrane conductances which together constitute a set of 31 parameters. The simulator outputs 3 voltage traces, which are condensed into 18 summary statistics \citep{Prinz2003,Prinz2004}. The prior is uniform over previously defined parameter ranges \cite{Prinz2004,Goncalves2020}. We are interested in inferring the posterior distribution of the parameters, given experimentally observed data \cite{Haddad2021}. 

\begin{figure}
\vspace{-3mm}
  \centering

  \begin{subfigure}[b]{0.495\columnwidth}
    \includegraphics[width=\textwidth]{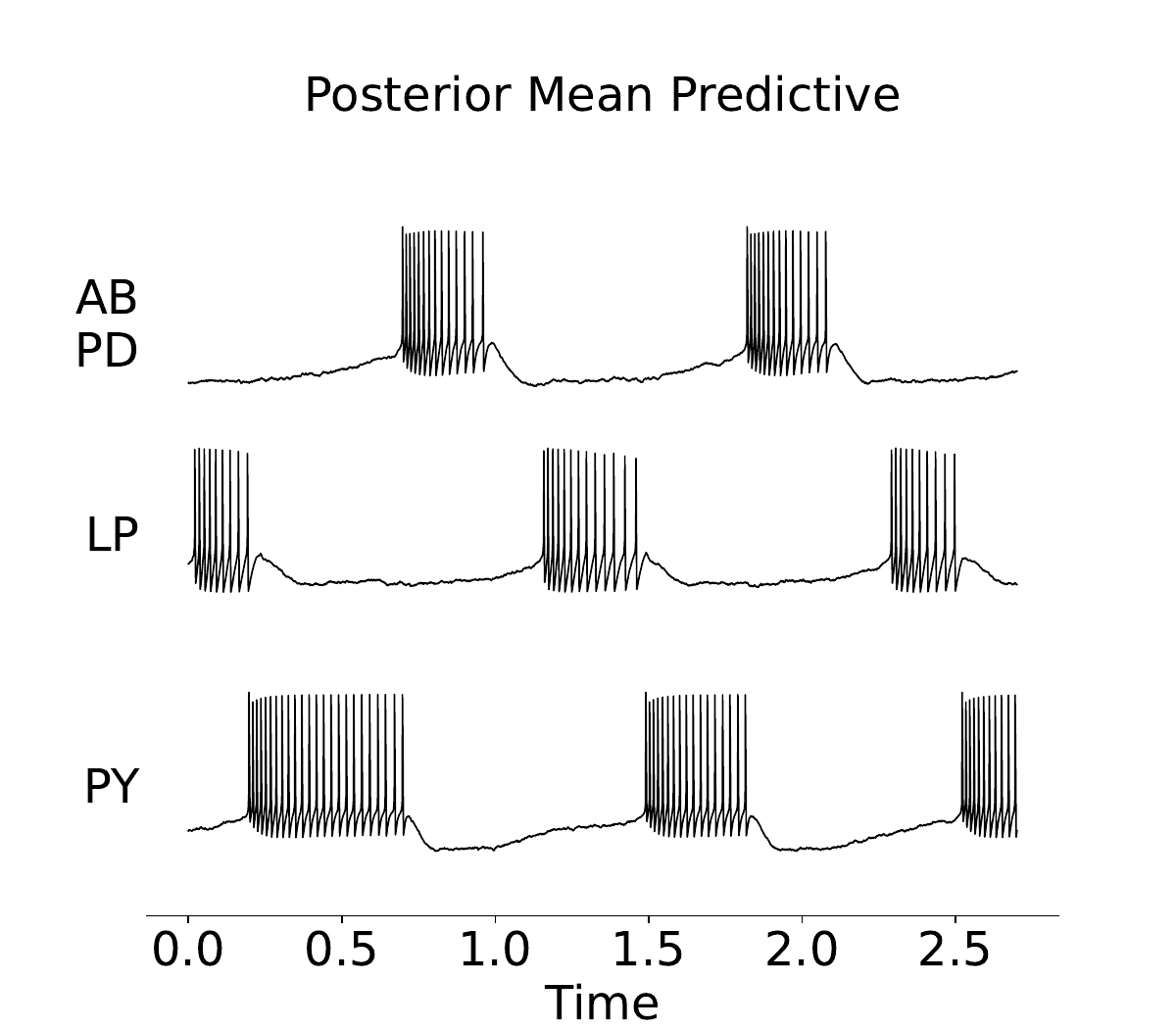}
    \caption{Posterior mean predictive.}
    \label{fig:pyloric-post-predictive}
  \end{subfigure}
  \hfill
  \begin{subfigure}[b]{0.495\columnwidth}
    \includegraphics[width=\textwidth]{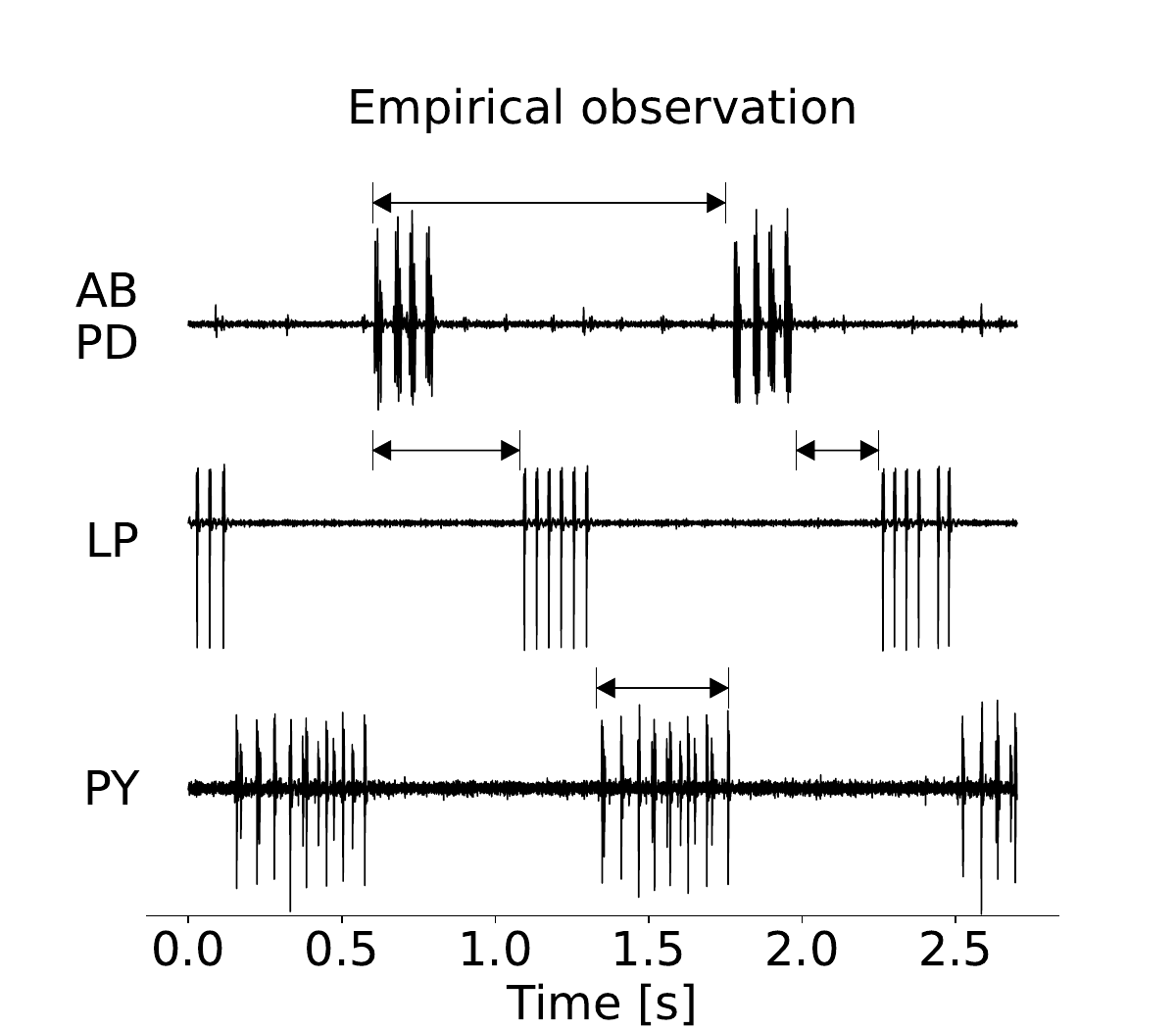}
    \caption{True observation.}
    \label{fig:pyloric-true-obs}
  \end{subfigure}

  \vspace{10pt} 

  \begin{subfigure}[b]{0.99\columnwidth}
    \includegraphics[width=\textwidth]{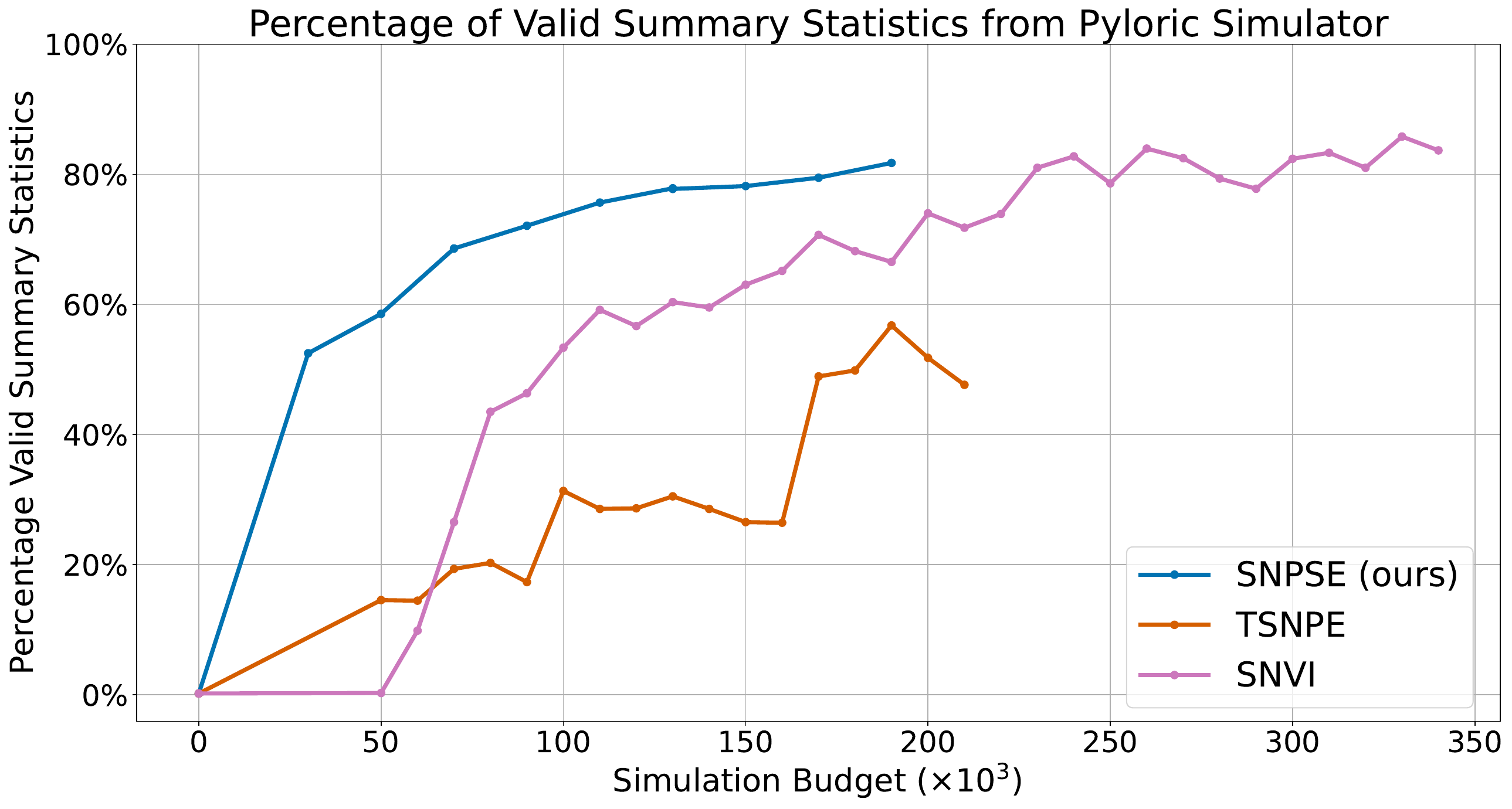}
    \caption{Percentage of valid summary statistics from the pyloric simulator against simulation budget for 3 different methods, SNPSE (ours, blue), TSNPE (orange), SNVI (pink).}
    \label{fig:pyloric-rej-rate}
  \end{subfigure}
  \vspace{-1mm}
  \caption{\textbf{Results for the Pyloric experiment.}}
  \label{fig:pyloric_plots}
  \vspace{-1mm}
\end{figure}

In this model, the volume of the parameter space which gives rise to meaningful summary statistics is very small. For example, over $99\%$ of prior samples input into the simulator result in neural traces with ill-defined summary statistics. This, alongside the significant simulator cost, renders posterior inference in this model a very challenging task. Previous work has performed amortised inference using NPE, although this requires several million simulations \cite{Goncalves2020,Deistler2022a}. More recent methods have adopted a sequential approach, reducing the number of samples required by 25 times or more \citep{Glockler2022,Deistler2022,Glaser2022}.

We applied TSNPSE to this problem, using an identical architecture to that used in our benchmark experiments to demonstrate the robustness of our approach. We performed inference over 9 rounds, with 30000 initial simulations, and 20000 added simulations in each round. Our results, including the percentage of valid summary statistics versus the number of simulations, and a posterior predictive sample, are provided in Figure \ref{fig:pyloric_plots}. We also provide a pairwise marginal plot of our final posterior approximation in Figure \ref{fig:pyloric-pairwise-marginal} (Appendix \ref{sec:pyloric-additional}). In the final round, we achieved 81\% valid summary statistics from the simulator (Figure \ref{fig:pyloric-rej-rate}), superior to the percentage achieved by other methods for the same simulation budget. We also note that the obtained posterior produces samples which closely match the observed data (Figure \ref{fig:pyloric-post-predictive}). In addition, the posterior marginals (Figure \ref{fig:pyloric-pairwise-marginal}) are very similar to others previously reported in the literature \citep{Deistler2022,Glockler2022}.

\section{Discussion}
\label{sec:discussion}

\paragraph{Limitations}
The main limitation of our approach relates to computational cost. In particular, TSNPSE requires computing $\mathrm{HPR}_{\varepsilon}$ of the approximate posterior to define the proposal, which involves computing the approximate posterior density over many samples. In TSNPE, which uses a normalising flow, this is relatively inexpensive as a single forward pass is required for sampling, and a single backward pass for density evaluation \cite{Dinh2017,Papamakarios2017}. In contrast, with TSNPSE, which uses a CNF, multiple forward passes are required for sampling, and multiple gradients of forward passes are required for density evaluation \cite{Chen2018a,Grathwohl2019}. Interestingly, this can be avoided using an alternative parameterisation of the score network; see \cref{app:alt-param} for further details.

It is worth noting that there are several ways  to reduce the cost of both sampling and likelihood evaluation in CNFs. For example, faster numerical ODE solvers can substantially reduce the number of forward passes required for sampling \citep[e.g.,][]{Lu2022,Zhang2023}. Meanwhile, the Skilling-Hutchinson trace estimator \cite{Skilling1989,Hutchinson1990} can be used to reduce the cost of the gradient computations required for likelihood evaluation \cite{Grathwohl2019}. 

In some sense, this comparison between TSNPE and TSNPSE reflects a wider discussion regarding the trade-offs between (discrete) normalising flows and CNFs. The former are associated with a lower computational cost, while the latter are much more flexible, which can result in more accurate inference \citep[e.g.,][]{Grathwohl2019,Finlay2020}. As such, preference for a method based on a discrete normalising flow (e.g., TSNPE) or a CNF (e.g., TSNPSE) will depend on the problem at hand. For example, for challenging real-world simulators, the additional cost incurred by a CNF may be negligible in comparison to the cost of acquiring simulations.

\paragraph{Future Work}
We highlight two directions for future work. First, with the exception of SNPSE-C, the sequential methods in this paper can also be applied to other methods based on CNFs. In this sense, a natural extension of our work would be to develop a sequential variant of FMPE \citep{Dax2023}. Second, in this paper we used a relatively simple neural network architecture, with a relatively small number of parameters, in large part to demonstrate the robustness of our approach. In contrast, the architectures used by diffusion models in other modalities are often highly specialised, and have received significant attention in their own right \citep[e.g.,][]{Karras2022}. Undoubtedly, further investigation into effective network design for SBI problems would be a fruitful direction for future work. 

\section*{Acknowledgements}
The authors are grateful to the three anonymous reviewers for their constructive feedback. The authors would also like to thank Tomas Geffner and Iain Murray for useful comments on early drafts of the paper. LS was supported by the UK Research and Innovation (UKRI) Engineering and Physical Sciences Research Council (EPSRC), grant number EP/V022636/1. JS was supported by the EPSRC Centre for Doctoral Training in Computational Statistics and Data Science, grant number EP/S023569/1.

\section*{Impact Statement}
This paper presents work whose goal is to advance the field of machine learning. There are several potential societal consequences of our work, none of which we feel must be specifically highlighted here.

\bibliography{references}
\bibliographystyle{icml2024}

\newpage
\appendix
\onecolumn

\section{Neural Posterior Score Estimation: Theoretical Results}
\label{ap:npse}

\subsection{Derivation of the NPSE Loss Function}
\label{ap:npse-1}
In this section we include a self-contained proof that the minimiser of the denoising posterior score matching objective in \eqref{eq:weightedposteriorDSM} is equal to the minimiser of the posterior score matching objective in \eqref{eq:weightedposteriorSM}. Similar results can also be found in \citet{Batzolis2021,Tashiro2021}. 

Our proof begins with the observation that 
\begin{align}
\mathcal{J}_{\mathrm{post}}^{\mathrm{DSM}}(\psi)
    &= \frac{1}{2} \int_0^T \lambda_t \mathbb{E}_{(\theta_t,x) \sim p_t(\theta_t,x)} \left[ ||s_{\psi}(\theta_t,x,t) - \nabla_{\theta}\log p_t(\theta_t|x)||^2 \right] \mathrm{d}t \\
    &= \frac{1}{2} \int_0^T \lambda_t \bigg[ \underbrace{\mathbb{E}_{(\theta_t,x) \sim p_t(\theta_t,x)} \big[ ||s_\psi(\theta_t,x,t)||^2 \big]}_{\Omega_t^{1}}-2 \underbrace{\mathbb{E}_{(\theta_t,x) \sim p_t(\theta_t,x)} [s^{\top}_{\psi}(\theta_t,x,t)\nabla_{\theta}\log p_t(\theta_t|x)]}_{\Omega_t^{2}}  \\
    &\hspace{15mm}+ \underbrace{\mathbb{E}_{(\theta_t,x) \sim p_t(\theta_t,x)} \big[ ||\nabla_{\theta}\log p_t(\theta_t|x)||^2\big]}_{\Omega_t^{3}} \bigg] \mathrm{d}t. \nonumber
\end{align}
For the first term $\Omega_t^{1}$, we have that
\begin{align}
 \Omega_t^{1} &= \int_{\mathbb{R}^d} \int_{\mathbb{R}^n} p_t(\theta_t,x) ||s_{\psi}(\theta_t,x,t)||^2 \mathrm{d}\theta_{t} \mathrm{d}x \\
  &= \int_{\mathbb{R}^d} \int_{\mathbb{R}^n} p_t(\theta_t|x)p(x)  ||s_{\psi}(\theta_t,x,t)||^2 \mathrm{d}\theta_{t} \mathrm{d}x  \\
  &= \int_{\mathbb{R}^d} \int_{\mathbb{R}^n} \left[\int_{\mathbb{R}^n} p_{t|0}(\theta_t|x,\theta_0)p(\theta_0|x) \mathrm{d}\theta_{0} \right] p(x) ||s_{\psi}(\theta_t,x,t)||^2 \mathrm{d}\theta_{t} \mathrm{d}x  \\
  &= \int_{\mathbb{R}^d} \int_{\mathbb{R}^n} \int_{\mathbb{R}^n} p_{t|0}(\theta_t|\theta_0)p(\theta_0|x) p(x)  ||s_{\psi}(\theta_t,x,t)||^2 \mathrm{d}\theta_{t} \mathrm{d}\theta_{0} \mathrm{d}x \\[2mm]
  &= \mathbb{E}_{(\theta_0,x) \sim p(\theta_0,x), \theta_t \sim p_{t|0}(\theta_t|\theta_0)} \left[ ||s_{\psi}(\theta_t,x,t)||^2 \right].
\end{align}

For the second term $\Omega_t^2$, we have that 
\begin{align}
  \hspace{-5mm} \Omega_{t}^2 &= \int_{\mathbb{R}^d} \int_{\mathbb{R}^n} p_t(\theta_t,x) s^{\top}_{\psi}(\theta_t,x,t) \nabla_{\theta} \log p_t(\theta_t|x) \mathrm{d}\theta_{t} \mathrm{d}x  \\
  &= \int_{\mathbb{R}^d} \int_{\mathbb{R}^n} p_t(\theta_t|x)p(x) s^{\top}_{\psi}(\theta_t,x,t) \nabla_{\theta} \log p_t(\theta_t|x) \mathrm{d}\theta_{t} \mathrm{d}x  \\
  &= \int_{\mathbb{R}^d} \int_{\mathbb{R}^n} p(x) s^{\top}_{\psi}(\theta_t,x,t) \nabla_{\theta} p_t(\theta_t|x) \mathrm{d}\theta_{t} \mathrm{d}x \\
  &= \int_{\mathbb{R}^d} \int_{\mathbb{R}^n} p(x) s^{\top}_{\psi}(\theta_t,x,t) \nabla_{\theta_t}\left[\int p_{t|0}(\theta_t|x,\theta_0)p(\theta_0|x) \mathrm{d}\theta_{0} \right]\mathrm{d}\theta_{t} \mathrm{d}x\\
  &= \int_{\mathbb{R}^d} \int_{\mathbb{R}^n} \int_{\mathbb{R}^n} p(\theta_0,x) s^{\top}_{\psi}(\theta_t,x,t) \nabla_{\theta_t}p_{t|0}(\theta_t|\theta_0) \mathrm{d}\theta_{t} \mathrm{d}\theta_{0} \mathrm{d}x  \\
  &= \int_{\mathbb{R}^d} \int_{\mathbb{R}^n} \int_{\mathbb{R}^n} p(\theta_0,x) p_{t|0}(\theta_t|\theta_0) s^{\top}_{\psi}(\theta_t,x,t) \nabla_{\theta_t} \log p_{t|0}(\theta_t|\theta_0) \mathrm{d}\theta_{t} \mathrm{d}\theta_{0} \mathrm{d}x  \\[2mm]
  &= \mathbb{E}_{(\theta_0,x)\sim p(\theta_0,x), \theta_t\sim p_{t|0}(\theta_t|\theta_0)}[s^{\top}_{\psi}(\theta_t,x,t) \nabla_{\theta_t}\log p_{t|0}(\theta_t|\theta_0)]. 
\end{align}

The third term $\Omega_t^{3}$ is independent of $\psi_{\text{post}}$. We thus have 
\begin{align}
\mathcal{J}_{\mathrm{post}}^{\mathrm{DSM}}(\psi)
 &\propto \frac{1}{2} \int_{0}^t \lambda_t \mathbb{E}_{(\theta_0,x)\sim p(\theta_0,x), \theta_t\sim p_{t|0}(\theta_t|\theta_0)} \left[ ||s_{\psi}(\theta_t,x,t)||^2-2 s^{\top}_{\psi}(\theta_t,x,t) \nabla_{\theta_t}\log p_{t|0}(\theta_t|\theta_0) \right]\mathrm{d}t \\
 &\propto \frac{1}{2} \int_0^t\lambda_t \mathbb{E}_{(\theta_0,x)\sim p(\theta_0,x), \theta_t\sim p_{t|0}(\theta_t|\theta_0)} \left[||s_{\psi}(\theta_t,x,t) - \nabla_{\theta_t}\log p_{t|0}(\theta_t|\theta_0) ||^2\right]\mathrm{d}t.
\end{align}

\subsection{Error Bounds for NPSE}
\label{ap:npse-error-bounds}
We now present error bounds for NPSE in the fully deterministic sampling regime, assuming an $L^2$ bound on the approximation error and a mild regularity condition on the target posterior distribution. Our results are based on those obtained in \citet{Benton2024}. 

\subsubsection{Notation}
\label{sec:theory-notation}
We begin by setting up some basic notation. We first recall the definition of the probability flow ODE, now for fixed $x=x_{\mathrm{obs}}$, and $t\in[0,1]$. This ODE is given by
\begin{equation}
\frac{\mathrm{d}\theta_{t}^{\vartheta}}{\mathrm{d}t} = \underbrace{\left[ - f(\theta_{t}^{\vartheta},t) + \frac{1}{2}g^2(t)\nabla_{\theta}\log p_{t}(\theta_{t}^{\vartheta}|x_{\mathrm{obs}})\right]}_{v(\theta_t^{\vartheta},t)}, \quad \theta_0^{\vartheta} = \vartheta \label{eq:prob-flow-ode}
\end{equation}
for each $\vartheta\in\mathbb{R}^d$. The probability flow ODE defines a deterministic coupling between the reference distribution and the target posterior distribution. In particular, if we define $(\theta_t)_{t\in[0,1]}$ by taking $\vartheta\sim \pi$, and setting $\theta_t = \theta_t^{\varphi}$ for all $t\in[0,1]$, then $\theta_1 \sim p(\cdot|x_{\mathrm{obs}})$. Throughout this section, we will write $v(\theta,t)$ for the velocity field defined by \eqref{eq:prob-flow-ode}, and $v_{\psi}(\theta,t)$ for the velocity field corresponding to \eqref{eq:prob-flow-ode} but where the score $\nabla_{\theta}\log p_t(\theta_t|x_{\mathrm{obs}})$ is replaced with its approximation $s_{\psi}(\theta_t,x_{\mathrm{obs}},t)$. We suppress notational dependence of these velocity fields on $x_{\mathrm{obs}}$, since $x_{\mathrm{obs}}$ is assumed to be fixed.

It is worth noting several modifications between the definition of the probability flow ODE in this appendix, and the definition in the main text. First, we here assume that time is rescaled so that the probability flow ODE runs for $t\in[0,1]$, rather than $t\in[0,T]$. Second, we now consider time to run in the opposite direction. In particular, running forward in time, the probability flow ODE in \eqref{eq:prob-flow-ode} transforms the reference distribution to the target distribution, rather than the other way round. We adopt this convention to remain consistent with the setup used in \citet{Benton2024}.

\subsubsection{Assumptions}
We impose the following assumptions, which represent analogues of Assumptions 1, 2, 3, and 4' introduced in \citet{Benton2024}.

\begin{customassumption}{A1}[Bound on Joint $L^2$ Approximation Error]
\label{assumption:a1}
    The true and approximate scores $\nabla_{\theta} \log p_t(\theta_t|x)$ and $s_{\psi}(\theta_t,x,t)$ satisfy $\int_0^1 \mathbb{E}_{p_t(\theta_t,x)}\left[ ||s_{\psi}(\theta_t,x,t) - \nabla_{\theta} \log p_t(\theta_t|x) ||^2 \right] \mathrm{d}t\leq \varepsilon^2$.
\end{customassumption}

\begin{customassumption}{A1'}[Bound on Conditional $L^2$ Approximation Error]
\label{assumption:a1-dash}
   The true and approximate scores $\nabla_{\theta} \log p_t(\theta_t|x_{\mathrm{obs}})$ and $s_{\psi}(\theta_t,x_{\mathrm{obs}},t)$ satisfy $\int_0^1 \mathbb{E}_{p_t(\theta_t|x_{\mathrm{obs}})}\left[ ||s_{\psi}(\theta_t,x_{\mathrm{obs}},t) - \nabla_{\theta} \log p_t(\theta_t|x_{\mathrm{obs}}) ||^2 \right] \mathrm{d}t\leq \varepsilon_{\mathrm{obs}}^2$.
\end{customassumption}

\begin{customassumption}{A2}[Existence and Uniqueness of Smooth Flows]
\label{assumption:a2}
    For each $\vartheta\in\mathbb{R}^d$, and $s\in[0,T]$, there exist
$(\eta_{s,t}^{\vartheta})_{t\in[s,1]}$ and $(\iota_{s,t}^{\vartheta})_{t\in[s,1]}$ starting in $\eta_{s,s}^{\vartheta} = \vartheta$ and $\iota_{s,s}^{\theta}=\vartheta$ with velocity fields $v_{\psi}(\varphi,t)$ and $v(\varphi,t)$ respectively. In addition, $\eta_{s,t}^{\vartheta}$ and $\eta_{s,t}^{\vartheta}$ are continuously differentiable in $\eta$, $s$, and $t$.
\end{customassumption}

\begin{customassumption}{A3}[Regularity of Approximate Score Function]
\label{assumption:a3}
    The score-network $s_{\psi}(\theta,x_{\mathrm{obs}},t)$ is differentiable in its first and last inputs. In addition, for each $t\in(0,1)$, there exists a constant $L_t$ such that $s_{\psi}(\theta,x_{\mathrm{obs}},t)$ is $L_t$ Lipschitz in $\theta$.
\end{customassumption}

\begin{customassumption}{A4}[Regularity of Data Distribution]
\label{assumption:a4}
     Let $\theta\sim p(\cdot|x_{\mathrm{obs}})$. Then, for any $\tau\in(0,\infty)$ and $\xi\sim\mathcal{N}(0,\tau^2 \mathbf{I})$ independent of $\theta$, there exists $\lambda\geq 1$ such that $||\mathrm{Cov}_{\xi | \theta' = \vartheta}(\xi)||_{\mathrm{op}}\leq \lambda \tau^2$ for all $\vartheta\in\mathbb{R}^d$, where $\theta' = \theta + \xi$. 
\end{customassumption}

Assumption \ref{assumption:a1} is arguably the most natural assumption on the training error since we learn the score-network $s_{\psi}$ by minimising the denoising posterior score matching objective in \eqref{eq:weightedposteriorDSM}. This is proportional to the $L^2$ approximation erro in \eqref{eq:weightedposteriorSM},  which appears in the LHS of the bound in Assumption \ref{assumption:a1}. On the other hand, Assumption \ref{assumption:a1-dash} is required to apply the results in \citet{Benton2024}. Below, we provide an additional technical assumption which can be used to translate Assumption \ref{assumption:a1} into Assumption \ref{assumption:a1-dash}.

\begin{customassumption}{B1} \label{assumption:b1}
    Let $\mathcal{A}_{\mathrm{obs}}^{\delta} = \{ x \in\mathbb{R}^p : ||x-x_{\mathrm{obs}}||_2 < \delta \}$, for $\delta>0$. There exists $\delta>0$ such that  $\inf_{x\in\mathcal{A}_{\mathrm{obs}}^{\delta}}p(x)> 0$ and
    $\int_0^1 \mathbb{E}_{p_t(\theta_t|x_{\mathrm{obs}})}|| s_\psi(\theta_t, x_{\mathrm{obs}}, t) - \nabla_{\theta} \log p_t(\theta_t|x_{\mathrm{obs}}) ||^2\mathrm{d}t \leq (C+1) \inf_{x\in \mathcal{A}_{\mathrm{obs}}^{\delta}} \int_0^1 \mathbb{E}_{p_t(\theta_t|x)}|| s_\psi(\theta_t, x, t) - \nabla_{\theta} \log p_t(\theta_t|x) ||^2\mathrm{d}t\vphantom{\int_0^1}$ for some $C\geq 0$.
\end{customassumption}

Alternatively, we can just impose Assumption \ref{assumption:a1-dash} directly. In this case, the results in \citet{Benton2024} can essentially be applied without modification. We refer to, e.g., \citet[Theorem 3.2]{Fu2024} for some conditions under which it is possible to obtain a bound of this type.

\subsubsection{Auxiliary Results}
\label{sec:aux-results}
In order to extend \citet[Theorem 6]{Benton2024} to our setting, we will require some simple additional results. We first establish a lemma which will allow us to translate Assumption \ref{assumption:a1} into Assumption \ref{assumption:a1-dash}.

\begin{lemma}
\label{lemma:conditional-joint}
    Suppose Assumption \ref{assumption:a1} and Assumption 
    \ref{assumption:b1} hold. Then Assumption \ref{assumption:a1-dash} holds. 
\end{lemma}
\begin{proof}
    Let $f(x) = 
    \int_0^1 \mathbb{E}_{p_t(\theta_t|x)}|| s_\psi(\theta_t, x, t) - \nabla_{\theta} \log p_t(\theta_t|x) ||^2\mathrm{d}t$. In addition, let $\smash{K = [ \int_{A_{\mathrm{obs}}^{\delta}} p(x)\mathrm{d}x]^{-1}}$ and $K_1 = K(1+C)$. We then have 
    \allowdisplaybreaks
\begin{align}
   f(x_{\mathrm{obs}}) 
   &= \frac{1}{\int_{\mathcal{A}_{\mathrm{obs}}^{\delta}}p(x)\mathrm{d}x} 
   \left[f(x_{\mathrm{obs}}) 
   \int_{\mathcal{A}_{\mathrm{obs}}^{\delta}}p(x)\mathrm{d}x 
   \right]  
   \\
   &= \frac{1}{\int_{\mathcal{A}_{\mathrm{obs}}^{\delta}}p(x)\mathrm{d}x} 
    \left[\big[\inf_{\mathrm{x\in\mathcal{A}_{\mathrm{obs}}^{\delta}}}f(x)\big]
    \int_{\mathcal{A}_{\mathrm{obs}}^{\delta}}p(x)\mathrm{d}x + \big[f(x_{\mathrm{obs}}) - \inf_{\mathrm{x\in\mathcal{A}_{\mathrm{obs}}^{\delta}}}f(x)\big]
    \int_{\mathcal{A}_{\mathrm{obs}}^{\delta}}p(x)\mathrm{d}x 
     \right]
    \\
    &\leq \frac{1}{\int_{\mathcal{A}_{\mathrm{obs}}^{\delta}}p(x)\mathrm{d}x} 
    \left[\big[\inf_{\mathrm{x\in\mathcal{A}_{\mathrm{obs}}^{\delta}}}f(x)\big]
    \int_{\mathcal{A}_{\mathrm{obs}}^{\delta}}p(x)\mathrm{d}x + C\big[ \inf_{\mathrm{x\in\mathcal{A}_{\mathrm{obs}}^{\delta}}}f(x)\big]
    \int_{\mathcal{A}_{\mathrm{obs}}^{\delta}}p(x)\mathrm{d}x 
     \right] \label{eq:plus-minus}
    \\
    &\leq \frac{C+1}{\int_{\mathcal{A}_{\mathrm{obs}}^{\delta}}p(x)\mathrm{d}x}  
    \left[ \int_{A_{\mathrm{obs}}^{\delta}} f(x)p(x)\mathrm{d}x \right] \label{eq:inf-def} \\[3mm]
    &\leq \frac{C+1}{\int_{\mathcal{A}_{\mathrm{obs}}^{\delta}}p(x)\mathrm{d}x} \left[\int_{\mathbb{R}^p} f(x)p(x)\mathrm{d}x\right] 
    \label{eq:f-positive} \\[3mm]
    &\leq K_1 \varepsilon^2 := \varepsilon_{\mathrm{obs}}^2, \label{eq:final-ineq}
\end{align} 
where in \eqref{eq:plus-minus} we have used Assumption 
\ref{assumption:b1}, in \eqref{eq:inf-def} we have used elementary properties of the infimum, in \eqref{eq:f-positive} we have used the fact that $f(x)\geq 0$ for all $x$, and in \eqref{eq:final-ineq} we have used Assumption \ref{assumption:a1}.
\end{proof}

We next establish a very straightforward lemma which will enable us to convert our $L^2$ bound on the approximate score function (Assumption \ref{assumption:a1-dash}) into an $L^2$ bound on the corresponding velocity field in the probability flow ODE.

\begin{lemma} 
\label{lemma:score-velocity}
Suppose Assumption \ref{assumption:a1-dash} holds. Suppose also that $\sup_{t\in[0,1]}g^{4}(t) <\infty$. Let $v(\theta,t)$ and $v_{\psi}(\theta,t)$ be the true and approximate velocity fields for the probability flow ODE, as defined in Section \ref{sec:theory-notation}. Then there exists $\varepsilon_1>0$ such that $\int_0^1 \mathbb{E}_{p_t(\theta_t|x_{\mathrm{obs}})}\left[ ||v_{\psi}(\theta_t,t) - v(\theta_t,t) ||^2 \right] \mathrm{d}t\leq \varepsilon_1^2$.
\end{lemma}
\begin{proof}
    Straightforwardly, we have that
    \allowdisplaybreaks
    \begin{align}
    \label{eq:bound-on-prob-flow}
        \int_0^1 &\mathbb{E}_{p_t(\theta_t|x_{\mathrm{obs}})}\left[ ||v_{\psi}(\theta_t,t) - v(\theta_t,t) ||^2 \right] \mathrm{d}t \\
        &= \frac{\inf_{t\in[0,1]} \frac{4}{g^4(t)}}{\inf_{t\in[0,1]} \frac{4}{g^4(t)}} \int_0^1 \mathbb{E}_{p_t(\theta_t|x_{\mathrm{obs}})}\left[ ||v_{\psi}(\theta_t,t) - v(\theta_t,t) ||^2 \right] \mathrm{d}t \\
        &\leq \frac{1}{\inf_{t\in[0,1]} \frac{4}{g^4(t)}} \int_0^1 \frac{4}{g^4(t)}\mathbb{E}_{p_t(\theta_t|x_{\mathrm{obs}})}\left[ ||v_{\psi}(\theta_t,t) - v(\theta_t,t) ||^2 \right] \mathrm{d}t \label{eq:ineq-integral} \\
        &=\frac{1}{4} \sup_{t\in[0,1]}[g^4(t)]\int_0^1 \mathbb{E}_{p_t(\theta_t|x_{\mathrm{obs}})}\left[ \bigg|\bigg|\frac{2\left[v_{\psi}(\theta_t,t)+f(\theta_t,t)\right]}{g^{2}(t)} - \frac{2\left[v(\theta_t,t)+f(\theta_t,t)\right]}{g^{2}(t)}\bigg|\bigg|^2 \right] \mathrm{d}t \\
        &= \frac{1}{4} \sup_{t\in[0,1]}[g^4(t)]\int_0^1 \mathbb{E}_{p_t(\theta_t|x_{\mathrm{obs}})}\left[ ||s_{\psi}(\theta_t,x_{\mathrm{obs}},t) - \nabla_{\theta} \log p_t(\theta_t|x_{\mathrm{obs}}) ||^2 \right] 
        \label{eq:sub-v} \\
        &\leq \frac{1}{4}\varepsilon^2\sup_{t\in[0,1]}[g^4(t)] := \varepsilon_1^2, \label{eq:sup-ineq}
    \end{align}
    where in \eqref{eq:ineq-integral} we have used elementary properties of the infimum, in \eqref{eq:sub-v}  we have used the definitions of $v(\theta,t)$ and $v_{\psi}(\theta,t)$, and in \eqref{eq:sup-ineq} we have used Assumption \ref{assumption:a1-dash}.
\end{proof}

\subsection{Main Result}
In order to state our main result, we will require some additional definitions. Following \citet[Corollary 2]{Benton2024}, we first define the quantities $(\beta_t)_{t\in[0,1]}$ and $(\gamma_t)_{t\in[0,1]}$ as
\begin{alignat}{3}
        &\text{VP ODE:}\quad &&\gamma_t = R\cos\left[\left(\frac{\pi}{2} - \delta\right) t\right], \quad &&\beta_t = \sin \left[\left(\frac{\pi}{2} - \delta\right)t\right], \\
        &\text{VE ODE:}\quad &&\gamma_t ~\text{ decreasing}, \quad &&\beta_t =1.
\end{alignat}
We also define $(K_t)_{t\in[0,1]}$ according to 
\begin{equation}
    K_t=\lambda \frac{|\dot{\gamma}_t|}{\gamma_t} + \min \left[ \lambda \frac{|\dot{\beta}_t|}{\beta_t},\lambda^{1/2}R \frac{|\dot{\beta}_t|}{\gamma_t} \right].
\end{equation}
Finally, we let $\mathcal{V}$ denote the class of functions $v:\mathbb{R}^d\times[0,T]\rightarrow\mathbb{R}^d$ which are $K_t$-Lipschitz in $\theta$ for all $t\in[0,1]$.

We are now ready to state our main result: a bound on the error of NPSE in the deterministic sampling regime in terms of the $L^2$ approximation error. 

\begin{theorem}
    Suppose that Assumption \ref{assumption:a1} and \ref{assumption:b1} hold, or that Assumption \ref{assumption:a1-dash} holds. Suppose also that Assumptions \ref{assumption:a2}, \ref{assumption:a3}, and \ref{assumption:a4} hold. Let $v_{\theta}\in\mathcal{V}$. Let $\tilde{\pi}_0=\mathcal{N}(0,\mathbf{I})$, and let $\tilde{\pi}_1$ equal $p(\cdot|x_{\mathrm{obs}})$ plus Gaussian noise with scale $\gamma_{1}\ll 1$. Let $(\eta_t)_{t\in[0,T]}$ be a flow starting in $\tilde{\pi}_0$ with velocity field $v_{\psi}$, and let $\hat{\pi}_1$ be the distribution of $\eta_{1}$. Then, with $\varepsilon_1$ defined as in Lemma \ref{lemma:score-velocity}, 
\begin{alignat}{2}
        &\text{\emph{VP ODE:}}\quad W_2(\hat{\pi}_1,\tilde{\pi}_1)\leq \varepsilon_1\left[\frac{e}{\gamma_1}\right]^{\lambda}, \\
        &\text{\emph{VE ODE:}}\quad W_2(\hat{\pi}_1,\tilde{\pi}_1)\leq \varepsilon_1\left[\frac{1}{\gamma_1}\right]^{\lambda}.
\end{alignat}
\end{theorem}

\begin{proof}
    The result follows directly from \citet[Theorem 6]{Benton2024}, setting the target distribution $\pi:=p(\cdot|x_{\mathrm{obs}})$, for some fixed $x_{\mathrm{obs}}$. We note, in particular, that Assumption 1 in \citet{Benton2024} follows from Assumption \ref{assumption:a1-dash} (or Assumptions \ref{assumption:a1} and \ref{assumption:b1} via Lemma \ref{lemma:conditional-joint}) and Lemma \ref{lemma:score-velocity}. Meanwhile, our Assumptions \ref{assumption:a2}, \ref{assumption:a3}, and \ref{assumption:a4} correspond directly to Assumptions 2, 3, and 4' in \citet{Benton2024}.
\end{proof}

\section{Neural Likelihood Score Estimation}
\label{sec:nlse}
\subsection{Overview}
In this section we outline an alternative method to the one described in Section \ref{sec:SGM} for learning an approximation to the perturbed posterior score $\nabla_{\theta} \log p_t(\theta_t|x)$. We refer to this approach as Neural Likelihood Score Estimation (NLSE). Our alternative approach is based on the following decomposition of the posterior score, which follows straightforwardly from Bayes' theorem:
 \begin{equation}
     \nabla_{\theta} \log p_t(\theta_t|x) =  \nabla_{\theta} \log p_t(x|\theta_t) + \nabla_{\theta} \log p_t(\theta_t), \label{eq:score-bayes}
 \end{equation}
 where $p_t(x|\theta_t) = \int p(x|\theta_0) p_{0|t}(\theta_0|\theta_t)\mathrm{d}\theta_0$ denotes the conditional density of $x$ given $\theta_t$.
 This decomposition suggests that, rather than directly targeting the score of the posterior, we could instead train a score network $\smash{s_{\psi_{\mathrm{lik}}}(\theta_t,x,t)\approx \nabla_{\theta}\log p_t(x|\theta_t)}$ for the score of the perturbed `likelihood' $p_t(x|\theta_t)$, and then estimate the perturbed posterior score by setting 
 \begin{equation}
     s_{\psi_{\text{post}}}(\theta_t,x,t) = s_{\psi_{\mathrm{lik}}}(\theta_t,x,t) + \nabla_{\theta}\log p_t(\theta_t). \label{eq:score_decomp}
 \end{equation}

 In certain cases, it is possible to compute the perturbed prior $p_t(\theta_t)$ in closed form, and thus obtain the perturbed prior score $\nabla_{\theta} \log p_t(\theta_t)$ via automatic differentiation (see Appendix \ref{app:perturbed_prior}). In cases where this is not possible, we can instead approximate this term using an additional score network $s_{\psi_{\text{pri}}}(\theta_t,t)\approx \nabla_{\theta}\log p_t(\theta_t)$ (see Appendix \ref{app:prior_estimation}).
 
 In order to train the time-varying score network $s_{\psi_{\mathrm{lik}}}(\theta_t,x,t)$, it is once again natural to minimise a weighted Fisher divergence, which now reads
 \begin{align}
\label{eq:weightedlikelihoodSM}
\mathcal{J}_{\mathrm{lik}}^{\mathrm{SM}}(\psi_{\mathrm{lik}}):=\frac{1}{2} \int_{0}^T\lambda_t \mathbb{E}_{p_t(\theta_t,x)}\big[ \big|\big| s_{\psi_{\mathrm{lik}}}(\theta_t,x,t) - \nabla_{\theta} \log p_t(x|\theta_t)\big|\big|^2\big]\mathrm{d}t.
 \end{align}
 Similar to \eqref{eq:weightedposteriorSM}, we cannot optimise this objective due to the intractable second term. However, by substituting \eqref{eq:score-bayes}, and arguing as in Appendix \ref{ap:npse-1}, one can show that it is equivalent 
 to minimise the denoising likelihood score matching objective function, given by\footnote{See also \citet[Theorem 1]{Chao2022} for a slightly more general version of this result.}
 \begin{align}
\label{eq:weightedlikelihoodDSM} \mathcal{J}^{\mathrm{DSM}}_{\mathrm{lik}}(\psi_{\mathrm{lik}}) := \frac{1}{2} \int_{0}^T\lambda_t \mathbb{E}_{p_{t|0}(\theta_t|\theta_0)p(\theta_0,x)}  \big[\big|\big|s_{\psi_\mathrm{lik}}(\theta_t,x,t)+ \nabla_{\theta} \log p_t(\theta_t)   - \nabla_{\theta} \log p_{t|0}(\theta_t|\theta_0)\big|\big|^2\big]\hspace{.5mm}\mathrm{d}t. 
 \end{align}
  Similar to \eqref{eq:weightedposteriorDSM},  given
a suitable choice for the drift and diffusion coefficients in
\eqref{eq:forwardSDE}, the scores $\nabla_{\theta_t}\log p_{t|0}(\theta_t|\theta_0)$ can be computed in closed
form. We can compute Monte Carlo estimates of \eqref{eq:weightedlikelihoodDSM}, and minimise this using standard methods to obtain $s_{\psi_{\mathrm{lik}}}(\theta_t,x,t)\approx \nabla_{\theta_t}\log p_t(x|\theta_t)$.

\subsection{Computing or Estimating the Perturbed Prior Score}
\label{app:prior}

\subsubsection{Computing the Perturbed Prior Score}
\label{app:perturbed_prior}
For certain choices of the prior, and certain choices of the drift and diffusion coefficients in the forward SDE \eqref{eq:forwardSDE}, we can obtain the perturbed prior $p_{t}(\theta_t) = \int_{\mathbb{R}^d}  p_{t|0}(\theta_t|\theta_0)p(\theta_0)\mathrm{d}\theta_0$ in closed form. We can then obtain the score of the perturbed prior $\nabla_{\theta} \log p_t(\theta_t)$ using automatic differentiation \citep[e.g.,][]{Bartholomew-Biggs2000}. 

Suppose, for example, that the drift and diffusion coefficients in \eqref{eq:forwardSDE} are given by $f(\theta_t,t)=0$ and $g(t) = \tau_t$, where $(\tau_t)_{t\in[0,T]}$ is a positive sequence of reals. In this case, we have $p_{t|0}(\theta_t|\theta)=\mathcal{N}(\theta_t|\theta,\tau_t^2 \mathbf{I})$, and can obtain $p_t(\theta_t)$ in closed form for the following common choices of prior.

\textbf{Uniform Prior}. 
Suppose that $p(\theta) = \mathcal{U}(\theta | a, b)$. We can then compute, writing $\Phi(\cdot|\mu,\sigma^2)$ is the CDF of a univariate Gaussian with mean $\mu$ and variance $\sigma^2$,  
\begin{align}
    p_t(\theta_t) 
    &= \int_{\mathbb{R}^d} p(\theta_0) p_{t|0}(\theta_t|\theta_0) \mathrm{d}\theta_0  \\
     &= \frac{1}{\prod_{i=1}^{d} (b_i - a_i)} \int_{[a_1,b_1] \times \cdots \times [a_d,b_d]} \mathcal{N}(\theta_t|\theta_0,\tau_t^2 \mathbf{I}) \mathrm{d}\theta_0 \\
     &= \frac{1}{\prod_{i=1}^{d} (b_i - a_i)} \prod_{i=1}^{d} \left( \Phi(b_i|\theta_{t,i}, \tau_{i,t}^2) - \Phi(a_i|\theta_{t,i}, \tau_{i,t}^2) \right).
\end{align}

\textbf{Gaussian Mixture Prior}. 
Suppose that $p(\theta) = \sum_{i=1}^n \alpha_i \mathcal{N}(\theta|\mu_i,\Sigma_i)$. Using standard results \citep[e.g.,][Equation 2.115]{Bishop2006}, we then have that
\begin{align}
    p_t(\theta_t) 
    &= \int_{\mathbb{R}^d} p(\theta_0) p_{t|0}(\theta_t|\theta_0) \mathrm{d}\theta_0  \\
    &= \sum_{i=1}^n \alpha_i \int_{\mathbb{R}^d} \mathcal{N}(\theta_0|\mu_i,\Sigma_i) \mathcal{N}(\theta_t|\theta_0,\tau_t^2 \mathbf{I}) \mathrm{d}\theta_0 \\
    &= \sum_{i=1}^n \alpha_i \mathcal{N}\left(\theta_t|\mu_i, \Sigma_i + \tau_t^2 \mathbf{I} \right).
\end{align}

\subsubsection{Estimating the Perturbed Prior Score}
\label{app:prior_estimation}
In cases where it is not possible to obtain the perturbed prior in closed form (e.g., the prior is implicit), we can instead learn an approximation $s_{\psi_{\mathrm{pri}}}(\theta_t,t)\approx \nabla_{\theta} \log p_{t}(\theta_t)$ using denoising score matching, by minimising a Monte Carlo estimate of 
\begin{equation}
{\mathcal{J}}_{\text{pri}}(\psi_{\text{pri}}) = \frac{1}{2} \int_{0}^{T} \lambda_t \mathbb{E}_{p(\theta_0) p_{t|0}(\theta_t|\theta_0)}\left[  \left|\left|s_{\psi_{\text{pri}}}(\theta_{t},t)- \nabla_{\theta} \log p_{t|0}(\theta_{t}|\theta_{0}) \right|\right|^2\right]. \label{eq:prior-dsm}
\end{equation}
We summarise this procedure below.
\begin{algorithm*}[!ht]
   \caption{Prior Score Estimation}
   \label{alg:1}
\begin{algorithmic}
\STATE {\bfseries Input:} prior $p(\theta)$, prior sample budget $N$, dataset $\mathcal{D} = \{\}$.
   \STATE {\bfseries Outputs:} $s_{\psi_{\mathrm{pri}}}(\theta_t, t) \approx \nabla_{\theta_t} \log p_t(\theta_t)$
  \FOR {$i=1:N$}
   	\STATE Sample $\theta_{i} \sim {p}(\theta)$.
	\STATE Add $\theta_i$ to $\mathcal{D}$.
  \ENDFOR
  \STATE Learn $s_{\psi_{\mathrm{pri}}}(\theta_t, t) \approx \nabla_{\theta_t} \log p_t(\theta_t)$ by minimising a Monte Carlo estimate of \eqref{eq:prior-dsm} based on $\mathcal{D}$.
  \STATE {\bfseries{Return:}} $s_{\psi_{\mathrm{pri}}}(\theta_t, t)$
\end{algorithmic}
\end{algorithm*}

\subsection{NPSE versus NLSE}
A natural question to ask is whether it is preferable to use NLSE or NPSE. In numerical testing, we observed significantly better performance for NPSE relative to NLSE in cases where it was not possible to compute the score of the perturbed prior, and it was thus necessary to approximate this quantity using an additional score network (see Appendix \ref{app:prior_estimation}). 

Meanwhile, in cases where it was possible to compute the perturbed prior analytically (see Appendix \ref{app:perturbed_prior}), we found little empirical difference between NPSE and NLSE. To illustrate this point, we provide results for four benchmark experiments in Figure \ref{fig:nlse}. For NPSE, we report results using both the VE SDE and VP SDE (see Appendix \ref{ap:sde}) for the forward noising process. For NLSE, we use the VE SDE (see Appendix \ref{ap:sde}), as this allowed us to easily compute the perturbed prior in closed form (see Appendix \ref{app:perturbed_prior}).  

\begin{figure*}[h!]
\centering
\includegraphics[width=1\textwidth, trim=60 20 40 20, clip]{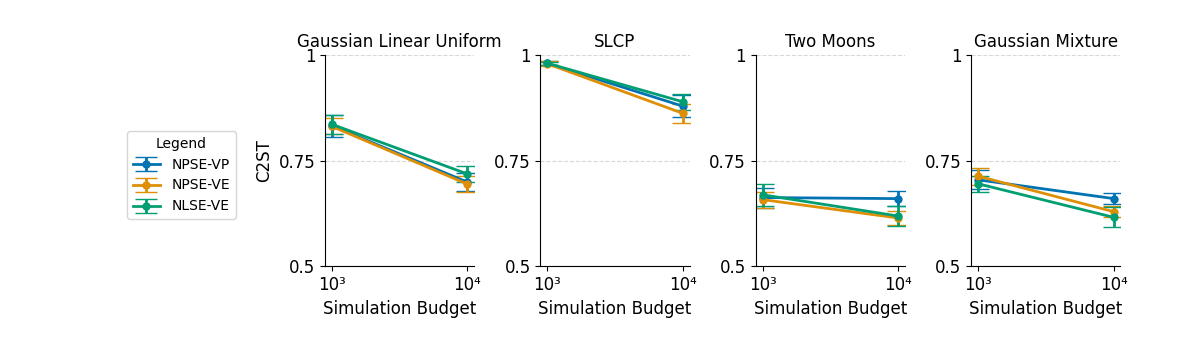}
  \vspace{-1mm}
\caption{\textbf{Comparison between NPSE and NLSE on four benchmark tasks.}}
  \label{fig:nlse}
\end{figure*}

\section{Sequential Neural Posterior Score Estimation: Additional Details}
\label{sec:sequential-additional}

In this section we provide further details of the various sequential methods introduced in Section \ref{sec:sequential}: TSNPSE  (Section \ref{sec:tnspse}), SNPSE-A (Section \ref{sec:snpse-a}), SNPSE-B (Section \ref{sec:snpse-b}), and SNPSE-C (Section \ref{sec:snpse-c}).

\subsection{TSNPSE}
\label{sec:tnspse}
We begin with our main sequential algorithm: TSNPSE. In particular, the following section contains a proof of the theoretical result (\cref{prop:main-result}) provided in the main text.

\subsubsection{Theoretical Results}
\begin{proposition}
\label{prop:main}
    Let $\tilde{p}^{r}(\theta) = \frac{1}{r}\sum_{s=0}^{r-1} \bar{p}^{s}(\theta)$, where $\bar{p}^{0}(\theta) = p(\theta)$ and $\bar{p}^{s}(\theta)$ is defined by \eqref{eq:truncation} for all $s\geq 1$. Suppose that 
    \begin{equation}
        \Theta_{\mathrm{obs}} \subseteq \mathrm{HPR}_\epsilon({p}_{\psi}^{s}(\theta|x_\mathrm{obs}))
    \end{equation}
    for all $s\geq 1$, where $\Theta_{\mathrm{obs}} = \mathrm{supp}(p(\cdot|x_{\mathrm{obs}}))$. 
    Then, writing $\tilde{p}_t^{r}(\theta_t, x)$ for the distribution of $(\theta_t, x)$ when $(\theta_0,x) \sim \tilde{p}^{r}(\theta,x)$, the minimiser $\psi^{*}$ of the loss function
    \begin{align}
    \label{eq:tsnpse-loss-v2}
&\mathcal{J}^{\mathrm{TSNPSE-SM}}_{\mathrm{post}}(\psi) 
    := \frac{1}{2} \int_{0}^{T} \lambda_t \mathbb{E}_{\tilde{p}_t^{r}(\theta_t, x)} \left[|| s_{\psi}(\theta_t, x, t) - \nabla_{\theta} \log p_t(\theta_t|x) ||^2\right] \mathrm{d}t, \\
\intertext{or, equivalently, of the loss function}
&\mathcal{J}^{\mathrm{TSNPSE-DSM}}_{\mathrm{post}}(\psi) 
    := \frac{1}{2} \int_{0}^{T} \lambda_t \mathbb{E}_{p_{t|0}(\theta_t|\theta_0)p(x|\theta_0)\tilde{p}^r(\theta_0)}\left[|| s_{\psi}(\theta_t, x, t) - \nabla_{\theta} \log p_{t|0}(\theta_t|\theta_0) ||^2\right] \mathrm{d}t, 
\label{eq:tsnpse-loss-denoising-v2}
\end{align}
satisfies $s_{\psi^{\star}}(\theta_t, x_\mathrm{obs}, t) = \nabla_{\theta} \log p_t(\theta_t|x_{\mathrm{obs}})$.
\end{proposition}

\begin{proof}
By definition, we have $\tilde{p}^{r}(\theta) \propto c^{r}(\theta) p(\theta)$, where $c^{r}(\theta) := \frac{1}{r}\sum_{s=0}^{r-1} \mathbb{I}\{\theta \in \Theta^{s}\}$, with $\Theta^{0} = \mathrm{supp} (p(\theta))$ and $\Theta^{s}=\mathbb{I}\{\theta\in\mathrm{HPR}_\varepsilon({p}_{\psi}^{s}(\theta|x_\mathrm{obs}))\}$ for all $s\geq 1$. Thus, in particular, we can write
\begin{equation}
    {p}^{r}(\theta) = \frac{c^{r}(\theta) p(\theta)}{Z^{r}} 
    := f^{r}(\theta) p(\theta), 
\end{equation}
where $f^{r}(\theta) = \frac{c^{r}(\theta)}{Z^{r}}$, and where $Z^{r}$ is the normalisation constant
\begin{equation}
    Z^{r} =\int_{\mathbb{R}^d}c^{r}(\theta)  p(\theta)\mathrm{d}\theta= \frac{1}{r}\sum_{s=0}^{r-1} \int_{\Theta^{s}} p(\theta)\mathrm{d}\theta.
\end{equation}
By definition, $\smash{c^{r}(\theta)=\frac{1}{r}\sum_{s=0}^{r-1} \mathbb{I}\{\theta \in \Theta^{s}\}=1}$ for all $\smash{\theta\in\cup_{s=0}^{r-1}\Theta^{s}}$. Under the assumption on $\Theta_{\mathrm{obs}}$, we thus also have that $c^{r}(\theta)=1$ for all $\theta\in\Theta_{\mathrm{obs}}$. It follows that 
\begin{equation}
    f^{r}(\theta) = \frac{1}{Z^{r}} = \mathrm{constant} = A_{r}
\end{equation}
for all $\theta\in\Theta_{\mathrm{obs}}$. Thus, in particular, we have that $p^{r}(\theta) = A_{r} \cdot p(\theta)$ for all $\theta\in\Theta_{\mathrm{obs}}$; see also \citet[Section 6.2]{Deistler2022}. 

Now, using standard results \citep[e.g.,][Theorem 1]{Batzolis2021}, we know that $\psi^{\star} = \arg \min \mathcal{J}^{\mathrm{TSNPSE-SM}}_{\mathrm{post}}(\psi) = \arg \min \mathcal{J}^{\mathrm{TSNPSE-DSM}}_{\mathrm{post}}(\psi)$ satisfies 
\begin{equation}
    s_{\psi^{\star}}(\theta_t, x, t) = \nabla_{\theta} \log \tilde{p}_t^{r}(\theta_t|x), \label{eq:truncated-minimiser}
\end{equation}
where, similar to before, $\tilde{p}_t^{r}(\theta_t| x) = \int_{\mathbb{R}^d} p_{t|0}(\theta_t| \theta_0) \tilde{p}^{r}(\theta_0|x) \mathrm{d}\theta_0$,
and $\tilde{p}^{r}(\theta_0|x) = \frac{\tilde{p}^{r}(\theta_0) p(x|\theta_0) }{\tilde{p}^{r}(x)}$. Thus, at the observation $x_{\mathrm{obs}}$, we have that
\allowdisplaybreaks
\begin{align}
    \tilde{p}_t^{r}(\theta_t| x_\mathrm{obs}) 
    &= 
    \int_{\mathbb{R}^d} p_{t|0}(\theta_t| \theta_0) \tilde{p}^{r}(\theta_0|x_\mathrm{obs}) \mathrm{d}\theta_0  \\
    &= \int_{\mathbb{R}^d} p_{t|0}(\theta_t| \theta_0) \frac{\tilde{p}^{r}(\theta_0){p}(x_\mathrm{obs}|\theta_0)}{\tilde{p}^{r}(x_\mathrm{obs})} \mathrm{d}\theta_0  \\
    &= \int_{\mathbb{R}^d} p_{t|0}(\theta_t| \theta_0) \frac{f^{r}(\theta_0) p(\theta_0){p}(x_\mathrm{obs}|\theta_0)}{\tilde{p}^{r}(x_\mathrm{obs})} \mathrm{d}\theta_0  \\
    &=\int_{\mathbb{R}^d} p_{t|0}(\theta_t| \theta_0) \frac{f^{r}(\theta_0) {p}(\theta_0|x_\mathrm{obs})p(x_{\mathrm{obs}})}{\tilde{p}^{r}(x_\mathrm{obs})} \mathrm{d}\theta_0  \\
    &\propto \int_{\mathbb{R}^d} f^{r}(\theta_0) p_{t|0}(\theta_t|\theta_0)  p(\theta_0|x_{\mathrm{obs}})\mathrm{d}\theta_0 \\
    &= 
     A_{r} \cdot \int_{\Theta_{\mathrm{obs}}} p_{t|0}(\theta_t|\theta_0)  p(\theta_0|x_{\mathrm{obs}})\mathrm{d}\theta_0 \\
     &= A_{r} \cdot p_t(\theta_t|x_\mathrm{obs}), \label{eq:perturbed-identity}
\end{align}
where the penultimate equality holds since $f^{r}(\theta) =A_{r}$ for all $\theta\in\Theta_{\mathrm{obs}}$, and $p(\theta|x_{\mathrm{obs}})=0$ for all $\theta\in\mathbb{R}^d \setminus\Theta_{\mathrm{obs}}$. Thus, combining \eqref{eq:truncated-minimiser} and the logarithmic derivative of \eqref{eq:perturbed-identity}, we conclude that
$s_{\psi^{\star}}(\theta_t, x_{\mathrm{obs}}, t)= \nabla_{\theta} \log \tilde{p}_t^{r}(\theta_t|x_{\mathrm{obs}})= \nabla_{\theta} \log p_t(\theta_t|x_{\mathrm{obs}})$ as required.
\end{proof}

\subsection{SNPSE-A}
\label{sec:snpse-a}
We now provide more details on SNPSE-A, so-named due to its connections with SNPE-A \cite{Papamakarios2016}. This algorithm is summarised in Algorithm \ref{alg:snpse-a}.

\subsubsection{Overview} The main steps involved in SNPSE-A can be summarised as follows. For $r=1$, sample parameters from the prior $\smash{\{\theta_{0,i}^{1}\}_{i=1}^{M}\sim p(\theta):=p^{0}_{\psi}(\cdot|x_{\mathrm{obs}})}$. Then, for $r=1,2$, 
\begin{itemize}
    \item[(i)] 
    Simulate new data $\smash{\{x_i^{r}\}_{i=1}^{M} \sim p(\cdot|\theta_{0,i}^{r})}$. Concatenate samples $\smash{\{(\theta_{0,i}^{r},x_i^{r})\}_{i=1}^M}$ with those from previous rounds to form $\smash{\{(\theta_{0,i},x_i)\}_{i=1}^{rM}:=\bigcup_{s=1}^{r}\{(\theta_{0,i}^{r}, x_i^{r})\}_{i=1}^{M} \sim \tilde{p}^{r}(\theta)p(x|\theta)}$, where $\smash{\tilde{p}^{r}(\theta) = \frac{1}{r}\sum_{s=0}^{r-1} p_{\psi}^{s}(\theta|x_\mathrm{obs})}$. Draw times $\{t_i\}_{i=1}^{rM}\sim\mathcal{U}(0,T)$, and samples $\{\theta_{t_i,i}\}_{i=1}^{rM} \sim p_{t|0}(\cdot|\theta_{0,i})$.
    \item[(ii)] Using these samples, train a time-varying score network $\tilde{s}^{r}_{\psi}(\theta_t,x,t)$ to approximate the score of the proposal posterior $\nabla_{\theta}\log \tilde{p}_t^{r}(\theta_t|x)$, by minimising a Monte Carlo estimate of the original denoising posterior score matching objective, but now over samples from the proposal prior. That is,
    \begin{align}
{\mathcal{J}}^{\mathrm{DSM-A}}_{\mathrm{post}}(\psi) =\frac{1}{2}\int_{0}^T\lambda_t &\mathbb{E}_{p_{t|0}({\theta}_t|\theta_0)p(x|\theta_0)\tilde{p}^{r}(\theta_0)}  \left[ \left|\left| \tilde{s}^{r}_{{\psi}}({\theta}_t,x,t) - \nabla_{{\theta}_t} \log p_{t|0}({\theta}_t|\theta_0)\right|\right|^2\right]\mathrm{d}t.
\label{eq:snpse-a} 
\end{align}
\item[(iii)] Draw samples $\{\theta^{r+1}_{T,i}\}_{i=1}^{M'}\sim \pi(\theta)$, where $M'\geq M$. Simulate the backward SDE \eqref{eq:backwardSDE} or 
the probability flow ODE \eqref{eq:backwardODE}, substituting $\tilde{s}^{r}_{\psi}(\theta_t,x_{\mathrm{obs}},t)\approx\nabla_{\theta}\log \tilde{p}_t^{r}(\theta_t|x_{\mathrm{obs}})$, to obtain samples $\smash{\{\tilde{\theta}^{r+1}_{0,i}\}_{i=1}^{M'}\sim {\tilde{p}}_{\psi}^{r}(\cdot|x_{\mathrm{obs}})\approx \tilde{p}^{r}(\cdot|x_{\mathrm{obs}})}$. 
\item[(iv)] Use (approximate) sampling-importance-resampling (SIR) \cite{Rubin1987,Rubin1988,Smith1992,Gelman1995} to recover samples $\smash{\{\theta_{0,i}^{r+1}\}_{i=1}^M \sim p_{\psi}^{r}(\cdot|x_{\mathrm{obs}})\approx p(\cdot|x_{\mathrm{obs}})}$. In particular, draw samples $\smash{\{\theta_{0,i}^{r+1}\}_{i=1}^M}$ with or without replacement from $\smash{\{\tilde{\theta}_{0,i}^{r+1}\}_{i=1}^{M'}}$ using sample probabilities $\smash{{w}_i^{r}}$ proportional to
\begin{equation}
{h}_i^{r} = \frac{p(\tilde{\theta}_i^{r+1})}{\tilde{p}^{r}(\tilde{\theta}_i^{r+1})} ~,\quad i\in[M']. \label{eq:importance-weights}
\end{equation}
This corresponds, at the level of densities, to updating the current posterior density estimate as \citep[see also][]{Papamakarios2016}
\begin{equation}
    p_{\psi}^{r}(\theta|x_{\mathrm{obs}}) \propto \frac{p(\theta)}{\tilde{p}^{r}(\theta)} \tilde{p}_{\psi}^r(\theta|x_{\mathrm{obs}}). \label{eq:new-posterior}
\end{equation}
\end{itemize}

\begin{algorithm}[h!]
\caption{SNPSE-A}
\label{alg:snpse-a}
\begin{algorithmic}
    \STATE \textbf{Inputs:} Observation $x_\mathrm{obs}$, prior $p(\theta) =: {p}_{\psi}^{0}(\theta|x_\mathrm{obs})$, simulator $p(x|\theta)$, simulation budget $N$, number of rounds $R$, (simulations-per-round $M=N/R$)
    \STATE \textbf{Outputs:} Samples $\theta \sim p^{r}_{\psi}(\theta|x_\mathrm{obs}) \approx p(\theta|x_\mathrm{obs})$
    \FOR{$r = 1,2$}
        \FOR{$i = 1,\dots,M$}
            \STATE Draw $\theta_i \sim {p}_{\psi}^{r-1}(\theta|x_\mathrm{obs})$ using \eqref{eq:new-posterior} (requires importance weights, see Appendix \ref{ap:iw-snpse-a} for details),
            $x_i \sim p(x|\theta_{i})$. 
            \STATE Add $(\theta_i, x_i)$ to $\mathcal{D}$.
        \ENDFOR
        \STATE Learn $\tilde{s}^{r}_{\psi}(\theta_t,x,t)\approx \nabla_{\theta} \log \tilde{p}^{r}_t(\theta_t|x)$ by minimising a Monte Carlo estimate of \eqref{eq:snpse-a} based on dataset $\mathcal{D}$.
        \STATE Get $\tilde{p}_\psi^{r}(\theta|x_{\mathrm{obs}})$ sampler by substituting $\tilde{s}^{r}_{\psi}(\theta_t,x_{\mathrm{obs}},t)\approx \nabla_{\theta} \log \tilde{p}_t^{r}(\theta_t|x_{\mathrm{obs}})$ into \eqref{eq:backwardSDE} or \eqref{eq:backwardODE}.
    \ENDFOR
    \STATE \textbf{Return:} $\theta \sim p_\psi^{r}(\theta|x_{\mathrm{obs}})$ using \eqref{eq:new-posterior} (requires importance weights, see Appendix \ref{ap:iw-snpse-a} for details).
\end{algorithmic}
\end{algorithm}

\subsubsection{Theoretical Justification}
We can formally justify this procedure as follows. First, using standard results on conditional denoising score matching \citep[e.g.,][]{Batzolis2021}, the minimiser 
\begin{equation}
    \psi^{*} = \mathrm{argmin}_{\psi}{\mathcal{J}}^{\mathrm{DSM-A}}_{\mathrm{post}}(\psi)
\end{equation}
is such that $\tilde{s}^{r}_{\psi^{*}}(\theta_t,x,t)=\nabla_{\theta}\log \tilde{p}_t^{r}(\theta_t|x)$ for almost all $\theta_t\in\mathbb{R}^d$, $x\in\mathbb{R}^p$, and $t\in[0,T]$. Thus, by substituting the score network $\smash{\tilde{s}_{\psi^{*}}^{r}(\theta_t,x,t)}$ into \eqref{eq:backwardSDE} or \eqref{eq:backwardODE} we can, in principle, generate samples from the true proposal posterior, 
\begin{equation}\{{\tilde{\theta}_i^{r+1}\}_{i=1}^{M'}\sim\tilde{p}^{r}(\theta|x_{\mathrm{obs}})}.
\end{equation}  
It follows, using classical results on SIR \citep[e.g.,][]{Gelfand1992}, that if we resample $\{\theta_i^{r+1}\}_{i=1}^M$ from $\{\tilde{\theta}_i^{r+1}\}_{i=1}^{M'}$ with or without replacement, using sample probabilities ${w}_i^{r}$ proportional to the importance weights
\begin{align}
{h}_i^{r} =\frac{p(\tilde{\theta}_i^{r+1}|x_{\mathrm{obs}})}{\tilde{p}^{r}(\tilde{\theta}_i^{r+1}|x_{\mathrm{obs}})} \propto \frac{p(\tilde{\theta}_i^{r+1})}{\tilde{p}^{r}(\tilde{\theta}_i^{r+1})} 
\label{eq:true-importance}
\end{align}
then, in the limit as $M'\rightarrow\infty$, the resulting samples will correspond to i.i.d. draws from the target posterior $p(\cdot|x_{\mathrm{obs}})$, as required. These are precisely the importance weights that we use in SNPSE-A, c.f. \eqref{eq:importance-weights}.

In practice, of course, we will never obtain the minimiser $\psi^{*}= \mathrm{argmin}_{\psi}{\mathcal{J}}^{\mathrm{DSM-A}}_{\mathrm{post}}(\psi)$ 
but instead some $\psi$ such that, hopefully, $\tilde{s}_{\psi}^{r}(\theta_t,x,t)\approx \nabla_{\theta} \log \tilde{p}_t^{r}(\theta_t|x)$ or, alternatively, such that $\tilde{p}_{\psi}^{r}(\theta|x_{\mathrm{obs}}) \approx \tilde{p}^{r}(\theta|x_{\mathrm{obs}})$. Here, as before, we use $\smash{{\tilde{p}}_{\psi}^{r}(\theta|x_{\mathrm{obs}})}$ to denote the approximation of the true proposal posterior $\smash{\tilde{p}^{r}(\theta|x_{\mathrm{obs}})}$ obtained by substituting $\smash{\tilde{s}_{\psi}^{r}(\theta_t,x,t)\approx\nabla_{\theta}\log \tilde{p}^{r}_t(\theta_t,x,t)}$ into the 
probability flow ODE \eqref{eq:backwardODE}.
Using the score network $\tilde{s}_{\psi}^{r}(\theta_t,x,t)$, we can now generate samples from an approximation of the proposal posterior, rather than the true proposal posterior:
\begin{equation}
    \{\tilde{\theta}_i^{r+1}\}_{i=1}^{M'}\sim \tilde{p}_{\psi}^{r}(\cdot|x_{\mathrm{obs}})\approx \tilde{p}^{r}(\theta|x_{\mathrm{obs}}).
\end{equation}
It follows, once again appealing to standard results on SIR, that in this case the correct probabilities to use in order to recover samples from the true posterior are sample probabilities $\tilde{w}_i^{r}$ proportional to the importance weights
\begin{align}
    \tilde{h}_i^{r} = \frac{p(\tilde{\theta}_i^{r+1}|x_{\mathrm{obs}})}{\tilde{p}^{r}_{\psi}(\tilde{\theta}_i^{r+1}|x_{\mathrm{obs}})}. 
\end{align}
These importance weights are only approximately equal to \eqref{eq:true-importance}, i.e., the importance weights that we actually use in SNPSE-A:
\begin{align}
\tilde{h}_i^{r} = \frac{p(\tilde{\theta}_i^{r+1}|x_{\mathrm{obs}})}{\tilde{p}^{r}_{\psi}(\tilde{\theta}_i^{r+1}|x_{\mathrm{obs}})} 
\approx \frac{p(\tilde{\theta}_i^{r+1}|x_{\mathrm{obs}})}{\tilde{p}^{r}(\tilde{\theta}_i^{r+1}|x_{\mathrm{obs}})} 
= {h}_i^r, \label{eq:importance-v2} 
\end{align}
since we will only ever learn an approximation of the true proposal posterior scores, $\smash{\tilde{s}_{\psi}^{r}(\theta_t,x,t)\approx \nabla_{\theta} \log \tilde{p}_t^{r}(\theta_t,x,t)}$, and thus an approximation of the true proposal posterior,  $\smash{\tilde{p}_{\psi}^{r}(\theta|x_{\mathrm{obs}})\approx \tilde{p}^{r}(\theta|x_{\mathrm{obs}})}$, when we minimise the SNPSE-A score-matching objective ${\mathcal{J}}^{\mathrm{DSM-A}}_{\mathrm{post}}(\psi)$ over a finite number of samples. 

It follows that, when we perform a post-hoc correction in the $r^{\mathrm{th}}$ round using sample probabilities proportional to ${h}_i^{r}$ rather than $\tilde{h}_i^{r}$, we necessarily introduce an additional approximation into the sequential procedure. This approximation is directly related to the scale of the mismatch between the true proposal posterior $\tilde{p}^{r}(\theta|x_{\mathrm{obs}})$, and the approximate proposal posterior $\tilde{p}_{\psi}^{r}(\theta|x_{\mathrm{obs}})$ learned in the $r^{\mathrm{th}}$ round.

\subsubsection{Computing the Importance Weights} \label{ap:iw-snpse-a}
SNPSE-A relies on being able to compute the importance weights in \eqref{eq:importance-weights}. In particular, it is necessary to compute the density ratio between the prior and the proposal prior, viz 
\begin{equation}
    \frac{p(\theta)}{\tilde{p}^{r}(\theta)} = \frac{p(\theta)}{\frac{1}{r}\sum_{s=0}^{r-1}{p}_{\psi}^{s}(\theta|x_{\mathrm{obs}})}. \label{eq:weights}
\end{equation}
Since the prior $p(\theta)$ is typically available in closed form, it just remains to compute the proposal prior $\tilde{p}^{r}(\theta)$. In the following sections, we outline several possible ways to compute or approxiate this term.

\textbf{Computing the Proposal Prior}. The first and most direct approach is to compute the proposal prior using the probability flow ODE \eqref{eq:backwardODE} and the instantaneous change-of-variables formula \eqref{eq:change_of_var}. To be precise, by substituting our estimate of the proposal posterior score,  $\tilde{s}_{\psi}^{r}(\theta_t,x_{\mathrm{obs}},t)\approx \nabla_{\theta} \log \tilde{p}_t^{r}(\theta_t|x_{\mathrm{obs}})$ into the probability flow ODE \eqref{eq:backwardODE}, it is possible to evaluate the approximate proposal posterior density $\tilde{p}_{\psi}^{r}(\theta|x_{\mathrm{obs}})\approx \tilde{p}^{r}(\theta|x_{\mathrm{obs}})$ via the instantaneous change of variable formula \eqref{eq:change_of_var}. 

Unfortunately, even with access to approximate proposal posterior densities $\tilde{p}_{\psi}^{r}(\theta|x_{\mathrm{obs}})$, it turns out that we can only compute the approximate posterior densities $p_{\psi}^{r}(\theta|x_{\mathrm{obs}})$ for $r=0,1$. Thus, by definition, we can only compute the proposal prior $\tilde{p}^{r}(\theta)$ and the importance weights $\smash{\frac{p(\theta)}{\tilde{p}^{r}(\theta)}}$ for $r=1,2$. In other words, it is only possible to compute the required weights in \eqref{eq:weights} for up to 2 rounds.

To illustrate this, we can consider explicitly the quantities required to compute the proposal priors $\smash{\tilde{p}^{1}(\theta),\tilde{p}^{2}(\theta),\dots}$ in rounds $r=1,2,3$. Recall that the proposal priors are defined as the mixture of the posterior estimates $p_{\psi}^{0}(\theta|x_{\mathrm{obs}}), p_{\psi}^{1}(\theta|x_{\mathrm{obs}}), \dots$ from the previous rounds, c.f. \eqref{eq:weights}. For these proposals to be computable, we must be able to express them in terms of the prior $p(\theta)$, which we assume is known, and the proposal posterior estimates $\smash{\tilde{p}_{\psi}^{1}(\theta|x_{\mathrm{obs}}),\tilde{p}_{\psi}^{2}(\theta|x_{\mathrm{obs}}),\dots}$, which we can always access using the instantaneous change of variables formula.

At the start of the $1^{\mathrm{st}}$ round, the current `posterior estimate' is initialised equal to the prior: $p_{\psi}^{0}(\theta|x_{\mathrm{obs}}) = p(\theta)$. Thus, in the $\smash{1^{\mathrm{st}}}$ round, the proposal prior is just equal to the prior: 
\begin{equation}
    \tilde{p}^{1}(\theta)=p^{0}_{\psi}(\theta|x_{\mathrm{obs}}) = p(\theta). \label{eq:1st-round}
\end{equation}
At the start of the $2^{\mathrm{nd}}$ round, the current posterior estimate is equal to the proposal posterior estimate obtained in the $1^{\mathrm{st}}$ round: $\smash{p_{\psi}^{1}(\theta|x_{\mathrm{obs}}) = \tilde{p}_{\psi}^{1}(\theta|x_{\mathrm{obs}})}$. This is because the proposal prior in the $\smash{1^{\mathrm{st}}}$ round is equal to the prior, and thus the proposal posterior in the $1^{\mathrm{st}}$ round coincides with the posterior. Thus, in the $\smash{2^{\mathrm{nd}}}$ round, the proposal prior is given by a mixture of the prior and the proposal posterior estimate obtained in the $\smash{1^{\mathrm{st}}}$ round:
\begin{align}
    \tilde{p}^{2}(\theta)&=\frac{1}{2}\left[p^{0}_{\psi}(\theta|x_{\mathrm{obs}}) + p^{1}_{\psi}(\theta|x_{\mathrm{obs}})\right] \\
    &=\frac{1}{2}\left[p(\theta) + \tilde{p}_{\psi}^{1}(\theta|x_{\mathrm{obs}})\right]. \label{eq:2nd-round}
\end{align}
At the start of $3^{\mathrm{rd}}$ round, the current posterior estimate is equal to the proposal posterior estimate obtained in the $2^{\mathrm{nd}}$ round, reweighted  by the ratio between the prior and the proposal prior in the $2^{\mathrm{nd}}$ round: \begin{equation}
p_{\psi}^2(\theta|x_{\mathrm{obs}}) = \frac{1}{Z_{\psi}^{2}} \frac{p(\theta)}{\tilde{p}^{2}(\theta)}\tilde{p}_{\psi}^{2}(\theta|x_{\mathrm{obs}}),
\end{equation}
where $\smash{Z_{\psi}^{2}}$ is an (intractable) normalising constant given by $\smash{Z_{\psi}^{2}={\tilde{p}^{2}(x_{\mathrm{obs}})}/{p(x_{\mathrm{obs}})}}$, where $\smash{p(x_{\mathrm{obs}}) = \int_{\mathbb{R}^d} p(\theta) p(x_{\mathrm{obs}}|\theta)\mathrm{d}\theta}$ and $\smash{\tilde{p}^{2}(x_{\mathrm{obs}}) = \int_{\mathbb{R}^d} \tilde{p}^{2}(\theta) p(x_{\mathrm{obs}}|\theta)\mathrm{d}\theta}$. Thus, in the $3^{\mathrm{rd}}$ round, the proposal prior is given by a mixture of the prior, the proposal posterior estimate obtained in the $\smash{1^{\mathrm{st}}}$ round, and the posterior estimate obtained in the $\smash{2^{\mathrm{nd}}}$ round:
\begin{align}
    \tilde{p}^{3}(\theta) &= \frac{1}{3}\left[p^{0}_{\psi}(\theta|x_{\mathrm{obs}}) + p^{1}_{\psi}(\theta|x_{\mathrm{obs}})+p^{2}_{\psi}(\theta|x_{\mathrm{obs}})\right] \\
    &= \frac{1}{3}\left[p(\theta) + \tilde{p}_{\psi}^{1}(\theta|x_{\mathrm{obs}}) + \frac{1}{Z_{\psi}^2}\frac{p(\theta)}{\frac{1}{2}\left[p(\theta) + \tilde{p}_{\psi}^{1}(\theta|x_{\mathrm{obs}})\right]}\tilde{p}_{\psi}^{2}(\theta|x_{\mathrm{obs}})\right]. \label{eq:3rd-round}
\end{align}
Crucially, this proposal not only depends on the prior $p(\theta)$ and the proposal posterior estimates $\smash{\tilde{p}_{\psi}^{1}(\theta|x_{\mathrm{obs}}),\tilde{p}_{\psi}^{2}(\theta|x_{\mathrm{obs}}),\dots}$, but also on an intractable normalising constant $\smash{Z_{\psi}^{2}}$. Thus, without additional approximations, we cannot use this approach in the $3^{\mathrm{rd}}$ round, or any future rounds.

\textbf{Approximating the Proposal Prior}. An alternative approach is to approximate the proposal prior, or each component of the proposal prior, using samples. In this case, we replace the proposal prior $\tilde{p}^{r}(\theta)$ in \eqref{eq:weights} by an approximate proposal prior $\hat{{p}}^{r}(\theta)\approx \tilde{p}^{r}(\theta)$, which we obtain using samples $\theta\sim \tilde{p}^{r}(\theta)$. The advantages of this approach are that (a) it can be applied if we use more than 2 rounds and (b) we no longer need to use the probability flow ODE \eqref{eq:backwardODE} and the instantaneous change-of-variables formula \eqref{eq:change_of_var} to compute densities.

To see this, let us once again consider the quantities (i.e., the proposal priors) required to compute the importance weights in rounds $r=1,2,3$. At the start of the $1^{\mathrm{st}}$ round, the current `posterior estimate' is once again defined to be equal to the prior. Thus, the proposal prior $\tilde{p}^{1}(\theta)$ is equal to the prior $p(\theta)$, as in \eqref{eq:1st-round}. In this case, we can just set the approximate proposal prior equal to the original proposal prior: 
\begin{equation}
    \hat{p}^{1}(\theta):=\tilde{p}^{1}(\theta).
\end{equation}
At the start of the $2^{\mathrm{nd}}$ round, the current posterior estimate is equal to the proposal posterior estimate from the $1^{\mathrm{st}}$ round, as argued after \eqref{eq:1st-round}. Thus, as before, the proposal prior  is equal to a mixture of the prior $p(\theta)$ and the proposal posterior estimate $\tilde{p}_{\psi}^{1}(\theta|x_{\mathrm{obs}})$ 
 from the $1^{\mathrm{st}}$ round:
 \begin{equation}
 \tilde{p}^{2}(\theta) = \frac{1}{2}[p(\theta) + \tilde{p}_{\psi}^{1}(\theta|x_{\mathrm{obs}})]
 \end{equation}
 Unlike before, suppose now that we can generate samples from $\smash{\tilde{p}_{\psi}^{1}(\theta|x_{\mathrm{obs}})}$ and thus from $\smash{\tilde{p}^{2}(\theta) =  \frac{1}{2}[p(\theta) + \tilde{p}_{\psi}^{1}(\theta|x_{\mathrm{obs}})]}$, but that we cannot explicitly evaluate $\smash{\tilde{p}_{\psi}^{1}(\theta|x_{\mathrm{obs}})}$, nor $\smash{\tilde{p}^{2}(\theta) =  \frac{1}{2}[p(\theta) + \tilde{p}_{\psi}^{1}(\theta|x_{\mathrm{obs}})]}$. For example, it may be too expensive to solve the instantaneous change-of-variables formula \eqref{eq:change_of_var} to evaluate these densities. We then have two natural options for approximating $\tilde{p}^{2}(\theta)$. The first is to directly approximate $\hat{p}^{2}(\theta)\approx \tilde{p}^{2}(\theta)$ using samples $\theta\sim \tilde{p}^{2}(\theta)$. The second is to approximate $\hat{p}_{\psi}^{1}(\theta|x_{\mathrm{obs}})\approx \tilde{p}_{\psi}^{1}(\theta|x_{\mathrm{obs}})$ using samples $\smash{\theta\sim \tilde{p}^{1}_{\psi}(\theta|x_{\mathrm{obs}})}$, and then to approximate $\smash{\tilde{p}^{2}(\theta)}$ using
\begin{equation}
    \hat{p}^{2}(\theta):= \frac{1}{2}\left[ p(\theta) + \hat{p}_{\psi}^{1}(\theta|x_{\mathrm{obs}})\right]. \label{eq:p-hat-2-approx}
\end{equation}
At the start of the $3^{\mathrm{rd}}$ round, the current posterior estimate is equal to the approximate proposal posterior estimate from the $2^{\mathrm{nd}}$ round, reweighted  by the ratio between the prior and the approximate proposal prior from the $2^{\mathrm{nd}}$ round:
\begin{equation}
    p_{\psi}^{2}(\theta|x_{\mathrm{obs}}) 
    = \frac{1}{\hat{Z}_{\psi}^{2}} \frac{p(\theta)}{\hat{p}^{2}(\theta)} \hat{p}^{2}_{\psi}(\theta|x_{\mathrm{obs}}), \label{eq:2nd-round-approx}
\end{equation}
where  $\smash{\hat{p}_{\psi}^{2}(\theta|x_{\mathrm{obs}})}$ is the approximation of the proposal posterior $\smash{\hat{p}^{2}(\theta|x_{\mathrm{obs}})\propto \hat{p}^{2}(\theta)p(x_{\mathrm{obs}}|\theta)}$ associated with the approximate proposal prior $\smash{\hat{p}^2(\theta)}$, and $\smash{\hat{Z}_{\psi}^2}$ is the appropriate normalising constant. Explicitly, we now have $\smash{\hat{Z}_{\psi}^{2}=\frac{\hat{p}^{2}(x_{\mathrm{obs}})}{p(x_{\mathrm{obs}})}}$, where $\smash{p(x_{\mathrm{obs}}) = \int_{\mathbb{R}^d} p(\theta) p(x_{\mathrm{obs}}|\theta)\mathrm{d}\theta}$ and $\smash{\hat{p}^{2}(x_{\mathrm{obs}}) = \int_{\mathbb{R}^d} \hat{p}^{2}(\theta) p(x_{\mathrm{obs}}|\theta)\mathrm{d}\theta}$. Thus, in the $3^{\mathrm{rd}}$ round, the proposal prior $\smash{\tilde{p}^{3}(\theta)}$ is equal to a mixture of the prior, the approximate proposal posterior estimate obtained in  $1^{\mathrm{st}}$ round, and the approximate posterior estimate in \eqref{eq:2nd-round-approx} obtained in the $2^{\mathrm{nd}}$ round. That is, the mixture defined in \eqref{eq:3rd-round}, but now with $\tilde{p}(\cdot)$ replaced everywhere by $\hat{p}(\cdot)$. 

We cannot directly evaluate $\smash{{p}_{\psi}^{2}(\theta|x_{\mathrm{obs}})}$ in \eqref{eq:2nd-round-approx} due to the intractable normalising constant $\smash{\hat{Z}_{\psi}^2}$. We can, however, still generate samples $\smash{\theta\sim p_{\psi}^{2}(\theta|x_{\mathrm{obs}})}$ using, e.g., SIR, since this only requires that we can sample from $\smash{\hat{p}_{\psi}^{2}(\theta|x_{\mathrm{obs}})}$, and that we can evaluate both $\smash{p(\theta)}$ and $\smash{\hat{p}^{2}(\theta)}$. Thus, by construction, we can also generate samples from the proposal prior, $\smash{\theta\sim \tilde{p}^{3}(\theta)}$. Similar to before, we can then approximate $\tilde{p}^{3}(\theta)$ using samples, either directly or by estimating the mixture components individually.

For subsequent rounds, we can proceed in precisely the same fashion. In particular, first generate samples from the current estimate of the proposal posterior. Then use SIR to obtain samples from the current posterior estimate. Finally, use these samples to approximate the current posterior estimate, and use this approximation to approximate the next proposal prior.

The disadvantage of this approach is that it necessitates additional approximations in each round. This being said, perhaps somewhat surprisingly, in other contexts the use of approximate importance weights rather than exact importance weights can actually improve performance \citep{Henmi2007, Delyon2016}; see also the discussion in \citet{Liu2017b}. Regarding the proposal prior approximations, there are several possibilities: e.g., a kernel density estimator \citep{Delyon2016}, or a normalising flow \citep{Papamakarios2021}. We leave a more thorough investigation of this approach to future work.

\textbf{Approximating the Importance Weights}. One final option is to approximate the density ratio in \eqref{eq:weights} using samples, rather than just approximating the proposal prior. 
This is the subject of (two sample) {density ratio estimation} (DRE) \cite{Sugiyama2012}. There are various approaches to two-sample DRE, amongst others, moment matching \cite{Gretton2009}, probabilistic classification \cite{Qin1998,Cheng2004,Bickel2007}, and ratio matching \cite{Sugiyama2008,Kanamori2009,Tsuboi2009,Yamada2009}. In our case, not only do we have access to samples from the prior, but we can also evaluate the density. In this setting, Stein density ratio estimation (SDRE) provides an alternative approach \cite{Liu2019b}. Once again, we leave a more detailed investigation into this approach to future work.

\subsection{SNPSE-B}
Next, we provide additional details on SNPSE-B, which can be seen as the score-based analogue of SNPE-B \cite{Lueckmann2017}. This algorithm is summarised in Algorithm \ref{alg:snpse-b}.
\label{sec:snpse-b}
\subsubsection{Overview}
The main steps involved in SNPSE-B are as follows. For $r=1$, sample parameters from the prior $\{\theta_{0,i}^{1}\}_{i=1}^M \sim p(\theta):= p_{\psi}^{0}(\cdot | x_{\mathrm{obs}})$. Then, for all $r\geq 1$, 
\begin{itemize}
\item[(i)]  
    Simulate new data $\smash{\{x_i^{r}\}_{i=1}^{M} \sim p(\cdot|\theta_{0,i}^{r})}$. Concatenate samples $\smash{\{(\theta_{0,i}^{r},x_i^{r})\}_{i=1}^M}$ with those from previous rounds to form $\smash{\{(\theta_{0,i},x_i)\}_{i=1}^{rM}:=\bigcup_{s=1}^{r}\{(\theta_{0,i}^{r}, x_i^{r})\}_{i=1}^{M} \sim \tilde{p}^{r}(\theta)p(x|\theta)}$, where $\smash{\tilde{p}^{r}(\theta) = \frac{1}{r}\sum_{s=0}^{r-1} p_{\psi}^{s}(\theta|x_\mathrm{obs})}$. Draw times $\{t_i\}_{i=1}^{rM}\sim\mathcal{U}(0,T)$, and samples $\{\theta_{t_i,i}\}_{i=1}^{rM} \sim p_{t|0}(\cdot|\theta_{0,i})$.
    \item[(ii)] Using these samples, train a time-varying score network ${s}_{\psi}(\theta_t,x,t)$ to approximate the score of the posterior $\nabla_{\theta}\log {p}_t(\theta_t|x)$, by minimising a Monte Carlo estimate of 
\begin{align}
{\mathcal{J}}^{\mathrm{DSM-B}}_{\mathrm{post}}(\psi) = \frac{1}{2}\int_{0}^T\lambda_t &\mathbb{E}_{p_{t|0}({\theta}_t|\theta_0)p(x|\theta_0)\tilde{p}^{r}(\theta_0)}  \left[ \frac{p(\theta_0)}{\tilde{p}^{r}(\theta_0)}\left|\left| {s}_{{\psi}}({\theta}_t,x,t) - \nabla_{{\theta}_t} \log p_{t|0}({\theta}_t|\theta_0)\right|\right|^2\right]\mathrm{d}t . \label{eq:snpse-b} 
\end{align}
    
\item[(iii)] Draw samples $\{\theta^{r+1}_{T,i}\}_{i=1}^{M}\sim \pi(\theta)$. Simulate the backward SDE \eqref{eq:backwardSDE} or the probability flow ODE \eqref{eq:backwardODE}, substituting ${s}_{\psi}(\theta_t,x_{\mathrm{obs}},t)\approx\nabla_{\theta}\log {p}_t(\theta_t|x_{\mathrm{obs}})$, to obtain samples $\smash{\{{\theta}^{r+1}_{0,i}\}_{i=1}^{M}\sim {{p}}_{\psi}^{r}(\cdot|x_{\mathrm{obs}})}\approx p(\cdot|x_{\mathrm{obs}})$. 
\end{itemize}

\begin{algorithm}[h!]
\caption{SNPSE-B}
\label{alg:snpse-b}
\begin{algorithmic}
    \STATE \textbf{Inputs:} Observation $x_\mathrm{obs}$, prior $p(\theta) =: {p}_{\psi}^{0}(\theta|x_\mathrm{obs})$, simulator $p(x|\theta)$, simulation budget $N$, number of rounds $R$, (simulations-per-round $M=N/R$)
    \STATE \textbf{Outputs:} $p_{\psi}(\theta|x_\mathrm{obs}) \approx p(\theta|x_\mathrm{obs})$
    \FOR{$r = 1,\dots,R$}
        \FOR{$i = 1,\dots,M$}
            \STATE Draw $\theta_i \sim {p}_{\psi}^{r-1}(\theta|x_\mathrm{obs})$, $x_i \sim p(x|\theta_{i})$ 
            \STATE Add $(\theta_i, x_i)$ to $\mathcal{D}$
        \ENDFOR
        \STATE Learn $s_{\psi}(\theta_t,x,t)\approx \nabla_{\theta}\log p_t(\theta_t|x)$ by minimising a Monte Carlo estimate of \eqref{eq:snpse-b} based on dataset $\mathcal{D}$ (requires importance weights, see Appendix \ref{ap:iw-snpse-b} for details).
        \STATE Get $p_\psi^{r}(\theta|x_{\mathrm{obs}})$ sampler by substituting ${s}_{\psi}(\theta_t,x_{\mathrm{obs}},t)\approx \nabla_{\theta} \log p_t(\theta_t|x_{\mathrm{obs}})$ into \eqref{eq:backwardSDE} or \eqref{eq:backwardODE}.
    \ENDFOR
    \STATE \textbf{Return:} $p_{\psi}^{R}(\theta|x_\mathrm{obs})$
\end{algorithmic}
\end{algorithm}

\subsubsection{Theoretical Justification}
The theoretical justification for SNPSE-B is rather straightforward. In particular, observe that
\begin{align}
{\mathcal{J}}^{\mathrm{DSM-B}}_{\mathrm{post}}(\psi) &= \frac{1}{2}\int_{0}^T\lambda_t \mathbb{E}_{p_{t|0}({\theta}_t|\theta_0)p(x|\theta_0)\tilde{p}^{r}(\theta_0)}  \left[ \frac{p(\theta_0)}{\tilde{p}^{r}(\theta_0)}\left|\left| {s}_{{\psi}}({\theta}_t,x,t) - \nabla_{{\theta}_t} \log p_{t|0}({\theta}_t|\theta_0)\right|\right|^2\right]\mathrm{d}t \\
&= \frac{1}{2}\int_{0}^T\lambda_t \mathbb{E}_{p_{t|0}({\theta}_t|\theta_0)p(x|\theta_0){p}(\theta_0)}  \left[ \left|\left| {s}_{{\psi}}({\theta}_t,x,t) - \nabla_{{\theta}_t} \log p_{t|0}({\theta}_t|\theta_0)\right|\right|^2\right]\mathrm{d}t.
\end{align}
This is nothing more than the original denoising posterior score matching objective in \eqref{eq:weightedposteriorDSM}, which we know is minimised by $\psi^{*}$ such that $s_{\psi^{*}}(\theta_t,x,t)=\nabla_{\theta}\log p_t(\theta_t|x)$ \citep[e.g.,][]{Batzolis2021}. Thus, substituting $\smash{{s}_{\psi}(\theta_t,x_{\mathrm{obs}},t)\approx \nabla_{\theta} \log p_t(\theta_t|x_{\mathrm{obs}})}$ into the backward SDE \eqref{eq:backwardSDE} or the probability flow ODE \eqref{eq:backwardODE}, will indeed result in samples approximately distributed according to $p(\theta|x_{\mathrm{obs}})$, with no need for an additional correction.

\subsubsection{Computing the Importance Weights} \label{ap:iw-snpse-b}

Similar to SNPSE-A, SNPSE-B relies on our ability to compute or approximate the importance weights ${p(\theta)}/{\tilde{p}^{r}(\theta)}$, which now appear in the objective function in \eqref{eq:snpse-b}. We refer to Appendix \ref{ap:iw-snpse-a} for a detailed discussion of how to compute or approximate these weights.

\subsection{SNPSE-C}
\label{sec:snpse-c}
Finally, we provide additional details regarding SNPSE-C, which can be seen as the score-based analogue of SNPE-C \cite{Greenberg2019}. This algorithm is summarised in Algorithm \ref{alg:snpse-c}.

\subsubsection{Overview}
\label{sec:snpse-c-overview}
The main components of the SNPSE-C algorithm are summarised below. For $r=1$, sample parameters from the prior $\{\theta_{0,i}^{1}\}_{i=1}^M\sim p(\theta):= p_{\psi}^{0}(\cdot|x_{\mathrm{obs}})$. For all $r\geq 1$:
\begin{itemize}
    \item[(i)] Simulate new data $\smash{\{x_i^{r}\}_{i=1}^{M} \sim p(\cdot|\theta_{0,i}^{r})}$. Concatenate samples $\smash{\{(\theta_{0,i}^{r},x_i^{r})\}_{i=1}^M}$ with those from previous rounds to form $\smash{\{(\theta_{0,i},x_i)\}_{i=1}^{rM}:=\bigcup_{s=1}^{r}\{(\theta_{0,i}^{r}, x_i^{r})\}_{i=1}^{M} \sim \tilde{p}^{r}(\theta)p(x|\theta)}$, where $\smash{\tilde{p}^{r}(\theta) = \frac{1}{r}\sum_{s=0}^{r-1} p_{\psi}^{s}(\theta|x_\mathrm{obs})}$. Draw times $\{t_i\}_{i=1}^{rM}\sim\mathcal{U}(0,T)$, and samples $\{\theta_{t_i,i}\}_{i=1}^{rM} \sim p_{t|0}(\cdot|\theta_{0,i})$.
    \item[(ii)] Using these samples, train a time-varying score network $\tilde{s}^{r}_{\psi}(\theta_t,x,t)$ to approximate the score of the proposal posterior $\nabla_{\theta}\log \tilde{p}_t^{r}(\theta_t|x)$, by minimising 
    a Monte Carlo estimate of 
    \begin{align}
{\mathcal{J}}^{\mathrm{DSM-C}}_{\mathrm{post}}(\psi) =
\frac{1}{2}\int_{0}^T\lambda_t &\mathbb{E}_{p_{t|0}({\theta}_t|\theta_0)p(x|\theta_0)\tilde{p}^{r}(\theta_0)}  \left[ \left|\left| \tilde{s}^{r}_{{\psi}}({\theta}_t,x,t) - \nabla_{{\theta}_t} \log p_{t|0}({\theta}_t|\theta_0)\right|\right|^2\right]\mathrm{d}t , 
\label{eq:snpse-c} 
\end{align}
where the  score network  $\tilde{s}^{r}_{\psi}(\theta_t,x,t)$ is defined as
\begin{align}
    \tilde{s}^{r}_{\psi}(\theta_t,x,t) &= s_{\psi}(\theta_t,x,t) + \nabla_{\theta} \log \tilde{p}^{r}_t(\theta_t) - \nabla_{\theta} 
    \log {p}_t(\theta_t). \label{eq:snpse_score_identity}
\end{align}
or some approximation thereof (see Appendix \ref{app:est-prop-prior-score}).
    \item[(iii)] Draw samples $\{\theta^{r+1}_{T,i}\}_{i=1}^{M}\sim \pi(\theta)$. Simulate the backward SDE \eqref{eq:backwardSDE} or the probability flow ODE \eqref{eq:backwardODE}, substituting $\smash{{s}_{\psi}(\theta_t,x_{\mathrm{obs}},t)}$ for $\smash{\nabla_{\theta}\log {p}_t(\theta_t|x_{\mathrm{obs}})}$, to obtain samples $\smash{\{{\theta}^{r+1}_{0,i}\}_{i=1}^{M}\sim {{p}}_{\psi}^{r}(\cdot|x_{\mathrm{obs}})}\approx p(\cdot|x_{\mathrm{obs}})$.
\end{itemize}

\begin{algorithm}[h!]
\caption{SNPSE-C}
\label{alg:snpse-c}
\begin{algorithmic}
    \STATE \textbf{Inputs:} Observation $x_\mathrm{obs}$, prior $p(\theta) =: {p}_{\psi}^{0}(\theta|x_\mathrm{obs})$, simulator $p(x|\theta)$, simulation budget $N$, number of rounds $R$, (simulations-per-round $M=N/R$)
    \STATE \textbf{Outputs:} $p_{\psi}(\theta|x_\mathrm{obs}) \approx p(\theta|x_\mathrm{obs})$
    \STATE Compute  $\nabla_{\theta}\log p_t(\theta_t)$ or estimate $s^{r}_{\psi}(\theta_t,t)\approx \nabla_{\theta} \log p_t(\theta_t)$ (see Algorithm \ref{alg:1}).
    \FOR{$r = 1,\dots,R$}
        \FOR{$i = 1,\dots,M$}
            \STATE Draw $\theta_i \sim p_{\psi}^{r-1}(\theta|x_\mathrm{obs})$, $x_i \sim p(x|\theta_{i})$ 
            \STATE Add $(\theta_i, x_i)$ to $\mathcal{D}$
        \ENDFOR
        \IF{$r > 1$}
            \STATE Learn $s_{\varphi}^{r,\mathrm{prop}}(\theta_t,t)  \approx \nabla_{\theta} \log \tilde{p}_{t}^{r}(\theta_t)$ by minimising \eqref{eq:proposal_score_dsm}.
            \STATE Define $\tilde{s}_{\psi}^{r}(\theta_t,x,t) := s_{\psi}(\theta_t,x,t) + s_{\varphi}^{r,\mathrm{prop}}(\theta_t,t) - \nabla_{\theta}\log p_t(\theta_t)$
        \ELSE
            \STATE Define $\tilde{s}^{r}_{\psi}(\theta_t,x,t) := s_{\psi}(\theta_t,x,t)$
        \ENDIF
        \STATE Learn $\tilde{s}^{r}_{\psi}(\theta_t,x,t)\approx \nabla_{\theta}\log \tilde{p}^{r}_{t}(\theta_t|x)$ by minimising a Monte Carlo estimate of \eqref{eq:snpse-c} based on dataset $\mathcal{D}$
        \STATE Get $p_\psi^{r}(\theta|x_{\mathrm{obs}})$ sampler by substituting $\smash{s_{\psi}(\theta_t,x_{\mathrm{obs}},t)\approx \nabla_{\theta} \log p_t(\theta_t|x_{\mathrm{obs}})}$ into \eqref{eq:backwardSDE} or \eqref{eq:backwardODE}. 
    \ENDFOR
    \STATE \textbf{Return:} $p_{\psi}^{R}(\theta|x_\mathrm{obs})$
\end{algorithmic}
\end{algorithm}

\subsubsection{Theoretical Justification}
We now outline the theoretical justification for SNPSE-C. We begin by noting that, via a repeated application of Bayes' Theorem, we have
\begin{align}
    p_t(\theta_t|x) &= \int p_{t|0}(\theta_t|\theta_0)p(\theta_0|x)\mathrm{d}\theta_0
    = \frac{p_t(\theta_t) \int_{\mathbb{R}^d} p_{0|t}(\theta_0|\theta_t)p(x|\theta_0) \mathrm{d}\theta_0}{p(x)} 
    = \frac{p_t(\theta_t)p_t(x|\theta_t)}
    {p(x)}, \label{eq:bayes1} \\
\tilde{p}^{r}_t(\theta_t|x) &= \int p_{t|0}(\theta_t|\theta_0)\tilde{p}^{r}(\theta_0|x)\mathrm{d}\theta_0
    = \frac{\tilde{p}_t^{r}(\theta_t) \int_{\mathbb{R}^d} \tilde{p}^{r}_{0|t}(\theta_0|\theta_t){p}(x|\theta_0) \mathrm{d}\theta_0}{\tilde{p}(x)} 
    = \frac{\tilde{p}_t^{r}(\theta_t)\tilde{p}^{r}_t(x|\theta_t)}
    {\tilde{p}(x)}, \label{eq:bayes2}
\end{align}
where in the final equality in each line we have identified $p_t(x|\theta_t) = \int p(x|\theta_0)p_{0|t}(\theta_0|\theta_t)\mathrm{d}\theta_0$ and $\tilde{p}^{r}_t(x|\theta_t) = \int p(x|\theta_0)\tilde{p}^{r}_{0|t}(\theta_0|\theta_t)\mathrm{d}\theta_0$. It follows, taking logarithmic derivatives of \eqref{eq:bayes1} and \eqref{eq:bayes2}, that
\begin{align}
    \nabla_{\theta} \log {p}_t(\theta_t|x) &= \nabla_{\theta} \log {p}_t(x|\theta_t) + \nabla_{\theta}\log {p}_t(\theta_t), \label{eq:score_id_1} \\
    \nabla_{\theta} \log \tilde{p}^{r}_t(\theta_t|x) &= \nabla_{\theta} \log \tilde{p}^{r}_t(x|\theta_t) + \nabla_{\theta}\log \tilde{p}^{r}_t(\theta_t).
    \label{eq:score_id_2}
\end{align}
Thus, in particular, we can relate the perturbed posterior score $\nabla_{\theta} \log p_t(\theta_t|x)$ to the perturbed proposal posterior $\nabla_{\theta} \log \tilde{p}^{r}_t(\theta_t|x)$ score according to 
\begin{equation}
    \nabla_{\theta}\log \tilde{p}^{r}_t(\theta_t|x)=\nabla_{\theta}\log p_t(\theta_t|x) + \nabla_{\theta}\log 
 \tilde{p}^{r}_t(\theta_t)  - \nabla_{\theta}\log p_t(\theta_t) + \nabla_{\theta}\log \tilde{p}^{r}_t(x|\theta_t) - \nabla_{\theta}\log  p_t(x|\theta_t), \label{eq:score-id-c}
\end{equation}
Now, suppose that we define $\tilde{s}_{\psi}^{r}(\theta_t,x,t)$ according to 
\begin{align}
    \tilde{s}^{r}_{\psi}(\theta_t,x,t)=s_{\psi}(\theta_t,x,t) + \nabla_{\theta}\log 
 \tilde{p}^{r}_t(\theta_t)  - \nabla_{\theta}\log p_t(\theta_t) + \nabla_{\theta}\log \tilde{p}_t^{r}(x|\theta_t)-  \nabla_{\theta}\log  p_t(x|\theta_t).  \label{eq:new-score-c}
\end{align}
By standard results on conditional denoising score matching \citep[e.g.,][]{Batzolis2021}, we know that \eqref{eq:snpse-c} is minimised by $\psi^{*}$ such that $\tilde{s}_{\psi^{*}}^{r}(\theta,x,t) = \nabla_{\theta} \log \tilde{p}_t^{r}(\theta|x)$ for almost every $\theta\in\mathbb{R}^d$, $x\in\mathbb{R}^p$, and $t\in[0,T]$. It follows, using this observation,  the definition in \eqref{eq:new-score-c}, and the identity in \eqref{eq:score-id-c},  that the minimiser $\smash{\psi^{*} = \mathrm{argmin}{\mathcal{J}}^{\mathrm{DSM}-C}_{\mathrm{post}}(\psi)}$ of \eqref{eq:snpse-c} is such that 
\begin{align}
    s_{\psi^{*}}(\theta,x,t) &= \tilde{s}^{r}_{\psi^{*}}(\theta_t,x,t) -  \nabla_{\theta}\log 
 \tilde{p}^{r}_t(\theta_t)  + \nabla_{\theta}\log p_t(\theta_t) - \nabla_{\theta}\log \tilde{p}_t^{r}(x|\theta_t) +  \nabla_{\theta}\log  p_t(x|\theta_t) \\
 &=\nabla_{\theta} \log \tilde{p}_t^{r}(\theta_t|x)-  \nabla_{\theta}\log 
 \tilde{p}^{r}_t(\theta_t)  + \nabla_{\theta}\log p_t(\theta_t) - \nabla_{\theta}\log \tilde{p}_t^{r}(x|\theta_t) +  \nabla_{\theta}\log  p_t(x|\theta_t) \\[1mm]
 &= \nabla_{\theta} \log p_t(\theta_t|x).
\end{align}
More generally, if we minimise \eqref{eq:snpse-c} to obtain $\tilde{s}_{\psi}^{r}(\theta_t,x,t)\approx \nabla_{\theta}\log \tilde{p}_t^{r}(\theta_t|x)$, then we automatically recover $s_{\psi}(\theta_t,x,t)\approx \nabla_{\theta} \log p_t(\theta_t|x)$ via the definition \eqref{eq:new-score-c}. That is, we can automatically transform an estimate of the proposal posterior score into an estimate of the posterior score. This approach is reminiscent of SNPE-C \cite{Greenberg2019}, also referred to as automatic posterior transformation (APT).

In practice, the definition in \eqref{eq:new-score-c} relies on several quantities - namely $\nabla_{\theta}\log \tilde{p}_t^{r}(\theta_t)$, $\nabla_{\theta} \log \tilde{p}_t^{r}(x|\theta_t)$ and $\nabla_{\theta} \log p_t(x|\theta_t)$ - to which we do not have immediate access. To make any progress, we will therefore need to make several simplifications and approximations. The first simplification is based on the observation that  $\nabla_{\theta} \log p_0(x|\theta_0) = \nabla_{\theta} \log \tilde{p}_0(x|\theta_0)$ and $\nabla_{\theta} \log p_T(x|\theta_T) \approx \nabla_{\theta} \log \tilde{p}_T(x|\theta_T)$. Thus, substituting into \eqref{eq:score-id-c}, we have 
\begin{align}
    \nabla_{\theta}\log \tilde{p}^{r}_{0}(\theta_{0}|x)&=\nabla_{\theta}\log p_{0}(\theta_{0}|x) + \nabla_{\theta}\log 
 \tilde{p}^{r}_{0}(\theta_0)  - \nabla_{\theta}\log p_0(\theta_0), \label{eq:snpse-c-zero}\\
 \nabla_{\theta}\log \tilde{p}^{r}_T(\theta_T|x)&\approx\nabla_{\theta}\log p_T(\theta_T|x) + \nabla_{\theta}\log 
 \tilde{p}^{r}_T(\theta_T)  - \nabla_{\theta}\log p_T(\theta_T).  \label{eq:snpse-c-T}
\end{align}
Inspired by \eqref{eq:snpse-c-zero} - \eqref{eq:snpse-c-T}, suppose that we define a sequence of distributions $\{p_t^{r,\mathrm{seq}}(\theta_t|x)\}_{t\in[0,T]}$ according to $p_t^{r,\mathrm{seq}}(\theta_t|x) \propto \frac{p_t(\theta_t)}{\tilde{p}_t^{r}(\theta_t)} \tilde{p}_t^{r}(\theta_t|x) \propto \frac{\tilde{p}_t^{r}(x|\theta_t)}{p_t(x|\theta_t)}p_t(\theta_t|x)$. By construction, the scores of this sequence of distributions satisfy the identity
\begin{equation}
    \nabla_{\theta} \log \tilde{p}_t^{r}(\theta_t|x) = \nabla_{\theta} \log p_t^{r,\mathrm{seq}}(\theta_t|x) + \nabla_{\theta} \log \tilde{p}_t^{r}(\theta_t) - \nabla_{\theta} \log {p}_t(\theta_t). \label{eq:new-score}
\end{equation}
In addition, based on \eqref{eq:new-score}, suppose that we now redefine the score network $\tilde{s}_{\psi}^{r}(\theta_t,x,t)$ according to 
\begin{equation}
    \tilde{s}_{\psi}^{r}(\theta_t,x,t) = s_{\psi}(\theta_t,x,t) + \nabla_{\theta} \log \tilde{p}_t^{r}(\theta_t) - \nabla_{\theta} \log p_t(\theta_t).  \label{eq:new-def}
\end{equation}
Then, arguing similarly to before, but now using \eqref{eq:new-def} and \eqref{eq:new-score} in place of \eqref{eq:new-score-c} and \eqref{eq:score-id-c}, it follows that the minimiser $\smash{\psi^{*} = \mathrm{argmin}{\mathcal{J}}^{\mathrm{DSM}-C}_{\mathrm{post}}(\psi)}$ of \eqref{eq:snpse-c} is such that 
\begin{align}
    s_{\psi^{*}}(\theta,x,t) &= \tilde{s}^{r}_{\psi^{*}}(\theta_t,x,t) -  \nabla_{\theta}\log 
 \tilde{p}^{r}_t(\theta_t)  + \nabla_{\theta}\log p_t(\theta_t) \\
 & = \nabla_{\theta} \log \tilde{p}_t^{r}(\theta_t|x)-  \nabla_{\theta}\log 
 \tilde{p}^{r}_t(\theta_t)  + \nabla_{\theta}\log p_t(\theta_t) = \nabla_{\theta} \log p^{r,\mathrm{seq}}_t(\theta_t|x).
\end{align}
Thus, in general, if we minimise \eqref{eq:snpse-c} to obtain $\tilde{s}_{\psi}^{r}(\theta_t,x,t)\approx \nabla_{\theta}\log \tilde{p}_t^{r}(\theta_t|x)$, then we automatically recover $s_{\psi}(\theta_t,x,t)\approx \nabla_{\theta} \log p_t^{r,\mathrm{seq}}(\theta_t|x)$ via the definition in \eqref{eq:new-def}. Crucially, by our construction of $\{p_t^{r,\mathrm{seq}}(\theta|x)\}_{t\in[0,T]}$, we have 
\begin{align}
    p_0^{r,\mathrm{seq}}(\theta_0|x) &\propto \frac{p_0(\theta_0)}{\tilde{p}_0(\theta_0)}\tilde{p}_0^{r}(\theta_0|x) = \frac{p(\theta_0)}{\tilde{p}(\theta_0)} \tilde{p}(\theta_0|x) \propto p(\theta_0|x), \\
    p_T^{r,\mathrm{seq}}(\theta_T|x) &\propto \frac{p_T(\theta_T)}{\tilde{p}_T(\theta_T)}\tilde{p}_T^{r}(\theta_T|x) \approx \frac{\mathcal{N}(\theta_T;0, \mathbf{I})}{\mathcal{N}(\theta_T;0, \mathbf{I})} \mathcal{N}(\theta_T;0, \mathbf{I}) = \mathcal{N}(\theta_T; 0,\mathbf{I}). 
\end{align}
Thus, in particular, $\{p_t^{r,\mathrm{seq}}(\theta_t|x)\}_{t\in[0,T]}$ defines a sequence of distributions which smoothly interpolate between the true posterior $p_0(\theta|x) = p(\theta|x)$, and the reference distribution $ \mathcal{N}(\theta;0,\mathbf{I})$. In general, of course, $p_t^{r,\mathrm{seq}}(\theta_t|x)$ will not coincide with $p_t(\theta_t|x)$, which means the sequence $\{p_t^{r,\mathrm{seq}}(\theta_t|x)\}_{t\in[0,T]}$ does not correspond to the standard sequence $\{p_t(\theta_t|x)\}_{t\in[0,T]}$ of distributions obtained by applying the forward SDE \eqref{eq:forwardSDE}, as described in Section \ref{sec:SGM}. Thus, in particular, substituting the score network $s_{\psi}(\theta_t,x,t)\approx \nabla_{\theta} \log p_t^{r,\mathrm{seq}}(\theta_t,x,t)$ obtained via \eqref{eq:new-def} into the backward SDE \eqref{eq:backwardSDE} will not result in samples from the correct posterior distribution.

Nonetheless, based on our observation above that $\{p_t^{r,\mathrm{seq}}(\theta_t|x)\}_{t\in[0,T]}$  smoothly interpolate between the true posterior $p_0(\theta|x) = p(\theta|x)$, and the reference distribution $p_T(\theta|x) = \mathcal{N}(\theta;0,\mathbf{I})$, we can still use this score network to obtain samples from the correct posterior. In particular, by using $s_{\psi}(\theta_t,x,t)\approx \nabla_{\theta} \log p_t^{r,\mathrm{seq}}(\theta_t,x,t)$ within an annealed MCMC algorithm, e.g., annealed Langevin dynamics \citep{Song2019,Geffner2022,Du2023}, we should still recover samples from the correct posterior. See also \citet[Section 4]{Du2023} for a related discussion, albeit in a very different context.

\subsubsection{Estimating the Proposal Prior Score}
\label{app:est-prop-prior-score}
Clearly, in order to define the score network $\tilde{s}_{\psi}^{r}(\theta_t,x,t)$ according to \eqref{eq:snpse_score_identity}, we must be able to compute (or approximate)  score of the perturbed proposal prior, namely, 
\begin{equation}
    \nabla_{\theta} \log \tilde{p}_t^{r}(\theta_t): = \nabla_{\theta} \log \left[\frac{1}{r}\sum_{s=0}^{r-1}{p}^{s}_{\psi,t}(\theta_t|x_{\mathrm{obs}})\right], \label{eq:proposal_score}
\end{equation}
where $\tilde{p}_t^{r}(\theta_t) = \int_0^t p_{t|0}(\theta_t|\theta_0) \tilde{p}^{r}(\theta_0)\mathrm{d}\theta_0$, ${p}_{\psi,t}^{s}(\theta_t|x_{\mathrm{obs}}) = \int_{0}^t p_{t|0}(\theta_t|\theta_0) {p}_{\psi}^{s}(\theta_0|x_{\mathrm{obs}})\mathrm{d}\theta_0$; and $\tilde{p}^{r}(\theta)$ and ${p}_{\psi}^{s}(\theta|x_{\mathrm{obs}})$ are defined as in Section \ref{sec:snpse-c-overview}.
Unfortunately, 
the score of the perturbed proposal prior cannot be written as a mixture of the score of the perturbed proposal priors from each of the previous rounds:
\begin{align}
    \nabla_{\theta} \log \tilde{p}_t^{r}(\theta_t) &\neq \frac{1}{r}\sum_{s=0}^{r-1} \nabla_{\theta}\log {p}_{\psi,t}^{s}(\theta_t|x_{\mathrm{obs}})
\end{align}
That is, we cannot compute $\nabla_{\theta} \log \tilde{p}_t^{r}(\theta_t)$ in \eqref{eq:proposal_score} using the score estimates 
obtained in previous rounds. Instead, to compute \eqref{eq:proposal_score}, and thus to use \eqref{eq:snpse_score_identity}, we will need an alternative approach. Below, we outline several possibilities.

\textbf{Approximating the Proposal Prior Score}. The first and perhaps most natural approach is to approximate  $\smash{s_{\varphi}^{r,\mathrm{prop}}(\theta_t,t)\approx \nabla_{\theta}\log \tilde{p}_t^{r}(\theta_t)}$ using a score network, and substitute this approximation into \eqref{eq:snpse_score_identity}. Given samples $\theta\sim \tilde{p}^{r}(\theta)$, we can train this network by minimising a Monte Carlo estimate of the standard denoising score matching objective, viz, 
\begin{equation}
{\mathcal{J}}^{\mathrm{DSM-C}}_{\mathrm{prop}}(\varphi) = \frac{1}{2}\int_0^T \lambda_t \mathbb{E}_{p_{t|0}(\theta_t|\theta_0)\tilde{p}^{r}(\theta_0)} \left[ || s_{\varphi}^{r,\mathrm{prop}}(\theta_t,t) - \nabla_{\theta}\log p_{t|0}(\theta_t|\theta_0)||^2\right]\mathrm{d}t.
    \label{eq:proposal_score_dsm}
\end{equation}
There are several advantages of this approach: it is conceptually very simple, and it fits rather naturally into our existing framework. At the same time, there are several disadvantages, particularly with regards to computational and memory costs. 

First, to obtain a sufficiently accurate score network $\smash{s_{\varphi}^{r,\mathrm{prop}}(\theta_t,t)\approx \nabla_{\theta}\log \tilde{p}_t^{r}(\theta_t)}$, it is desirable to use a large number of samples $\theta\sim \tilde{p}^{r}(\theta)$ from the proposal prior to form the Monte Carlo estimate of \eqref{eq:proposal_score_dsm}. By definition of the proposal prior, this requires generating (and storing) a large number of samples $\theta\sim {p}_{\psi}^{s}(\theta|x_{\mathrm{obs}})$, for $s=1,\dots,r-1$, which are obtained by repeated simulation of the relevant backward SDE \eqref{eq:backwardSDE} or probability flow ODE \eqref{eq:backwardODE}. Second, minimising \eqref{eq:proposal_score_dsm} is itself a costly procedure, possibly requiring a large number of iterations in order to converge. The score network $s_{\varphi}^{r,\mathrm{prop}}(\theta_t,t)\approx \nabla_{\theta}\log \tilde{p}_t^{r}(\theta_t)$ must be re-learned after each round, which may amount to a significant additional computational cost over a large number of rounds.

Finally, the proposal prior score network $s_{\varphi}^{r,\mathrm{prop}}(\theta_t,t)\approx \nabla_{\theta}\log \tilde{p}_t^{r}(\theta_t)$ is substituted into \eqref{eq:snpse_score_identity} to compute $\tilde{s}_{\psi}^{r}(\theta_t,x,t)$, which is then learned by minimising the SNPSE-C objective in \eqref{eq:snpse-c}. Thus, minimising \eqref{eq:snpse-c} requires an additional neural network pass at each training iteration. In addition, any error in the approximation $s_{\varphi}^{r,\mathrm{prop}}(\theta_t,t)\approx \nabla_{\theta}\log \tilde{p}_t^{r}(\theta_t)$ will adversely affect learning an accurate approximation $\tilde{s}_{\psi}^{r}(\theta_t,x,t)\approx \nabla_{\theta} \log \tilde{p}_t^{r}(\theta_t|x)$. 

\textbf{Approximating the Density Ratio Score}. An alternative approach is instead to approximate the score of the ratio of the proposal prior density and the prior density, viz
\begin{equation}
    \nabla_{\theta} \log \left[ \frac{\tilde{p}_t^{r}(\theta_t)}{p_t(\theta_t)}\right] = \nabla_{\theta} \log \tilde{p}_t^{r}(\theta_t) - \nabla_{\theta} \log p_t(\theta_t). \label{eq:score-ratio}
\end{equation}
Inspired by the recent perspective in \citet{Liu2023} on Stein variational gradient descent \cite{Liu2016b}, we can approximate this term using a normalised, kernel-based estimate. In particular, let $k:\mathbb{R}^d\times\mathbb{R}^d\rightarrow\mathbb{R}$ be a positive semi-definite kernel. We can then approximate
\begin{align}
    \nabla_{\theta} \log \left[ \frac{\tilde{p}_t^{r}(\theta_t)}{p_t(\theta_t)}\right] &\approx \frac{\int_{\mathbb{R}^d} k(\theta_t,\theta_t') \nabla_{\theta'} \log \left[ \frac{\tilde{p}_t^{r}(\theta_t')}{p_t(\theta_t')}\right] \tilde{p}_t^{r} (\theta_t') \mathrm{d}\theta_t'}{\int_{\mathbb{R}^d}k(\theta_t,\theta_t')\tilde{p}_t^{r}(\theta_t')\mathrm{d}\theta_t'}  \label{eq:line1} \\
    &
    = - \frac{\int_{\mathbb{R}^d} \left[ \nabla_{\theta'} k(\theta_t,\theta_t') + \nabla_{\theta'} \log p_t(\theta_t')\right]\tilde{p}_t^{r}(\theta_t') \mathrm{d}\theta_t'}{{\int_{\mathbb{R}^d}k(\theta_t,\theta_t')\tilde{p}_t^{r}(\theta_t')\mathrm{d}\theta_t'}}  \\
    &= -\frac{\mathbb{E}_{p_{t|0}(\theta_t'|\theta_0')\tilde{p}^{r}(\theta_0')}\left[\nabla_{\theta'} k(\theta_t,\theta_t') + \nabla_{\theta'} \log p_t(\theta_t')\right]}{\mathbb{E}_{p_{t|0}(\theta_t'|\theta_0')\tilde{p}^{r}(\theta_0')}\left[k(\theta_t,\theta_t')\right]}, \label{eq:svgd} 
\end{align}
where the second line follows using integration by parts, and holds under mild regularity conditions \citep[e.g.,][]{Liu2016b,Korba2020}. Crucially, the expectations in \eqref{eq:svgd} only depend on samples $\theta'_{t}\sim p_{t|0}(\cdot|\theta'_0)$, where $\theta'_0\sim \tilde{p}^{r}(\cdot)$, and can therefore be approximated using Monte Carlo.

\textbf{Approximating the Proposal Prior}. One final approach, somewhat different in spirit from the previous two, is to learn an approximation $\hat{p}^{r}(\theta)\approx \tilde{p}^{r}(\theta)$ of the proposal prior, or else an approximation $\hat{p}^{s}_{\psi}(\theta|x_{\mathrm{obs}})\approx {p}_{\psi}^{s}(\theta|x_{\mathrm{obs}})$ of each component of the proposal prior, for which it is possible to exactly compute the score of the corresponding perturbed proposal prior
\begin{equation}
    \nabla_{\theta}\log \hat{p}_t^{r}(\theta_t) := \nabla_{\theta} \left[ \frac{1}{r} \sum_{s=0}^{r-1} \hat{p}_{\psi,t}^{s}(\theta_t|x_{\mathrm{obs}})\right].
\end{equation}
There are several possible choices of surrogate proposal prior for which this calculation is possible. These include a mixture of Gaussians, a (continuous) uniform distribution, and an atomic (i.e., discrete uniform) distribution \citep[e.g.,][]{Greenberg2019}. We refer to Appendix \ref{app:perturbed_prior} for further details of how to compute the perturbed prior score in each of these cases. We leave further investigation of this approach to future work.

\subsection{TSNPSE vs SNPSE-A vs SNPSE-B vs SNPSE-C}
\begin{figure*}[h!]
\centering
\includegraphics[width=.9\textwidth, trim=40 20 40 20, clip]{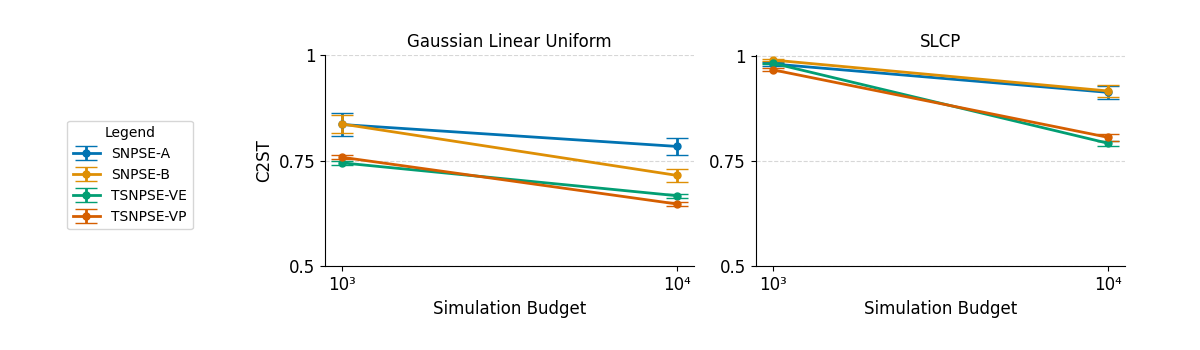}
  \vspace{-2mm}
\caption{\textbf{Comparison between SNPSE-A, SNPSE-B, and TSNPSE on two benchmark tasks.}}
  \label{fig:snpse-ab}
\end{figure*}

In Figure \ref{fig:snpse-ab}, we provide a comparison between TSNPSE, SNPSE-A, and SNPSE-B, for two of the benchmark tasks described in \citet{Lueckmann2021} (SLCP and GLU). We omit the corresponding results for SNPSE-C since, in our empirical testing, this method failed to provide meaningful results (e.g., C2ST $\approx$ 1). This, we suspect, is due to the significant approximation error incurred when estimating the score of the proposal prior, as described in Section \ref{app:est-prop-prior-score}. 

In both of these task, TSNPSE significantly outperforms both SNPSE-A and SNPSE-B, a finding which was also replicated in other tasks. We suspect that this is largely due to the error associated with the approximate importance weight correction used by SNPSE-A, and the high-variance updates associated with the use of importance weights in the loss function used by SNPSE-B. We note that the performance of SNPSE-B could likely be improved using the techniques recently introduced in  \citet{Xiong2023}.

\section{Dealing with Multiple Observations}
\label{app:LFI_multiple_obs}
In this section, we discuss how to adapt our methods to the task of generating samples from $p(\theta_t|x_\mathrm{obs}^{1},\dots,x_\mathrm{obs}^{n})$ for any set of observations $\{x_\mathrm{obs}^{1},\dots,x_\mathrm{obs}^{n}\}$. As noted in \citet{Geffner2022}, it is not possible to factorise the multiple-observation posterior score $\nabla_{\theta} \log p_t(\theta_t|x_{\mathrm{obs}}^{1},\dots,x_{\mathrm{obs}}^{n})$ in terms of the single-observation posterior scores $\nabla_{\theta} \log p_t(\theta_t|x_{\mathrm{obs}}^{i})$, and the prior score $\nabla_{\theta} \log p(\theta_t)$. Thus, a naive implementation of NPSE would require training a score network $s_{\psi}(\theta_t,x^{1},\dots,x^{n},t)\approx \nabla_{\theta} \log p_t(\theta_t|x^{1},\dots,x^{n})$ using samples $(\theta,x^{1},\dots,x^{n})\sim p(\theta)
\prod_{j=1}^n p(x^{i}|\theta)$. This requires calling the simulator $n$ times for every parameter sample $\theta$, and is thus highly sample inefficient.

To circumvent this issue, \citet{Geffner2022} introduce a new method based on the observation that $p(\theta|x^{1},\dots,x^{n}) \propto p(\theta)^{1-n} \prod_{i=1}^n p(\theta|x^{i})$. In particular, they propose to use the sequence of densities
\begin{equation}
    p^{\mathrm{(bridge)}}_t(\theta_t|x^{1},\dots,x^{n}) \propto (p(\theta_t)^{1-n})^{\frac{T-t}{T}}\prod_{i=1}^n p_t(\theta_t|x^{i}). \label{eq:geffner}
\end{equation}
Importantly, the density at $t=0$ coincides with the target distribution $p(\theta|x^{1},\dots,x^{n})$, while the density at $t=T$ is a tractable Gaussian. In addition, the score of these densities can be decomposed into the single-observation posterior scores $\nabla_{\theta} \log p_t(\cdot|x^{i})$, and the (known) prior score $\nabla_{\theta} \log p(\cdot)$, as 
\begin{equation}
    \nabla_{\theta} \log p^{\mathrm{(bridge)}}_t(\theta_t|x^1,\dots,x^n) = 
    \frac{(1-n)(T-t)}{T} \nabla_{\theta} \log p(\theta_t) + \sum_{i=1}^n \nabla_{\theta} \log p_t(\theta_t|x^i). \label{eq:geffner_scores}
\end{equation}
Thus, in particular, it is only necessary to learn a single score network $s_{\psi_{\text{post}}}(\theta_t,x,t)\approx \nabla_{\theta} \log p_t(\theta_t|x)$, which can be trained using samples $(\theta,x)\sim p(\theta)p(x|\theta)$. After learning this score network, one can then generate samples from the posterior by running the reverse diffusion with 
\begin{equation*}
    \nabla_{\theta} \log p^{\mathrm{(bridge)}}_t(\theta_t|x^1_{\text{obs}},\dots,x^n_{\text{obs}}) \approx \frac{(1-n)(T-t)}{T} \nabla_{\theta} \log p(\theta_t) + \sum_{i=1}^n s_{\psi}(\theta_t,x^i_{\text{obs}},t).
\end{equation*}
It is worth emphasising that, other than for $t=0$, $p^{\mathrm{(bridge)}}_t(\theta_t|x^{1},\dots,x^{n})$ do not coincide with the true perturbed multi-observation posterior densities $p_t(\theta_t|x^{1},\dots,x^{n})$. Thus, to generate samples, one must use an annealed MCMC algorithm \citep[e.g.,][Algorithm 1]{Geffner2022}, rather than directly integrating the reverse-time SDE. 

We now propose an alternative approach, based on a very similar idea to the one in \citet{Geffner2022}. In particular, in place of \eqref{eq:geffner}, we now propose the sequence of densities
\begin{equation}
    p^{\mathrm{(bridge)}}_t(\theta_t|x^{1},\dots,x^{n}) \propto (p_t(\theta_t))^{1-n} \prod_{i=1}^n p_t(\theta_t|x^{i}).\label{eq:our_densities}
\end{equation}
This sequence of densities has all of the desirable properties of \eqref{eq:geffner}. The density at $t=0$ coincides with the target $p(\theta_t|x^{1},\dots,x^{n})$, and the density at $t=T$ is a tractable Gaussian. Moreover, we can factorise these densities in terms of the single-observation posterior scores $\nabla_{\theta} \log p_t(\theta_t|x^{i})$, and the perturbed {prior} score $\nabla_{\theta} \log p_t(\theta_t)$, as
\begin{equation}
    \nabla_{\theta} \log p^{\mathrm{(bridge)}}_t(\theta_t|x^1,\dots,x^n) = (1-n)\nabla_{\theta} \log p_t(\theta_t) + \sum_{i=1}^n \nabla_{\theta} \log p_t(\theta_t|x^i). 
    \label{eq:our_scores}
\end{equation}
Similar to above, it is then only necessary to learn a single score network $s_{\psi_{\text{post}}}(\theta_t,x,t)\approx \nabla_{\theta} \log p_t(\theta_t|x)$, which we can train using samples $(\theta,x)\sim p(\theta)p(x|\theta)$. Clearly, the expressions in \eqref{eq:our_densities} - \eqref{eq:our_scores} are very similar to the ones given in \eqref{eq:geffner} - \eqref{eq:geffner_scores}, with the only difference appearing in the first term. These quantities coincide at time zero, but will otherwise differ. The advantage of \eqref{eq:geffner_scores}, i.e., the scheme proposed in \citet{Geffner2022}, is that it only requires access to the score of the prior (rather than the score of the perturbed prior), and is thus very straightforward to implement.

\section{Additional Experimental Details} 

\subsection{Benchmark Tasks}
\label{sec:benchmarks}

We consider the following set of benchmark tasks, described in \citet[Appendix T]{Lueckmann2021}. 

\paragraph{Gaussian Linear.}
This simple experiment involves inferring the mean of a 10-dimensional Gaussian, in which the covariance is fixed. The prior is a Gaussian, given by $p(\theta) = \mathcal{N}(0,0.1 \mathbf{I})$, as is the simulator, $p(x|\theta) = \mathcal{N}(x|\theta,0.1 \mathbf{I})$.

\paragraph{Gaussian Mixture.} 
This task, introduced by \citet{Sisson2007}, appears frequently in the SBI literature \cite{Beaumont2009,Lueckmann2021}. It consists of a uniform prior $p(\theta)=\mathcal{U}(-10,10)$, and a simulator given by $p(x|\theta) = 0.5\mathcal{N}(x|\theta,\mathbf{I}) + 0.5\mathcal{N}(x|\theta,0.01\mathbf{I})$, where $\theta,x\in\mathbb{R}^2$.

\paragraph{Two Moons.} 
This two-dimensional experiment consists of a uniform prior given by $p(\theta)=\mathcal{U}(-1,1)$, $\theta \in \mathbb{R}^2$, and a simulator defined by
\begin{align}
    x|\theta = \begin{pmatrix} r\cos(\alpha) + 0.25 \\ r\sin(\alpha) \end{pmatrix} + \begin{pmatrix} -|\theta_1 + \theta_2|/\sqrt{2} \\ (-\theta_1 + \theta_2)/\sqrt{2} \end{pmatrix} 
\end{align}
where $\alpha\sim \mathcal{U}(-\pi/2, \pi/2)$ and $r\sim \mathcal{N}(0.1,0.01^2)$. It defines a posterior distribution over the parameters which exhibits both local (crescent shaped) and global (bimodal) features, and is frequently used to analyse how SBI methods deal with multimodality \cite{Greenberg2019,Glockler2022}. 

\paragraph{Gaussian Linear Uniform.} 
This task consists of a uniform prior $p(\theta) = \mathcal{U}(-1,1)$, and a Gaussian simulator $p(x|\theta) = \mathcal{N}(x|\theta, 0.1\mathbf{I})$, where $\theta,x\in\mathbb{R}^{10}$. This example allows us to determine how algorithms scale with increased dimensionality, as well as with truncated support. 

\paragraph{Bernoulli GLM.} This experiment consists of a generalised linear model (GLM) with Bernoulli observations, used to simulate the activity of a neuron depending on a single set of covariates \cite{DeNicolao1997,Lueckmann2017}. The task is to infer a 10-dimensional parameter $\theta=(\beta,\mathbf{f})\in\mathbb{R}^{10}$, where $\beta\sim\mathcal{N}(0,2)$ and $\mathbf{f}\sim\mathcal{N}(0,(\mathbf{F}^T\mathbf{F})^{-1})$, where $\mathbf{F}$ encourages smoothness by penalizing the second-order differences in the vector of
parameters. The observed data $\mathbf{x}\in\mathbb{R}^{10}$ are the sufficient statistics for this GLM.

\paragraph{SLCP.}
This task, introduced by \citet{Papamakarios2019}, is designed to have a simple likelihood and a complex posterior. The prior is a five-dimensional uniform distribution $p(\theta) = \mathcal{U}(-3,3)$, while the likelihood for the eight-dimensional data is Gaussian, but with mean and covariance which are highly non-linear functions of the parameters. This defines a complex posterior distribution over the parameters, with four symmetrical modes and vertical cut-offs. 

\paragraph{SIR.}
This is an epidemiological model in which individuals from a population move between 3 possible compartments: (S)usceptible, (I)nfected and (R)emoved. The task involves inferring a two-dimensional model parameter $\theta=(\beta,\gamma)$, where $\beta \sim \mathrm{LogNormal}(\log(0.4), 0.5)$ is the contact rate between susceptible and infected, and $\gamma \sim \mathrm{LogNormal}(\log(0.8), 0.2)$ is the mean removal rate. The data $x=(x_1,\dots,x_{10})\in\mathbb{R}^{10}$ consist of 10 equally spaced noisy recordings $x_i \sim \mathrm{Bin}(1000, \frac{I}{N})$, where $I$ denotes the number of individuals in the infected compartment, and is simulated according to a set of ODEs. 

\paragraph{Lotka Volterra.}
This experiment involves a classical model used in ecology to model predator-prey populations \cite{Lotka1920}. The task is to infer a four dimensional parameter $\theta = (\alpha, \beta, \gamma, \delta)$ which governs the growth rates and the interactions of the predator and prey populations, which are described by a system of ODEs. The priors for these parameters are given by $\alpha, \gamma \sim \mathrm{LogNormal}(-0.125, 0.5)$, and $\beta, \delta \sim \mathrm{LogNormal}(-3, 0.5)$. The observed data $x=(x_1,\dots,x_{10})\in\mathbb{R}_{+}^{20}$, with each $x_i\in\mathbb{R}_{+}^2$, consist of 10 evenly spaced recordings of both predator and prey populations.

\subsection{Real-World Neuroscience Problem}
\label{sec:pyloric-additional}

\paragraph{Additional Implementation Details.} To deal with ill-defined summary statistics, we follow the approach adopted by \citet{Deistler2022}, and replace invalid summary statistics with a value two standard deviations below the prior predictive of the summary statistic. We use the VP SDE to diffuse samples (see Appendix \ref{ap:sde} for more details).

\begin{figure}[H]
  \centering
\includegraphics[width=.93\textwidth, trim=70 70 70 70, clip]{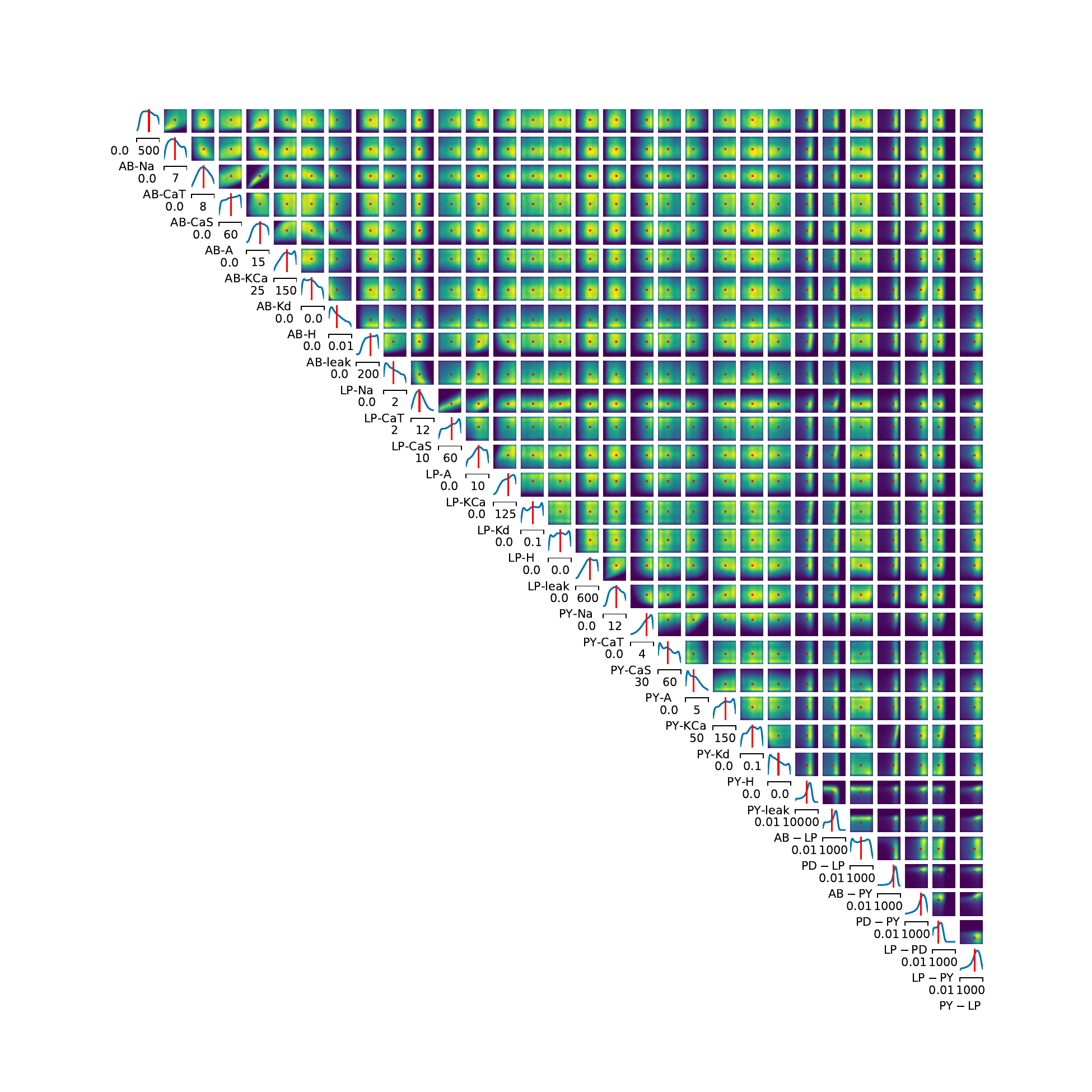}
\vspace{-2mm}
    \caption{\textbf{Pairwise marginal plot for the posterior approximation obtained in the Pyloric experiment. The posterior mean is plotted in red.}}
  \label{fig:pyloric-pairwise-marginal}
\vspace{-3mm}
\end{figure}

\begin{figure}[H]
  \centering
\includegraphics[width=0.35\textwidth, trim=00 00 00 00, clip]{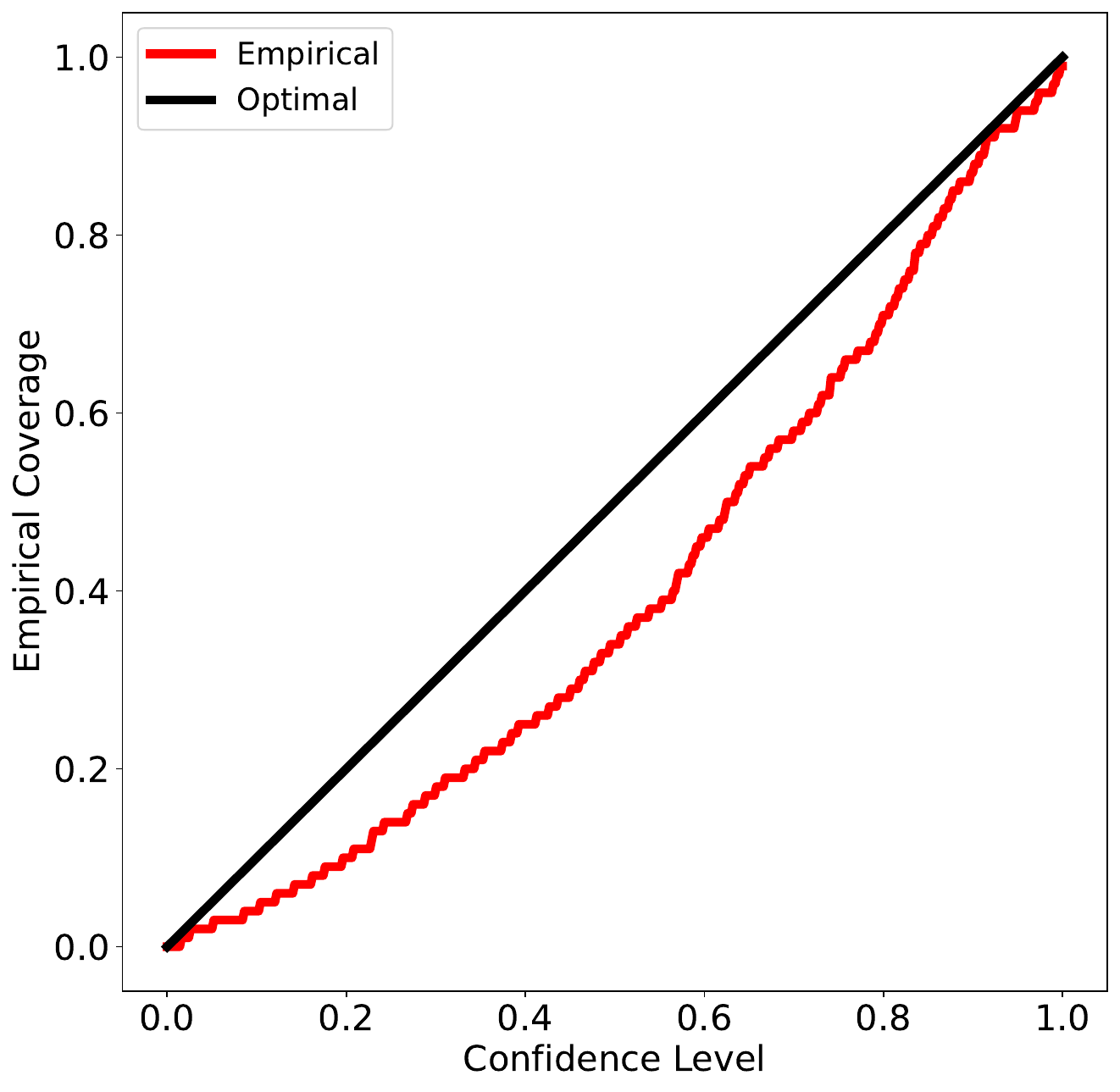}
\vspace{-3mm}
    \caption{\textbf{Coverage plot for the Pyloric experiment.}}
  \label{fig:pyloric-sbcc}
\end{figure}

\paragraph{Additional Numerical Results.} Additional results for this experiment are provided in Figures \ref{fig:pyloric-pairwise-marginal} - \ref{fig:pyloric-sbcc}. In Figure \ref{fig:pyloric-pairwise-marginal}, we provide a pairwise marginal plot for the posterior approximation obtained by TSNPSE. Our approximation has similar characteristics to those previously obtained in the literature; see, e.g., \citet{Deistler2022} and \citet{Glockler2022}. Meanwhile, Figure \ref{fig:pyloric-sbcc} shows the expected coverage of the approximate posterior, computed according to the simulation-based coverage calibration (SBCC) procedure described in \citet{Deistler2022}. This plot indicates that, for mid-low confidence levels, the empirical expected coverage is smaller than the confidence level (i.e., the posterior is overconfident). Importantly, however, the empirical expected coverage approximately matches the confidence level for high confidence levels. We expect that, as suggested in \citet{Hermans2022}, an ensemble of approximate neural posteriors estimators could be used to obtain a more conservative posterior.

\vspace{-4mm}

\subsection{Implementation Details}
\label{sec:implementation}

\subsubsection{SDE} \label{ap:sde}
For the benchmark experiments, we consider two different choices for the forward noising process: the variance exploding SDE (VE SDE) and the variance preserving SDE (VP SDE). See \citet[Appendices B - C]{Song2021} for further details.

\paragraph{VE SDE.} The VE SDE is defined according to
\begin{equation}
    \mathrm{d}\theta_t = \sigma_{\text{min}}\left(\tfrac{\sigma_{\min}\bigstrut[b]}{\sigma_{\max}}\right)^t\sqrt{2\log \tfrac{\sigma_{\max}\bigstrut[b]}{\sigma_{\min}}}\mathrm{d}w_t~,~~~t\in(0,1]. \label{eq:VESDE}
\end{equation}
We set $\sigma_{\text{min}}=0.01$ for the 2-dimensional experiments, SIR and Two Moons, and $\sigma_{\text{min}}=0.05$ for all other experiments, while $\sigma_{\text{max}}$ is chosen according to Technique 1 in \citet{Song2020a}. This SDE defines the transition density \begin{equation}
    p_{t|0}(\theta_t|\theta_0) = \mathcal{N}\left(\theta_t\left|\theta_0, \sigma_{\min}^2\left(\tfrac{\sigma_{\max}\bigstrut[b]}{\sigma_{\min}}\right)^{2t}\mathbf{I}\right.\right).
\end{equation}

\paragraph{VP SDE.}
The VP SDE is defined according to
\begin{equation}
    \mathrm{d}\theta_t = - \frac{1}{2} \beta_t\theta_t \mathrm{d}t + \sqrt{\beta_t} \mathrm{d}w_t~,~~~t\in(0,1], \label{eq:VPSDE}
\end{equation}
where $\beta_t = \beta_{\text{min}} + t (\beta_{\text{max}} - \beta_{\text{min}})$.
In our experiments, we set $\beta_{\text{min}}=0.1$ and $\beta_{\text{max}}=11.0$, following \citet{Song2020a}. This SDE defines the transition density \begin{equation}
    p_{t|0}(\theta_t|\theta_0) = \mathcal{N}\left(\theta_t\left|\theta_0 e^{\frac{1}{2}\int_{0}^{t}\beta_s\mathrm{d}s}, \mathbf{I} - \mathbf{I} e^{\int_{0}^{t}\beta_s\mathrm{d}s}\right.\right).
\end{equation}

\subsubsection{Network Architecture and Training} 
\label{ap:hyperparameters}

\paragraph{\textbf{$\theta_t$ embedding network}.} 3-layer fully-connected MLP with 256 hidden units in each layer. The input dimension is $d$ $(\theta\in\mathbb{R}^{d})$ and the output dimension from the final layer is determined by $\mathrm{max}(30, 4\cdot d)$. We denote this embedding $\theta_\mathrm{emb}$.

\paragraph{\textbf{$x$ embedding network}.} 3-layer MLP fully-connected with 256 hidden units in each layer. The input dimension is $p$ $(x\in\mathbb{R}^{p})$ and the output dimension from the final layer is determined by $\mathrm{max}(30, 4\cdot p)$. We denote this embedding $x_\mathrm{emb}$.

\paragraph{\textbf{$t$ sinusoidal embedding}.} We embed $t$ into 64 dimensions, denoted $t_\mathrm{emb}$. Inspired by \citet{Vaswani2017}, we use sinusoidal embeddings defined by
\begin{align} 
    (t_\mathrm{emb})_i = \begin{dcases}
      \sin\left(\frac{t}{10000^{(i-1)/31}}\right) & \text{if $i \leq 32$,} \\
      \cos\left(\frac{t}{{10000^{((i-32)-1)/31}}}\right) & \text{if $i > 32$.}
    \end{dcases}
\end{align}

\paragraph{\textbf{Score network}.} Finally, we concatenate $[\theta_\mathrm{emb}, x_{\mathrm{emb}}, t_{\mathrm{emb}}]$ and feed this into a 3-layer fully-connected MLP with 256 hidden units in each layer whose output dimension is $d$.

\paragraph{\textbf{Activation Function}.} We use SiLU activation functions between layers for all MLP networks.

\paragraph{\textbf{Optimizer}.} We use Adam \cite{Kingma2015} to train the networks with a learning rate of $1 \times 10^{-4}$. We hold $15\%$ of the data back as a validation set: we compute the loss function on these samples after each training step, if this loss does not decrease for 1000 steps then we stop training and return the network which gave the lowest validation loss. The maximum number of training iterations is 3000 for sequential experiments and 3000 for non-sequential experiments. For experiments with a simulation budget of either 1000 or 10000, our batch size is 50 for non-sequential experiments and 200 for sequential experiments. For simulation budgets of 100000 we employ a bigger batch size of 500 for both sequential and non-sequential. 

\subsubsection{Miscellaneous}
\label{sec:misc}
\paragraph{\textbf{Sampling}.} We use the probability flow ODE for sampling. To solve this ODE, we use an off-the-shelf solver (RK45). 

\paragraph{\textbf{Estimating $\mathrm{\mathbf{HPR}}_\varepsilon$ and Sampling the Truncated Proposal}.} We follow the approach described in \citet[Section 6.3]{Deistler2022}. We first simulate 20000 samples from our approximate posterior via (the time-reversal of) the probability flow ODE in \eqref{eq:backwardODE} using our approximation of $\nabla_\theta \log p_t(\theta_t|x_\mathrm{obs})$. We then compute the likelihood of these samples via the instantaneous change-of-variables formula in \eqref{eq:change_of_var}. Finally, we calculate the truncation boundary, $\kappa$, by taking the $(\varepsilon = 5 \times 10^{-4})^{\mathrm{th}}$ quantile. This quantity defines the log-probability rejection threshold for rejection sampling. 

To sample the truncated proposal, we use rejection sampling. In particular, we repeat the following procedure until the appropriate number of samples have been accepted: sample $\theta \sim p(\theta)$, compute the likelihood under our approximate posterior using the instantaneous change-of-variables formula in \eqref{eq:change_of_var}, and accept the sample if the likelihood is greater than $\kappa$, otherwise reject. As outlined in \citet[Section 3.2]{Deistler2022a}, one could also use other sampling schemes besides rejection sampling, such as Sampling Importance Resampling (SIR). 

In practice, computing likelihoods via the instantaneous change of variables formula in \eqref{eq:change_of_var} is a computationally expensive procedure, and thus we introduce an additional rejection sampling step to minimise the number of likelihood evaluations required.\footnote{See Appendix \ref{app:alt-param} for an alternative approach which reduces the cost of computing an (unnormalised) likelihood to the cost of a single forward pass of the neural network.} In particular, we first perform a cheap initial rejection step directly on the prior samples by identifying whether they are within the (empirical) hypercube occupied by the approximate posterior samples. This is appropriate since the support of the approximate posterior typically takes up a fraction of the prior space; and can significantly reduce the number of likelihood calculations required.  

\paragraph{\textbf{Standardization}.} We centre both $\theta_t$ and $x$ before being input into the score network, by subtracting an estimate of the mean and dividing by the standard deviation in each dimension. 

\paragraph{\textbf{Number of Rounds}} We use $R=10$ rounds for all sequential experiments, unless otherwise specified. The simulation budget is equally divided between rounds.

\section{Comparison to Flow Matching Posterior Estimation}
\label{app:fmpe}
\begin{figure*}[h!]
  \centering
  \includegraphics[width=\textwidth, trim=90 20 40 20, clip]{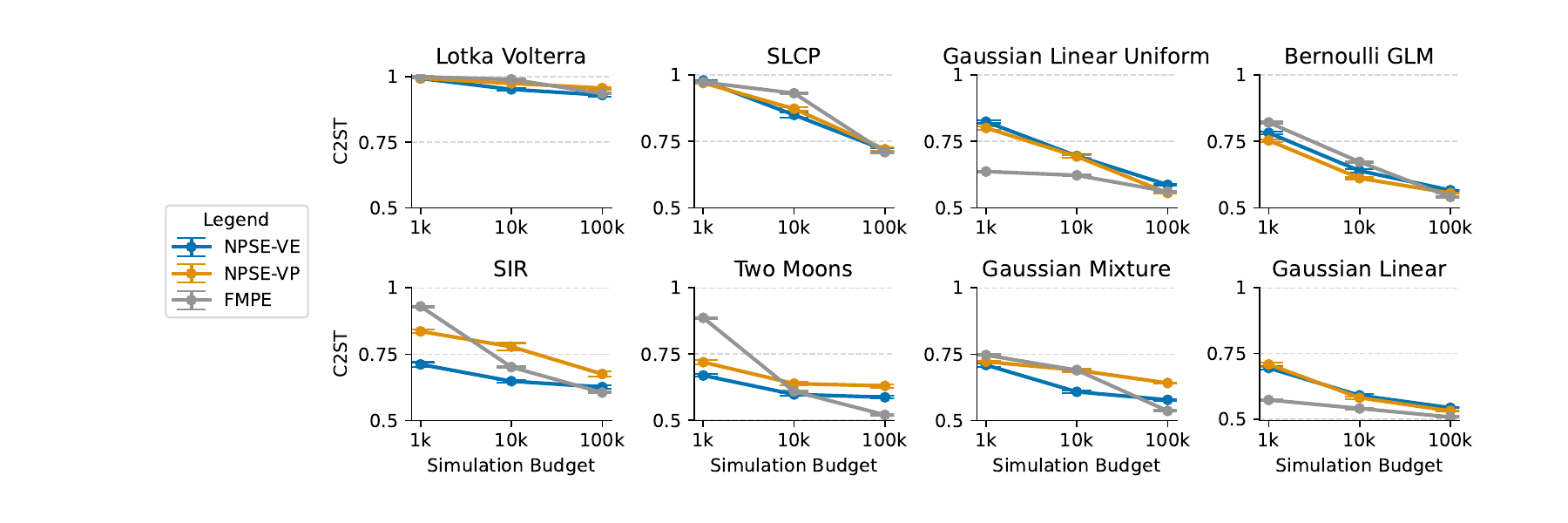}
  \vspace{-7mm}
  \caption{\textbf{Comparison between NPSE and FMPE on eight benchmark tasks.}}
  \label{fig:nonseq-fmpe}
\end{figure*}

In this section we provide an additional comparison between our non-sequential method (NPSE) and flow matching posterior estimation (FMPE)  \citep{Dax2023}. In particular, results for the eight SBI benchmark experiments described in Section \ref{sec:benchmarks}, for simulation budgets of 1000, 10000, and 100000, are provided in Figure \ref{fig:nonseq-fmpe}. 

It is worth noting that the results for FMPE are taken directly from \citet{Dax2023}. As a consequence, there is a difference in the way in which hyperparameters are tuned between the two methods. For FMPE, as detailed in \citet[Appendix C]{Dax2023}, hyperparameter tuning is performed on an experiment-by-experiment basis. This is achieved by sweeping over values of 5 hyperparameters (layer width, number of layers, learning rate, batch size, and a loss-weighting hyperparameter) for each experiment, and selecting the model which has the lowest validation loss on a held-out dataset. In contrast, NPSE uses a single set of hyperparameters for all experiments. 

\section{Alternative Parameterisation of the Score Network} \label{app:alt-param}
In this section we discuss how an alternative energy-based parameterisation of the diffusion model \citep[e.g.,][]{Du2023} can significantly reduce the computational cost of TSNPSE. Specifically, this parameterisation circumvents the need to use the instantaneous change-of-variables formula \eqref{eq:change_of_var} to compute likelihoods, which is required for the truncated proposal used in TSNPSE (see Section \ref{sec:misc}). 

As discussed in \citet{Salimans2021,Du2023}, there are multiple ways to parameterise the posterior score estimate used in diffusion models. In this paper, we have directly modelled the score using a vector-valued score network $s_{\psi}:\mathbb{R}^d\times\mathbb{R}^p\times[0,T]\rightarrow\mathbb{R}^d$. As an alternative, one could parameterise a scalar-valued energy function $E_{\psi}:\mathbb{R}^d\times\mathbb{R}^p\times[0,T]\rightarrow\mathbb{R}$, and then model the score as the gradient of this function: $s_{\psi}(\theta_t,x,t) =-\nabla_{\theta_t} E_{\psi}(\theta_t,x,t)$.\footnote{There are various ways in which one could parameterise the energy function $E_{\psi}$. The simplest is to parameterise $E_{\psi}$ as a feedforward neural network, whose final layer has a single output \citep[e.g.,][]{Nijkamp2020}. Several other parameterisations have also been considered in \citet{Salimans2021,Du2023}.}
In this case, we automatically also obtain an estimate of the perturbed posterior density as 
\begin{equation}
    p_t(\theta_t|x) \approx \frac{\exp \left[ - E_{\psi}(\theta_t,x,t)\right]}{Z_{\psi}(x,t)},
\end{equation}
where $Z_{\psi}(x,t) = \int_{\mathbb{R}^d} \exp\left[ - E_{\psi}(\theta_t,x,t)\right]\mathrm{d}\theta_t$ is an (unknown) normalising constant. Recalling that $p_0(\cdot|x):=p(\cdot|x)$, and noting that the normalising constant is independent of $\theta$, it follows in particular that
\begin{equation}
p(\theta|x_{\mathrm{obs}}) 
\appropto \exp\left[ - E_{\psi}(\theta,x_{\mathrm{obs}},0)\right].
\end{equation}
Now, observe that estimating $\mathrm{HPR}_\varepsilon$ and sampling the truncated proposal (i.e., evaluating if samples are inside $\mathrm{HPR}_{\varepsilon}$) actually only requires knowledge of the likelihood up to a normalisation constant (see Section \ref{sec:misc}). Thus, under the energy-based parameterisation, estimating $\mathrm{HPR}_\varepsilon$ and sampling the truncated proposal only requires a single forward pass of the neural network $E_{\psi}$. This is significantly cheaper than using a score-based parameterisation, which not only requires multiple forward passes, but also the gradient of multiple forward passes, to solve \eqref{eq:backwardODE}  and \eqref{eq:change_of_var}. 

It is worth noting that the choice of parameterisation can have a significant impact on the sample quality of (unconditional) score-based generative models; see, e.g., \citet[][Section 4.1]{Salimans2021} or \citet[][Appendix E]{Du2023}. It is likely that the same is true for the conditional score-based generative models used in TSNPSE. With this in mind, further empirical testing is required to understand whether the speed-up associated with the use of an energy-based parameterisation in TSNPSE comes at any cost in terms of sample quality.

\end{document}